\documentclass{article}
\usepackage{iclr2027_conference,times}

\usepackage{amsmath,amsfonts,bm}

\def\eqref#1{equation~\ref{#1}}

\def\1{\bm{1}}

\DeclareMathAlphabet{\mathsfit}{\encodingdefault}{\sfdefault}{m}{sl}
\SetMathAlphabet{\mathsfit}{bold}{\encodingdefault}{\sfdefault}{bx}{n}

\usepackage{hyperref}
\usepackage{url}
\usepackage{graphicx}
\usepackage{tikz}
\usepackage{multirow}
\usepackage{amsmath,amssymb}
\usepackage{algorithm}
\usepackage{algpseudocode}
\usepackage{pdfpages}
\usepackage{listings}
\usepackage{wrapfig}
\usepackage{placeins}

\title{How Much Can We Prune? Exploring the Performance Limits of Visual Token Pruning}

\iclrfinalcopy
\author{%
\textbf{Janghoon Cho, Shubhankar Borse, Reza Pourreza}\\
\textbf{Juntae Lee, Jihwan Bang, Munawar Hayat, Fatih Porikli}\\[4pt]
Qualcomm AI Research\thanks{Qualcomm AI Research is an initiative of Qualcomm Technologies, Inc.}\\
}

\begin{document}
\addtocontents{toc}{\protect\setcounter{tocdepth}{-1}}

\maketitle
\fancyhead[L]{Preprint}
\begingroup
\setlength{\parskip}{3pt plus 0.5pt minus 0.5pt}
\setlength{\textfloatsep}{10pt plus 2pt minus 2pt}
\setlength{\floatsep}{8pt plus 2pt minus 2pt}
\setlength{\intextsep}{6pt plus 1pt minus 1pt}
\setlength{\abovecaptionskip}{4pt}
\setlength{\belowcaptionskip}{0pt}
\setlength{\columnsep}{12pt}
\makeatletter
\renewcommand{\paragraph}{\@startsection{paragraph}{4}{\z@}{0.8ex plus 0.2ex minus 0.1ex}{-1em}{\normalsize\bfseries}}
\makeatother
\raggedbottom
\begin{abstract}
Visual token pruning is increasingly used to reduce the memory and inference costs of vision-language models. Existing work has largely focused on designing method-specific selectors and demonstrating incremental improvements over prior methods at selected token budgets. It remains unclear whether the resulting performance reflects an inherent limit of visual token pruning or the limitations of current token-selection algorithms. We take a complementary perspective and ask: \emph{how well can visual token pruning perform when offline search cost is not the primary constraint?} We formulate pruning for general image understanding as selecting, under a fixed budget, one shared visual-token subset that supports a broad distribution of questions about an image. We approximate this utility with a finite auxiliary QA set generated by a frontier VLM and optimize a teacher-forced answer-probability objective with the evaluated model frozen. Motivated by the conceptual monotonicity and diminishing returns of adding visual evidence, we construct nested subsets greedily and introduce \emph{QA mini-batch greedy} (QA-MBG), which estimates token marginal gains from randomly sampled QA components at each step. Empirically, QA-MBG is both cheaper and more effective than full greedy. With one frontier-VLM call and approximately $0.5$ H100 GPU-hours per image, it produces a nested subset trajectory covering keep ratios up to $0.3$. Across Qwen3-VL-4B and Gemma~4 E4B on $900$ images from nine benchmarks, we find significant headroom over evaluated pruning baselines on held-out QA and image-grounded long-form generation, while the searched subsets substantially narrow the gap to full-token inference.
\end{abstract}

\section{Introduction}
\label{sec:introduction}

Vision-language models (VLMs) obtain strong visual understanding by converting an image into hundreds or thousands of visual tokens and processing them together with text \citep{bai2025qwen3vl,wang2025internvl35}. These tokens substantially increase language-model prefill computation, decoding-time attention cost, KV-cache memory, and end-to-end latency. Visual token pruning has therefore become a major direction for efficient VLM inference. Recent methods score or consolidate tokens using attention, query relevance, feature similarity, redundancy, diversity, or learned selection modules, often retaining only a small fraction of the original sequence \citep{chen2024fastv,alvar2025divprune,wen2025dart,yang2025visionzip}.

Most work follows the same evaluation pattern: propose a new selection heuristic and demonstrate an incremental accuracy--efficiency improvement over previous selectors at selected token budgets. This competition improves deployable algorithms, but leaves a more fundamental question unanswered: \emph{how much performance can visual token pruning actually preserve at a fixed budget?} A gap from full-token inference conflates unavoidable information loss with recoverable selector error, while small differences among current methods do not show that they are close to the best available subset. Diagnostic studies further reveal strong sensitivity to evaluation protocol and cases in which simple rules rival elaborate methods \citep{wen2025rightproblem,peng2025uniprunebench,liao2026vtcbench}. The achievable performance frontier of visual token pruning remains unexplored.

We expose this frontier by relaxing the search-time compute constraint. Rather than designing another inexpensive inference-time heuristic, we invest substantial offline computation to discover a strong subset for each image. Every discovered subset is a feasible operating point at its retained-token budget; a large, reproducible gap over existing selectors therefore directly measures algorithmic headroom. This changes the question from which current heuristic wins to how much room remains for better pruning algorithms.

The key challenge is defining subset quality for \emph{general image understanding}. A subset selected for one known query can discard evidence needed by another. We instead seek one shared subset that supports the distribution of questions that could reasonably be asked about an image, formalized as expected question-answering utility under a cardinality constraint. Because this distribution is unavailable, we approximate it with a finite, diverse auxiliary QA set generated by a frontier VLM. The evaluation questions and answers are never used to score candidate subsets in the main experiment; after search, the same nested trajectory is reused for held-out benchmark questions and long-form description prompts.

This formulation connects visual-token selection to classical information-gathering problems, where adding observations yields monotone, diminishing-return utility and enables near-optimal greedy selection \citep{krause2008nearoptimal}. Frozen-VLM QA utility is only approximately monotone and submodular, but it exhibits the same saturation pattern empirically. We therefore build a nested trajectory by repeatedly adding the token with the largest estimated marginal gain. Full greedy is prohibitively expensive because every step evaluates every remaining token against every auxiliary QA.

We introduce \emph{QA mini-batch greedy} (QA-MBG), which samples a fresh mini-batch of auxiliary questions at each step, evaluates all remaining token candidates on those components, and adds the best token. This is the decomposable-objective counterpart of mini-batch submodular maximization \citep{schwartzman2024minibatch}, and differs from stochastic greedy, which samples token candidates while evaluating the full objective \citep{mirzasoleiman2015lazier}. QA-MBG reduces QA evaluations and improves the irreversible sequence of greedy decisions: it raises held-out QA and long-form performance over full greedy while producing more balanced gains across auxiliary questions. Candidate batching, prefix-KV reuse, and teacher-forced scoring make the offline search practical. One frontier-VLM call and approximately $0.5$ H100 GPU-hours per image produce a complete nested trajectory through keep ratio $0.3$.

We evaluate Qwen3-VL-4B and Gemma~4 E4B on $900$ images and $2{,}226$ held-out questions from nine benchmarks. Across both model families, searched subsets establish a clear empirical frontier above the evaluated pruning baselines and substantially narrow the gap to full-token inference. The gains are especially pronounced on document, chart, infographic, and diagram understanding, and transfer to image-grounded long-form generation despite search using a short-answer objective.

Our contributions are threefold:
\begin{itemize}
    \item We formulate visual token pruning for general image understanding as selecting one shared subset that maximizes expected QA utility under a fixed token budget, and use frontier-generated auxiliary QAs to expose its empirical performance frontier.
    \item We introduce QA mini-batch greedy, which samples objective components rather than token candidates and improves both search cost and downstream subset quality relative to full greedy and token-level stochastic greedy.
    \item Across two VLM families, nine benchmarks, held-out QA, and long-form generation, we reveal substantial headroom over existing pruning methods, showing that current selectors are far from exhausting what is achievable at the same visual-token budgets.
\end{itemize}

\section{Related Work}
\label{sec:related-work}

\paragraph{Visual token pruning in VLMs.}
Visual token pruning reduces multimodal inference cost by retaining a subset of visual tokens before or within the language model. Existing selectors use multi-stage dropping \citep{liu2024mustdrop}, attention or head importance \citep{zhu2026hawk,gao2026quietprune}, learned selection \citep{zhu2025visionselector}, diversity \citep{alvar2025divprune}, redundancy \citep{wen2025dart}, or combinations of relevance and contextual coverage \citep{pei2025greedyprune,zhang2025cdpruner,oh2026anchorprune,wang2026eadp}. Query-conditioned selection targets evidence for a particular question, whereas query-agnostic selection seeks to retain information without access to that question; SparseVILA explicitly distinguishes these settings \citep{khaki2025sparsevila}. Adaptive methods additionally vary the budget with the input, as in GlimpsePrune, PromPrune, and OccamToken \citep{zeng2025glimpseprune,lee2026promprune,li2026occamtoken}. These approaches primarily target efficient, deployable selection. Our focus is on the performance achievable by a shared image-token subset at a specified budget, selected using auxiliary QA supervision and reused without conditioning selection on the evaluation question.

\paragraph{Oracle analyses and search-based selection.}
SparseVILA constructs a query-aware greedy oracle that selects visual tokens to maximize agreement with the unpruned model's responses, and uses its deterioration across subsequent conversation rounds to illustrate the limitations of irreversible query-dependent pruning \citep{khaki2025sparsevila}. Our setting instead uses diverse auxiliary QA pairs to search for one subset intended to support multiple evaluation questions. EvoComp performs evolutionary search for visual-token masks that minimize response loss, constrains selection through semantic grouping, and uses the resulting masks to supervise a learned compressor \citep{song2026evocomp}. We share the use of answer-based search signals, but study a different target: the achievable performance of a shared multi-QA subset and the effect of search granularity on that performance, rather than compressor training. Related oracle analyses include the label-conditioned Grad-CAM oracle in video classification studied by Principles of Visual Tokens \citep{hao2025principles}. Our contribution therefore lies in combining shared-subset exploration, evaluation with target-question annotations withheld from candidate scoring, and an analysis of QA mini-batch greedy, rather than in introducing answer-guided token search itself.

\paragraph{Subset optimization.}
Classical stochastic greedy reduces oracle evaluations in cardinality-constrained subset selection by sampling candidate elements at each iteration \citep{mirzasoleiman2015lazier}. Mini-batch submodular maximization instead considers a decomposable objective, $F=\sum_i f_i$, and resamples objective components at every greedy step \citep{schwartzman2024minibatch}. Our auxiliary-QA objective has the same decomposition across questions: QA-MBG samples QA components while evaluating every remaining visual-token candidate. This directly separates QA-MBG from token-candidate stochastic greedy and provides the closest combinatorial-optimization precedent for our search rule. Because frozen-VLM answer probability is only approximately monotone and submodular, its existing approximation guarantee does not directly transfer; we therefore evaluate both the full auxiliary objective and downstream performance of the resulting trajectories.

\section{Method}
\label{sec:method}

\begin{figure}[t]
 \centering
 \resizebox{\linewidth}{!}{\input{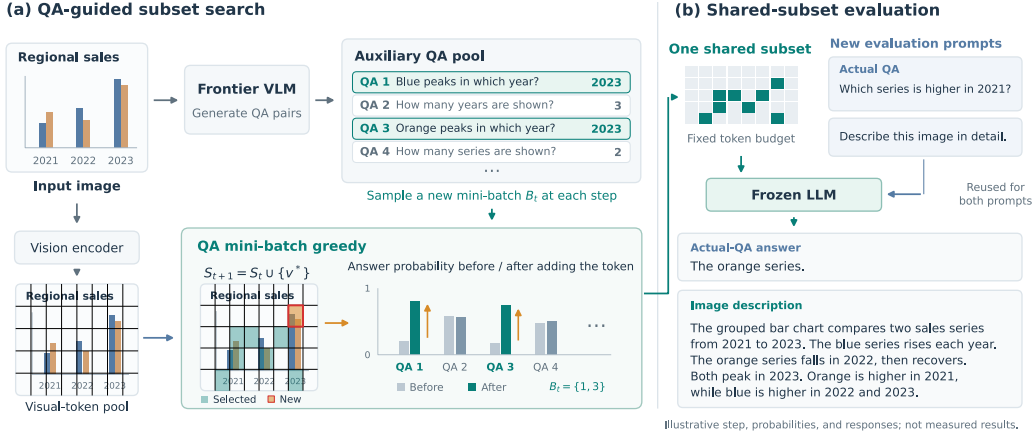}}
 \caption{QA-guided subset search and shared-subset evaluation.}
 \label{fig:overview}
\end{figure}

Figure~\ref{fig:overview} illustrates the pipeline: auxiliary QAs guide subset search, and the selected visual tokens are reused across evaluation prompts. The highlighted patch and probability changes illustrate one possible search step rather than measured effects for that example.

\subsection{Problem formulation: shared subsets for image understanding}
\label{sec:formulation}

\paragraph{A fixed visual-token budget.}
Let $I$ be an image and $Z_I=(z_1,\ldots,z_n)$ its visual embeddings at the input to a frozen language model with parameters $\theta$. We select indices from $V_I=\{1,\ldots,n\}$ after visual encoding and before language-model processing. For $S\subseteq V_I$, $Z_I[S]$ denotes the retained embeddings, with their original order and positional information preserved. Given a keeping ratio $\rho\in[0,1]$, the budget is $b=\lfloor\rho n\rfloor$. Our goal is to find a single subset of this size that preserves general image understanding across questions. Writing the abstract understanding ability as $U_I$, the ideal problem is
\begin{equation}
 S^{\star}_{U,b}\in\operatorname*{arg\,max}_{S\subseteq V_I,\;|S|=b} U_I(S).
 \label{eq:ideal-subset}
\end{equation}
The subset depends on the image and budget, but is shared across evaluation questions. We use an exact cardinality constraint for subset search. Experiments report realized token counts because model-specific rounding can produce small differences at the same requested ratio.

\paragraph{Conceptual structure of understanding.}
Additional useful visual clues should expand what can be understood about an image. A new clue also often contributes less when much of the relevant evidence is already available. These intuitions motivate an idealized monotone submodular view of $U_I$: for $A\subseteq B\subseteq V_I$ and $v\notin B$,
\begin{align}
 U_I(A)&\leq U_I(B), \label{eq:conceptual-monotone}\\
 U_I(A\cup\{v\})-U_I(A)&\geq U_I(B\cup\{v\})-U_I(B).
 \label{eq:conceptual-submodular}
\end{align}
Equation~(\ref{eq:conceptual-monotone}) expresses monotonicity and Eq.~(\ref{eq:conceptual-submodular}) diminishing returns. These properties describe the predominant effect of accumulating visual evidence: although counterexamples can arise from complementary clues, entangled embeddings, or distraction by additional tokens, most additions preserve or improve usable evidence and contribute less once related evidence is already available. We therefore use monotone submodularity as an approximate structural model for guiding incremental selection.

\paragraph{From abstract ability to question-answering probability.}
To operationalize understanding, let $\mathcal{D}_I$ be a target distribution over image-grounded questions, covering the aspects of $I$ that a user could ask about. For a question $q$, let $\mathcal{A}_I(q)$ denote its set of correct answers. A model-dependent measure of image understanding is the expected probability of answering correctly:
\begin{equation}
 F_I(S)=\mathbb{E}_{q\sim\mathcal{D}_I}
 \left[\Pr_{\widehat a\sim p_\theta(\cdot\mid q,Z_I[S])}
 \bigl(\widehat a\in\mathcal{A}_I(q)\bigr)\right],
 \qquad
 S^{\star}_{F,b}\in\operatorname*{arg\,max}_{S\subseteq V_I,\;|S|=b}F_I(S).
 \label{eq:population-objective}
\end{equation}
The distribution over questions specifies which kinds of understanding are valued; there is no uniform distribution over all possible natural-language questions.

\paragraph{Finite auxiliary QA approximation.}
Neither the question distribution nor all correct answers are available. We therefore prompt a frontier VLM with the image to produce a finite auxiliary set $\mathcal{Q}_I=\{(q_j,a_j)\}_{j=1}^{M}$. The generation prompt seeks diverse, image-grounded questions spanning the evidence types appropriate to the target task distribution. General understanding here is operationalized as support for multiple questions through one shared subset, with transfer to general description prompts evaluated separately. A generated answer provides reference supervision, rather than an exhaustive set of valid answers. Let $f_j(S)$ be the reference-answer score for $(q_j,a_j)$. We optimize
\begin{equation}
 \widehat F_I(S)=\frac{1}{M}\sum_{j=1}^{M}f_j(S),
 \qquad
 S^{\star}_{\mathrm{aux},b}\in\operatorname*{arg\,max}_{S\subseteq V_I,\;|S|=b}\widehat F_I(S).
 \label{eq:empirical-objective}
\end{equation}
For a tokenized reference answer $a_j=(a_{j,1},\ldots,a_{j,L_j})$, we score each answer token by teacher forcing the preceding reference tokens:
\begin{equation}
 \begin{aligned}
 p_{j,\ell}(S)&=p_\theta\!\left(a_{j,\ell}\mid q_j,Z_I[S],a_{j,<\ell}\right),\\
 f_j(S)&=\exp\!\left(\frac{1}{L_j}\sum_{\ell=1}^{L_j}\log p_{j,\ell}(S)\right)
 =\left(\prod_{\ell=1}^{L_j}p_{j,\ell}(S)\right)^{1/L_j}.
 \end{aligned}
 \label{eq:answer-score}
\end{equation}
Thus $f_j$ is the length-normalized geometric mean of the reference-token probabilities, not the raw sequence probability or the arithmetic mean of token probabilities. Length normalization removes the direct multiplicative penalty for longer answers, but does not turn the score into the probability that a freely generated answer is correct. Equation~(\ref{eq:empirical-objective}) averages these scores equally across QA pairs. All candidates are scored with the same reference answer and preceding reference-answer tokens for a given QA pair; their retained visual context differs.
The auxiliary QA set approximates the breadth of questions, while the reference-answer score provides a fine-grained measure of subset quality. Rather than autoregressively generating an answer and reducing the result to a binary correct/incorrect outcome, we retain the teacher-forced probability assigned to the reference answer. This soft score distinguishes subsets even when they yield the same discrete answer and therefore measures changes in image understanding at a finer resolution. It is also substantially faster than decoding because all reference-answer positions are evaluated in parallel in a single causal pass.

\subsection{Optimization with QA mini-batch greedy}
\label{sec:optimization}

\paragraph{Full greedy and its cost.}
Exhaustive search over $\binom{n}{b}$ subsets is infeasible. Starting with $S_0=\varnothing$, forward greedy adds the remaining token with the largest marginal gain:
\begin{equation}
 \Delta_j(v\mid S)=f_j(S\cup\{v\})-f_j(S),\qquad
 v_t\in\operatorname*{arg\,max}_{v\in V_I\setminus S_t}
 \frac{1}{M}\sum_{j=1}^{M}\Delta_j(v\mid S_t),\qquad
 S_{t+1}=S_t\cup\{v_t\}.
 \label{eq:full-greedy}
\end{equation}
Searching to $b_{\max}$ requires $\mathcal{O}(M b_{\max} n)$ QA--candidate scoring evaluations: the number of steps times the remaining candidates times the number of QA pairs.

\paragraph{Sampling candidates or objective components.}
Token-level stochastic greedy randomly samples a fraction of the remaining visual-token candidates at each step and evaluates only those candidates on the complete auxiliary-QA objective. It reduces computation by considering fewer token choices, following the candidate-sampling idea of stochastic greedy \citep{mirzasoleiman2015lazier}.

QA mini-batch greedy instead samples a QA index set $B_t\subseteq\{1,\ldots,M\}$, uniformly without replacement, of size $m=\max\{1,\lceil\alpha M\rceil\}$ for QA fraction $\alpha\in(0,1]$. We resample $B_t$ at every step and use the same mini-batch for all remaining candidates. The update uses
\begin{equation}
 \widehat\Delta_t(v\mid S_t)=\frac{1}{m}\sum_{j\in B_t}\Delta_j(v\mid S_t),
 \qquad
 v_t\in\operatorname*{arg\,max}_{v\in V_I\setminus S_t}\widehat\Delta_t(v\mid S_t).
 \label{eq:qa-greedy}
\end{equation}
For a fixed current subset and remaining token, fresh uniform QA sampling gives
\begin{equation}
 \mathbb{E}_{B_t}\!\left[\widehat\Delta_t(v\mid S_t)\mid S_t\right]
 =\widehat F_I(S_t\cup\{v\})-\widehat F_I(S_t).
 \label{eq:unbiased-marginal}
\end{equation}
For a fixed current subset and candidate token, averaging this estimate over repeated random QA mini-batches recovers the marginal gain computed on the complete generated-QA pool. This statement applies only to that finite pool: it does not imply that the pool represents every possible question, nor that maximizing noisy estimates selects the same token as full greedy. The QA batch size therefore controls both scoring cost and variability in the trajectory. Setting $m=M$ recovers full greedy, while $m=1$ uses one QA pair per step. The number of QA--candidate scoring evaluations decreases linearly with the QA sampling fraction.

\begin{wrapfigure}{r}{0.50\textwidth}
\begin{minipage}{\linewidth}
\begin{algorithm}[H]
\caption{QA mini-batch greedy.}
\label{alg:qa-greedy}
\small
\begin{algorithmic}[1]
\Require $V_I$, scores $\{f_j\}_{j=1}^{M}$, $b_{\max}$, $m$
\State $S_0\gets\varnothing$
\For{$t=0,\ldots,b_{\max}-1$}
  \State Sample a uniform $m$-subset $B_t$ of QAs
  \ForAll{$v\in V_I\setminus S_t$}
    \State $g_t(v)\gets\frac{1}{m}\sum_{j\in B_t}f_j(S_t\cup\{v\})$
  \EndFor
  \State $v_t\gets\arg\max_{v\in V_I\setminus S_t}g_t(v)$
  \State $S_{t+1}\gets S_t\cup\{v_t\}$
\EndFor
\State \Return $(S_0,\ldots,S_{b_{\max}})$
\end{algorithmic}
\end{algorithm}
\end{minipage}
\end{wrapfigure}
\noindent\textbf{One search, multiple budgets.}\enspace

Algorithm~\ref{alg:qa-greedy} returns one nested trajectory whose prefix $S_b$ serves each requested budget and all evaluation questions about the image. One run therefore yields every budget up to $b_{\max}$, amortizing search across the entire budget curve. Because all budgets are prefixes of the same trajectory, comparisons across keeping ratios do not benefit from independent search restarts. QA mini-batching perturbs the estimated gains and can therefore produce a different trajectory from candidate sampling or full greedy. Because one token is added at every step, the full auxiliary objective need not improve monotonically; we treat the resulting subsets as achievable operating points rather than certified optima.

\paragraph{Efficient candidate scoring.}
The greedy structure admits efficient batching: candidates at the same step have equal visual-token length and require no sequence padding. Their fixed system-and-image prefixes also allow candidate-specific KV states to be reused across sampled QA suffixes. Teacher forcing scores all reference-answer positions in parallel without autoregressive decoding, while cached visual embeddings avoid repeated vision encoding. Together, these optimizations preserve exhaustive candidate coverage while reducing repeated computation within each greedy step. Unlike token-candidate sampling, QA-MBG spends the saved computation on fewer objective components while still comparing every remaining token. Appendix~\ref{app:search-execution} gives implementation and runtime details.

\section{Experiments}
\label{sec:experiments}

\subsection{Experimental setup}
\label{sec:experimental-setup}

\paragraph{Evaluation sets.}
We use two cohorts from nine benchmarks: AI2D \citep{kembhavi2016ai2d}, ChartQA \citep{masry2022chartqa}, DocVQA \citep{mathew2021docvqa}, InfographicVQA \citep{mathew2022infographicvqa}, MME \citep{fu2023mme}, MMStar \citep{chen2024mmstar}, RealWorldQA \citep{xai2024realworldqa}, TextVQA \citep{singh2019textvqa}, and V$^*$Bench \citep{wu2024vstar}. The \emph{main set} contains $100$ images per benchmark ($9\times100$: $900$ images, $2{,}226$ actual QAs); the \emph{small set} contains $20$ per benchmark ($9\times20$: $180$ images, $392$ actual QAs). Actual-QA evaluation uses all associated benchmark questions. These annotations are separate from the generated auxiliary QA pool, although their semantic content can overlap. Long-form evaluation uses two description prompts per image.

\paragraph{Models.}
We evaluate Qwen3-VL-4B-Instruct \citep{bai2025qwen3vl} and Gemma~4 E4B-it \citep{gemmateam2026gemma4}, keeping their parameters frozen. Our search selects visual tokens after visual encoding and projection, before language-model processing. Each selected subset is shared across evaluation questions about the same image. We use an image-processor cap of $1$M pixels for Qwen3-VL and a maximum of $1{,}120$ visual soft tokens for Gemma~4. Architecture-specific integration and baseline adaptations are detailed in Appendix~\ref{app:implementations}.

\paragraph{Auxiliary QA generation.}
We use GPT-5.6-SOL to generate image-grounded auxiliary question--answer pairs; the generation protocol, prompts, validation, and provenance are documented in Appendix~\ref{app:qa-generation}. Because generated auxiliary QAs can semantically overlap with actual QAs, we compare the original search with a search after removing overlapping auxiliary QAs. We also evaluate dataset-agnostic generation and repeat QA generation with Opus~5 to test sensitivity to the generation protocol and frontier model. The resulting $30{,}273$ auxiliary QAs average $33.64$ per image. A selected trajectory is held fixed during actual-QA and long-form evaluation, without re-searching for each prompt.

\paragraph{Search and baselines.}
The main search configuration uses QA mini-batch greedy with QA sampling fraction $\alpha=0.2$, evaluating all remaining token candidates at each step. It optimizes the length-normalized geometric-mean answer-probability objective in Eq.~(\ref{eq:answer-score}), averaged over the sampled auxiliary QA pairs. We first interpret headroom against question-independent post-vision subset selectors, including Uniform, FLoC, and DivPrune, which also retain a shared subset of unchanged visual embeddings. The broader comparison includes feature merging, query-aware selection, and decoder-internal pruning. These methods can access different selection signals, change retained features, or process all tokens in early decoder layers. Prior pruning methods do not share a common model backbone, and LLaVA---widely used in earlier work---is an older architecture that is not representative of the contemporary VLMs evaluated here. We therefore implement architecture-specific adaptations for Qwen3-VL and Gemma~4; Appendix~\ref{app:implementations} details their pruning locations and schedules. Full-token inference provides an uncompressed reference.

\paragraph{Choice of QA mini-batch fraction.}
The QA mini-batch fraction $\alpha$ is the only search hyperparameter of our method. We set $\alpha=0.2$ empirically using the small-set experiments; Section~\ref{sec:experimental-analysis} describes the selection analysis.

\paragraph{Budgets and reporting.}
We report visual-token keeping ratios up to $0.30$. Subsets at different budgets are prefixes of one nested search trajectory for each image. The top row of Figure~\ref{fig:main-native-results} expresses native benchmark performance relative to the corresponding full-token model, whose reference level is $100\%$. The searched trajectories provide denser budget coverage than the fixed baseline operating points. Comparisons at a given retained-token budget measure achievable quality; equal final token counts do not imply equal computational cost across pre-LLM and progressive decoder-internal methods. Search execution and measured runtime are reported in Appendix~\ref{app:search-execution}.

\paragraph{Long-form evaluation.}
We reuse the selected image representation for two general image-description prompts (P1/P2), without optimizing a new subset for either prompt. GPT-5.6-Luna evaluates the generated descriptions against the original image, with candidate answers anonymized and shuffled. Figure~\ref{fig:main-native-results}(c,d) reports mean overall image-grounded quality on a $0$--$5$ scale over the evaluated image--prompt groups. Full-token answers are comparison candidates rather than ground-truth references; their ratings can vary with the other candidates in each judging group. The fixed-subset baselines and the generation and judging protocol are described in Appendices~\ref{app:decoder-baselines} and~\ref{app:longform}.

\subsection{Main results}
\label{sec:experimental-results}

\begin{figure}[!t]
 \centering
 \includegraphics[width=\linewidth]{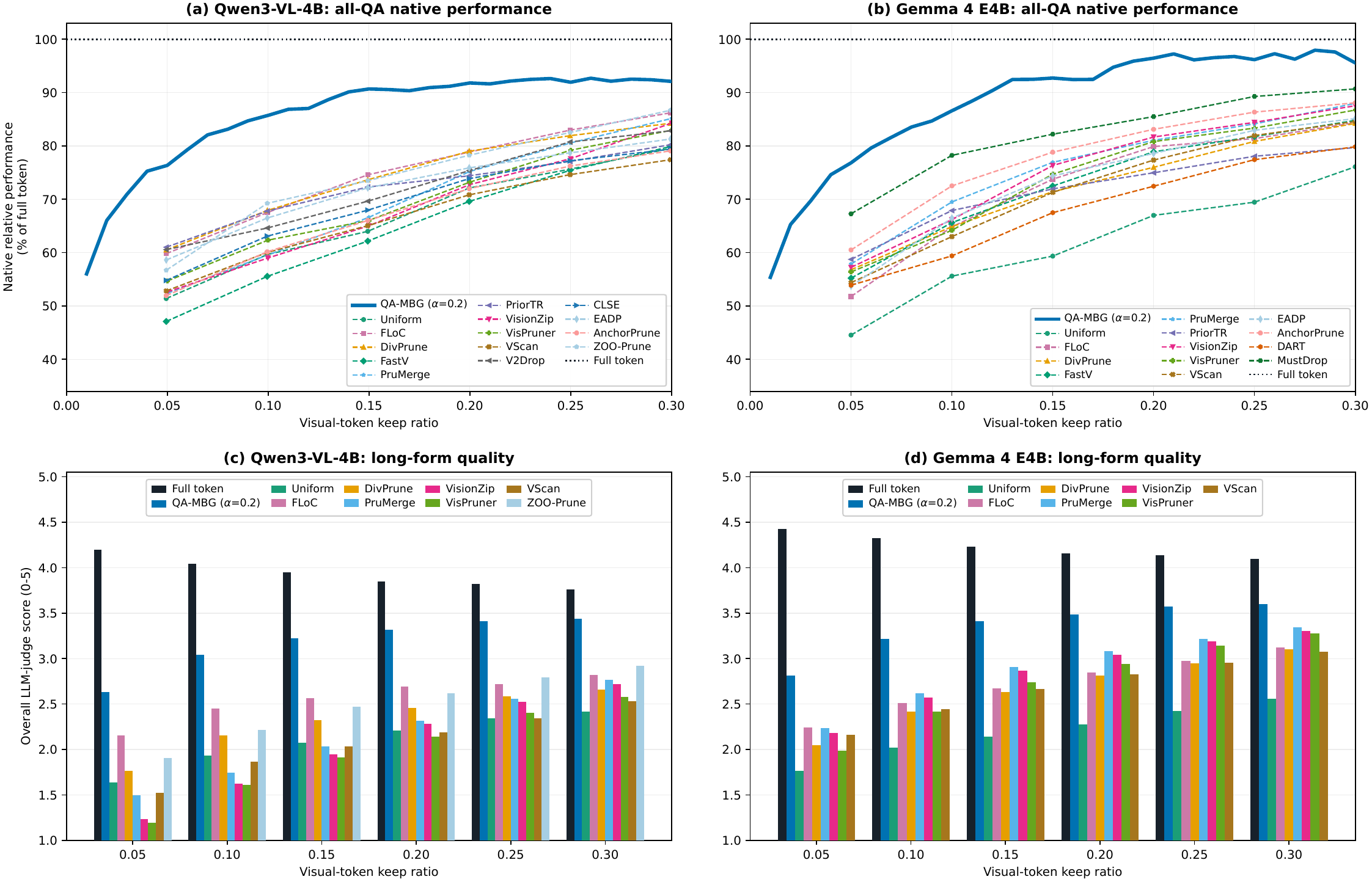}
 \caption{Actual-QA and long-form performance on the main set.}
 \label{fig:main-native-results}
\end{figure}

\paragraph{Achievable headroom across two model families.}
Figure~\ref{fig:main-native-results}(a,b) shows that QA-guided subset search achieves higher aggregate native benchmark performance than the plotted baselines at comparable token budgets for both Qwen3-VL and Gemma~4. The separation is particularly pronounced at low and moderate keeping ratios and remains visible toward $0.30$. Thus the performance lost by existing selectors is not determined by the token budget alone: stronger retained subsets are available under the same model and budget constraints.

\paragraph{Performance as the budget increases.}
The explored curves rise rapidly at small budgets and then approach a plateau closer to the full-token reference. The baselines also improve with additional tokens, narrowing the gap at larger budgets, but do not close it over the displayed range. Notably, this headroom also persists against query-aware pruning methods that select tokens using the evaluation question itself, even though our search produces one shared subset for all questions about an image. Observing this pattern in both model families supports the usefulness of shared-subset exploration beyond a single VLM architecture. The gap to direct shared-subset selectors demonstrates achievable quality beyond those selection rules under the auxiliary-supervision protocol. Comparisons with merging and decoder-internal adapters broaden the empirical context, while retaining the differences in representation and computation described above.

\paragraph{Transfer to long-form descriptions.}
Figure~\ref{fig:main-native-results}(c,d) shows that the same searched representations also improve long-form quality over the plotted compression baselines at all six evaluated budgets for both models. These prompts request broad descriptions rather than the short answers used by the search objective. The advantage therefore extends beyond the native short-answer evaluation. For Qwen3-VL, QA-MBG retains a clear margin over the strongest plotted baseline throughout the range, while its gap to full-token descriptions narrows as the budget increases. Averaged over the six budgets, its paired advantage over FLoC, the baseline with the highest six-budget mean, is $0.612$ judge-score points ($95\%$ CI $[0.578,0.645]$). This evaluates whether the selected image information remains useful under new prompts and a different answer format on the same images. The judge assesses each answer against the image rather than treating the full-token response as ground truth.

\paragraph{Paired uncertainty analysis.}
A benchmark-stratified, image-level paired bootstrap with $10{,}000$ replicates yields positive lower endpoints for all $252$ pointwise $95\%$ confidence intervals: $84$ Qwen and $78$ Gemma actual-QA comparisons, and $48$ Qwen and $42$ Gemma long-form comparisons. Against the baseline with the highest observed six-budget mean in each model/task, mean gains are $10.12$ and $5.10$ native-score percentage points for Qwen and Gemma actual QA, and $0.61$ and $0.45$ judge-score points for their long-form descriptions, respectively. These are individual comparison intervals, not a simultaneous confidence guarantee across all comparisons. Actual-QA bootstrap estimates use image-balanced native scores, rather than the full-token-normalized metric in Figure~\ref{fig:main-native-results}. Appendix~\ref{app:paired-bootstrap} details the resampling protocol, comparison summaries, and scope of inference.

\subsection{Analysis}
\label{sec:experimental-analysis}

We examine the choice of search strategy, its effect on auxiliary-objective optimization, sensitivity to benchmark conditioning and direct QA overlap, the size of the supervision pool, and the auxiliary-QA generator. These analyses complement the paired uncertainty estimates and long-form evaluation above.

\begin{figure}[!t]
\centering
\includegraphics[width=\linewidth]{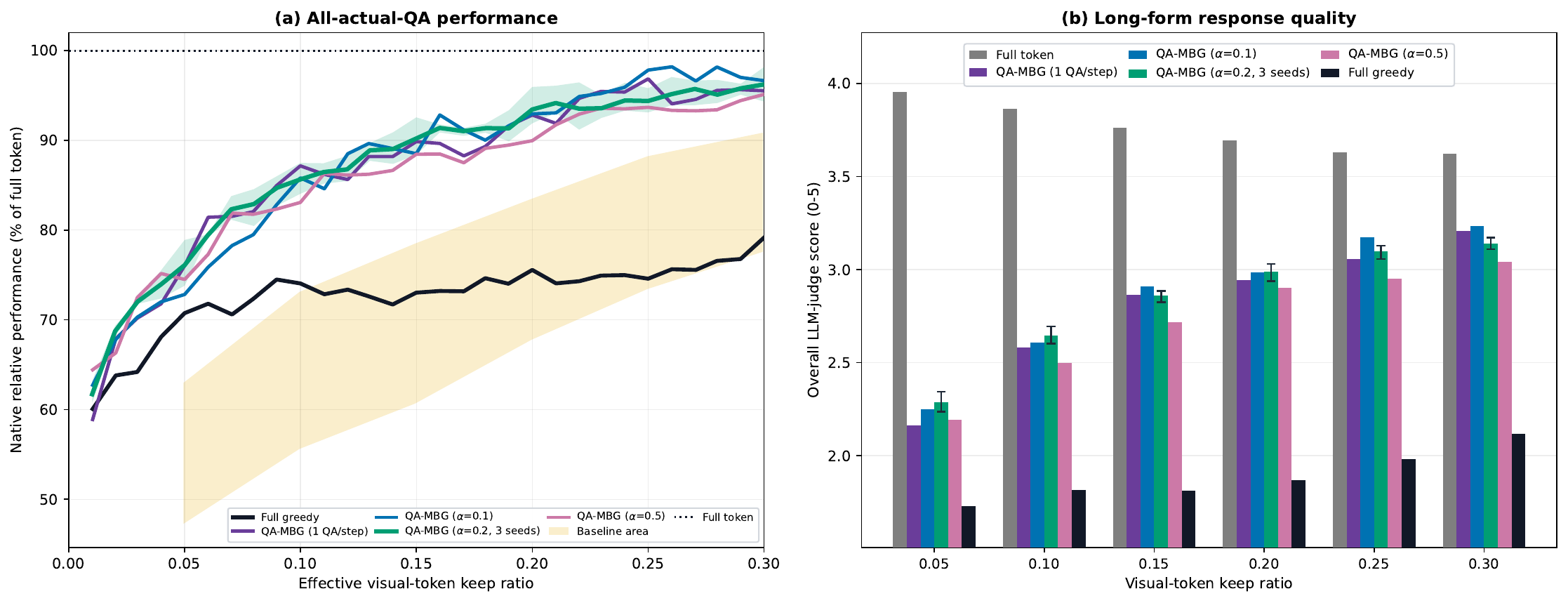}
\caption{QA mini-batch greedy versus full greedy on the small set.}
\label{fig:greedy-comparison}
\end{figure}

\paragraph{QA mini-batching versus full greedy.}
Figure~\ref{fig:greedy-comparison} compares QA batch sizes on the small set. Green shading and error bars show the $\alpha=0.2$ three-seed min--max ranges, not confidence intervals; the yellow envelope spans baseline methods rather than repeated runs. Other settings are single runs (Appendix~\ref{app:search-seeds}). At keep ratio $0.20$, $\alpha=0.2$ achieves mean relative actual-QA performance of $93.44\%$ across three seeds (range $91.90$--$95.95\%$), versus $75.56\%$ for full greedy and $89.96\%$ for the reported $\alpha=0.5$ run. Long-form quality is $2.989$ (range $2.939$--$3.031$), versus $1.869$ for full greedy and $2.903$ for $\alpha=0.5$. All three $\alpha=0.2$ seeds exceed full greedy at the six standard budgets in both evaluations, so the advantage is not confined to the original sampling trajectory. The larger fraction $0.5$ yields lower long-form quality than every $0.2$ seed at all six budgets despite evaluating more QA components per candidate. This supports $0.2$ as a practical setting, while $0.1$ and one QA per step remain competitive. The reversal---QA-MBG succeeding where full greedy underperforms---is particularly striking. We hypothesize that full greedy repeatedly fits the same finite generated-QA objective, whereas resampling QA components changes the objective estimate at every step and acts as a regularizer against overfitting to pool-specific artifacts. The following QA-sampling-dynamics analysis probes this hypothesis. Appendix~\ref{app:greedy-comparison} gives complete results, including the supplementary token-level control and seed protocol.

\paragraph{QA sampling dynamics.}
On the small set, QA sampling improves the lower tail of generated-QA probabilities compared with full greedy. At keeping ratio $0.30$, $\alpha=0.5$ achieves the highest overall mean ($0.7203$ versus $0.6790$ for full greedy) and bottom-$20\%$ mean ($0.3566$ versus $0.2756$) among the tested fractions. The lower tail is recomputed at each step; it is not a fixed group of initially difficult questions. At $\alpha=0.2$, the overall mean is similar to full greedy while the lower tail improves. Step-level diagnostics show focused gains on sampled QAs alongside temporary regressions on unsampled QAs, consistent with resampling redistributing improvements across the search trajectory. These diagnostics characterize optimization of the generated-QA objective; they do not imply a monotonic benefit from smaller fractions. Downstream transfer is assessed separately using actual QAs and long-form descriptions. Appendix~\ref{app:qa-sampling-mechanism} provides definitions, trajectories, and numerical summaries.

\paragraph{Controlling the QA distribution.}
When the target question distribution is known, as in benchmark-specific QA generation, auxiliary QAs can align with the actual-QA evaluation and introduce potential question-level leakage. We use two controls to separate this effect from headroom attributable to token selection. First, dataset-agnostic generation withholds benchmark names, task profiles, dataset-specific instructions, and same-benchmark examples while retaining the image-grounding and diversity requirements. At keep ratio $0.20$, it achieves $92.79\%$ full-token-relative performance, compared with $96.18\%$ for the original setting and $83.47\%$ for the strongest baseline. Its margin over the strongest baseline remains positive at all six budgets, showing that benchmark information helps but is not required for the observed headroom.

For the overlap control, after generating the auxiliary QAs, we use a text-only GPT-5.6-Luna judge to compare generated and actual QA pairs and filter generated QAs that ask about semantically equivalent visual facts. We then re-run subset search with the filtered generated-QA pool. At keep ratio $0.20$, filtered search reaches $91.36\%$ full-token-relative performance, compared with $96.18\%$ without filtering and $83.47\%$ for the strongest baseline. The filtered trajectory remains above the strongest baseline at all six budgets. Actual-QA annotations are used only to define which generated QAs are filtered; they never score candidate subsets. Because filtering also removes potentially useful visual supervision, the performance drop does not measure leakage alone. Together, the controls show that task conditioning and semantic overlap strengthen, but do not explain, the measured headroom. Appendices~\ref{app:qa-removal} and~\ref{app:dataset-agnostic} provide protocols and full results.

\paragraph{Number of generated QAs.}
On the small set, we vary the search pool using nested $25\%$, $50\%$, $75\%$, and $100\%$ subsets. At keep ratio $0.20$, relative performance increases from $86.33\%$ with one quarter of the pool to $96.18\%$ with the complete pool; using three quarters reaches $93.87\%$. Larger pools generally improve aggregate performance, with benchmark-dependent and occasionally non-monotonic effects. This measures sensitivity to the amount of auxiliary supervision. The per-step QA fraction remains $0.2$, so increasing the pool also increases the sampled QA count and search compute; it does not isolate diversity at fixed compute. Appendix~\ref{app:qa-count-ablation} provides the construction, exact counts, and results.

\paragraph{Auxiliary-QA generator.}
Replacing GPT-5.6-SOL with Opus~5 on the small set produces a closely matched actual-QA curve: the six-budget mean relative performance is $93.14\%$ versus $91.69\%$, with a mean absolute pointwise difference of $2.30$ points. Opus-generated subsets also exceed the best plotted long-form baseline at each budget by $0.56$ judge-score points on average ($3.21$ versus $2.65$). Appendix~\ref{app:generator-robustness} gives the per-budget results.

\paragraph{Qualitative examples.}
Appendix~\ref{app:qualitative} presents one selected favorable example from each benchmark, showing retained visual tokens, actual-QA responses, and long-form descriptions for the same image. These cases illustrate preserved evidence and differences in description quality; aggregate comparisons use the full evaluation cohorts above.

\section{Conclusion}
\label{sec:conclusion}

We ask how much performance visual token pruning can preserve when offline search cost is relaxed. We formulate general-understanding pruning as selecting one image-specific subset shared across questions, approximate its utility with frontier-generated auxiliary QAs, and optimize it with QA mini-batch greedy. Across Qwen3-VL and Gemma~4, the resulting trajectories reveal substantial headroom over evaluated pruning baselines on both actual QA and long-form descriptions, while narrowing the gap to full-token inference. On the small set, QA mini-batching also outperforms full greedy with fewer QA evaluations; its step-wise dynamics are consistent with a regularizing effect from resampling objective components. The headroom persists when benchmark information is withheld, semantically overlapping generated QAs are filtered, and the auxiliary-QA generator is changed. It also remains with smaller generated-QA pools, although performance generally improves as more auxiliary QAs are used.

These results establish a strong empirical frontier across the evaluated models and supervision settings. The consistent gaps show that a substantial portion of current pruning loss is recoverable selector error rather than inevitable information loss from the token budget. Visual token pruning can therefore perform substantially better than current heuristics, and the searched trajectories provide concrete targets for learning or designing the next generation of efficient selectors.

\par
\endgroup

\clearpage
\section*{AI Use Statement}
Frontier models are integral to the method and evaluation: GPT-5.6-SOL and Opus~5 generate auxiliary QAs, while GPT-5.6-Luna evaluates long-form outputs and semantic QA overlap. AI assistants also supported code development, language editing, and figure and table formatting. The authors designed the study and verified the implementations, analyses, claims, citations, and final manuscript.

\section*{Ethics Statement}
We use public research benchmarks and pretrained models, and collect no new human-subject or personally identifying data. Generated QAs and LLM judgments may inherit model biases or errors; we therefore report aggregate results and targeted controls rather than treating them as ground truth. The offline search costs approximately $0.5$ H100 GPU-hours per image and is intended to diagnose achievable headroom, not as a deployment-time algorithm.

\section*{Reproducibility Statement}
The main text and appendix document the evaluated cohorts and models, generation and judging prompts, scoring objective, search algorithm and hyperparameter, baseline adaptations, random seeds, runtime measurements, and bootstrap procedure. Exact replication depends on access to the named hosted frontier models; we report their identities, prompts, and provenance to make this dependency explicit.

\bibliography{iclr2027_conference}
\bibliographystyle{iclr2027_conference}

\clearpage
\appendix
\addtocontents{toc}{\protect\setcounter{tocdepth}{1}}
\phantomsection
\pdfbookmark[0]{Appendix}{appendix-overview}
\begin{center}
{\LARGE\bfseries Appendix\par}\vspace{0.7em}
{\large How Much Can We Prune?\par Exploring the Performance Limits of Visual Token Pruning\par}
\end{center}
\vspace{1em}
\renewcommand{\contentsname}{Appendix Contents}
{\small\tableofcontents}
\clearpage

\setlength{\parskip}{3pt plus 0.5pt minus 0.5pt}
\setlength{\textfloatsep}{10pt plus 2pt minus 2pt}
\setlength{\floatsep}{8pt plus 2pt minus 2pt}
\setlength{\intextsep}{8pt plus 2pt minus 2pt}
\renewcommand{\topfraction}{0.9}
\renewcommand{\bottomfraction}{0.8}
\renewcommand{\textfraction}{0.08}
\renewcommand{\floatpagefraction}{0.8}
\setcounter{topnumber}{3}
\setcounter{bottomnumber}{3}
\setcounter{totalnumber}{5}
\raggedbottom
\FloatBarrier
\section{Baseline Implementation Details}
\label{app:implementations}

We evaluate architecture-specific adapters for Qwen3-VL-4B and Gemma~4 E4B. The descriptions below distinguish unchanged-token selection, feature merging, and decoder-internal pruning. All adapters use frozen model weights; adaptations to architectures different from those in the source papers are stated explicitly.

\paragraph{Comparison overview.}
Table~\ref{tab:app-baseline-overview} classifies the evaluated adapters. Post-ViT denotes reduction after visual encoding and before the first LLM layer; Inside LLM denotes decoder-internal reduction. Query-aware methods explicitly use the evaluation question. Progressive methods prune at multiple decoder depths. Q and G denote Qwen3-VL and Gemma~4, respectively.

\begin{table}[htbp]
\centering\footnotesize
\setlength{\tabcolsep}{4pt}
\renewcommand{\arraystretch}{1.16}
\caption{Taxonomy of the evaluated baseline adapters.}
\label{tab:app-baseline-overview}
\begin{tabular}{lp{0.42\linewidth}ccc}
\hline
Pruning stage & Method / reference & \shortstack{Query-\\aware} & Progressive & Models\\\hline
\multirow{10}{*}{Post-ViT} & Uniform (position-only control) & No & No & Q, G\\
 & FLoC \citep{cho2026floc} & No & No & Q, G\\
 & DivPrune \citep{alvar2025divprune} & No & No & Q, G\\
 & EADP \citep{wang2026eadp} & Yes & No & Q, G\\
 & AnchorPrune \citep{oh2026anchorprune} & Yes & No & Q, G\\
 & ZOO-Prune \citep{kim2026zooprune} & No & No & Q\\
 & VisionZip \citep{yang2025visionzip} & No & No & Q, G\\
 & PruMerge \citep{shang2024prumerge} & No & No & Q, G\\
 & VisPruner \citep{zhang2025vispruner} & No & No & Q, G\\
 & VScan$^{a}$ \citep{zhang2025vscan} & No & No & Q, G\\\hline
\multirow{6}{*}{Inside LLM} & FastV \citep{chen2024fastv} & Yes & No & Q, G\\
 & PriorTR \citep{chen2026priortr} & Yes & No & Q, G\\
 & V2Drop$^{b}$ \citep{chen2025v2drop} & No direct & 3 stages & Q\\
 & CLSE \citep{chen2026clse} & Yes & 3 stages & Q\\
 & DART \citep{wen2025dart} & Yes & No & G\\
 & MustDrop$^{c}$ \citep{liu2024mustdrop} & Yes & No & G\\\hline
\end{tabular}
\par\vspace{4pt}
\begin{minipage}{\linewidth}\footnotesize
$^{a}$ Our VScan adapter uses vision-side local/global selection and merging; it does not include the original method's additional LLM-stage reduction.
$^{b}$ V2Drop ranks visual hidden-state changes without an explicit question-attention or question-embedding term. It is evaluated through decoder prefill, not through the fixed pre-LLM subset-transfer protocol.
$^{c}$ Our Gemma MustDrop adapter uses one decoder pruning stage, rather than the original multi-stage pipeline.
\end{minipage}
\end{table}

\subsection{Models, budgets, and integration}
\label{app:model-integration}

\paragraph{Qwen3-VL.}
We use \texttt{Qwen/Qwen3-VL-4B-Instruct} \citep{bai2025qwen3vl} with an image-processor cap of $1{,}003{,}520$ pixels. The visual-token count $n$ is measured after the visual merger and therefore varies across images. Post-merger methods select embeddings immediately before their insertion into the language-model sequence. Retained tokens are restored to their original order, and the sequence, masks, and multimodal position information are compacted consistently. Text tokens are retained. The default attention backend is FlashAttention~2; attention-exposing paths, including FastV and PriorTR, use SDPA. Prompt, image-processing, decoding, and checkpoint settings are held constant across methods within a comparison, apart from the attention-backend changes required by an adapter.

\paragraph{Gemma~4.}
We use \texttt{google/gemma-4-E4B-it} \citep{gemmateam2026gemma4} with \texttt{max\_soft\_tokens=1120}. Post-vision selection acts on projected soft tokens after the vision tower, pooler, and \texttt{embed\_vision}; it neither resizes the image nor selects raw vision patches. Retained indices preserve the original spatial/sequence order. The same positions compact the input embeddings, attention mask, multimodal token-type IDs, and Gemma per-layer inputs, while retaining text tokens. The resulting prefix length determines the language-model context and cache length. Vision-attention selectors compute received-attention mass using native Q/K projections and multidimensional rotary positions, average over queries and heads, and reduce to the soft-token pooler grid. 

\paragraph{Integer budgets and operating points.}
All comparisons use logged retained-token counts and effective ratios $|S|/n$, rather than assuming identical $n$ across images. The Qwen3-VL native-performance suite evaluates QA-MBG with QA fraction $0.2$ at requested ratios $0.01,0.02,\ldots,0.30$, and fixed baselines at $0.05,0.10,\ldots,0.30$. Each adapter receives a per-image integer target. Gemma's standard post-vision evaluation uses $\max\{1,\lfloor\rho n\rfloor\}$, whereas its long-form comparison runner uses $\max\{1,\operatorname{round}(\rho n)\}$ for both QA-MBG and fixed baselines to match batch shapes. These integer conventions can produce slightly different effective ratios. Saved QA-MBG trajectories are reused at the requested counts without search on the evaluation question.

\subsection{Post-vision selectors}
\label{app:postvision}

\paragraph{Uniform and facility location.}
Uniform selects evenly spaced indices as a question-independent control. FLoC \citep{cho2026floc} was proposed for efficient visual-token compression in long-video understanding. We adapt its facility-location selection to the post-vision embeddings of a single image. Given cosine similarities $s_{uv}$ between projected visual embeddings, its selection objective is
\begin{equation}
 G(S)=\sum_{u=1}^{n}\max_{v\in S}s_{uv}.
 \label{eq:app-floc}
\end{equation}
Qwen3-VL uses greedy exact-budget selection and Gemma uses lazy-greedy selection. Both retain the selected embeddings without merging discarded features. DivPrune \citep{alvar2025divprune} selects for visual diversity using embedding distances. The Gemma implementation uses farthest-first cosine-distance selection, initialized with the token having the largest nearest-neighbor distance; subsequent additions maximize minimum distance to the selected set.

\paragraph{Query-aware selectors.}
EADP \citep{wang2026eadp} combines global and local visual-to-instruction relevance with weighted facility location. Local relevance is entropy-filtered and spatially smoothed when a valid grid is available. The frozen Qwen3-VL settings use $\alpha=0.5$, $\beta=2.0$, entropy keep ratio $0.2$, entropy temperature $100.0$, and local temperature $0.01$, without external query enrichment. The Gemma adapter uses $\alpha=0.5$ and $\beta=1.0$ in projected soft-token space. These are EADP parameters, distinct from the QA sampling fraction in Section~\ref{sec:optimization}.

AnchorPrune \citep{oh2026anchorprune} ranks anchors by maximum visual-to-question cosine relevance. Both adapters use $k_{\min}=5$, novelty threshold $\tau=0.2$, patience $3$, and maximum-anchor ratio $0.5$, then fill the remaining budget using diversity and importance. Gemma uses final-ViT received-attention mass for the second-stage importance signal. These are adaptations of the Qwen2.5-VL-oriented method. Both EADP and AnchorPrune may select different subsets for different evaluation prompts.

\paragraph{Importance, context, and feature merging.}
Table~\ref{tab:app-vision-baselines} summarizes our adaptations of VisionZip \citep{yang2025visionzip}, PruMerge \citep{shang2024prumerge}, VisPruner \citep{zhang2025vispruner}, and VScan \citep{zhang2025vscan}. All are question-independent in this setup. A method that merges discarded features retains a compressed representation, rather than a literal subset of unchanged embeddings; this distinction is preserved when interpreting the comparisons.

\begin{table}[htbp]
\centering\small
\caption{Vision-side baseline implementations. Layer numbers follow the adapter configuration.}
\label{tab:app-vision-baselines}
\begin{tabular}{p{0.13\linewidth}p{0.38\linewidth}p{0.38\linewidth}}
\hline
Method & Qwen3-VL adaptation & Gemma~4 adaptation\\\hline
VisionZip & Final-vision-attention dominant tokens plus contextual representatives; discarded context is merged. & Final-ViT received attention; $90\%$ dominant budget by default, uniform contextual sampling, and cosine-nearest cluster averaging.\\[4pt]
PruMerge & Final-vision-attention representatives; nearest final-ViT keys assign discarded tokens to cluster means. Image and DeepStack features are updated consistently. & Mean received attention substitutes for the absent ViT CLS token. Top-$k$ representatives receive cosine-nearest cluster means.\\[4pt]
VisPruner & $75\%$ importance allocation, with contextual reconstruction and redundancy reduction for the remaining budget. & $75\%$ received-attention allocation; even/odd residual-token matching removes redundancy. Selected features are unchanged.\\[4pt]
VScan & Local importance from vision layer~0 with $4\!\times\!4$ windows, plus final-layer global importance; exact-budget filling, de-duplication, and feature merging. & Half local layer-0 coverage and half final-layer global selection (extra token to global for odd budgets); cosine-nearest cluster averaging.\\\hline
\end{tabular}
\end{table}

\paragraph{ZOO-Prune.}
ZOO-Prune \citep{kim2026zooprune} is evaluated on Qwen3-VL using merger-sensitivity scores with sequential diversity-aware selection and is an adaptation of a LLaVA-targeted release.

\subsection{Decoder-internal adapters and comparison scope}
\label{app:decoder-baselines}

Decoder-internal methods begin prefill with all visual tokens and compact the sequence at one or more language-model layers. Their equal final token counts do not imply equal prefill cost to pre-LLM selection. Table~\ref{tab:app-decoder-baselines} records the configured layers and ranking rules. Qwen attention-score paths and the Gemma decoder-internal adapters require batch size one.

\begin{table}[htbp]
\centering\small
\caption{Decoder-internal implementations and configured pruning layers. A dash denotes a method not included in that model's main comparison. References are given in Table~\ref{tab:app-baseline-overview}.}
\label{tab:app-decoder-baselines}
\begin{tabular}{p{0.13\linewidth}p{0.11\linewidth}p{0.13\linewidth}p{0.48\linewidth}}
\hline
Method & Qwen layers & Gemma layers & Selection rule and adaptation\\\hline
FastV & 2 & 5 & Head-averaged attention from the last valid text token; exact top-$k$ retention.\\[3pt]
PriorTR & 3 & 5 & Question-conditioned visual attention relative to a text-attention prior. Qwen aggregates question tokens and heads without instruction exclusion or feature-norm mixing. Gemma scores $p\log((p+\epsilon)/(q+\epsilon))$ using normalized question and text-anchor attention.\\[3pt]
V2Drop & 3, 16, 22 & --- & L2 feature-change ranking, with calibrated cumulative targets $\rho^{1/3},\rho^{2/3},\rho$.\\[3pt]
CLSE & 1, 10, 19 & --- & Attention/evolution top-$k$ selection with \texttt{clse\_score\_type=clse\_attn} and the same calibrated cumulative targets.\\[3pt]
DART & --- & 6 & Up to four visual and four question pivots selected by L1 norm; retain visual pivots, add cosine-dissimilar tokens, and fill by visual L1 norm.\\[3pt]
MustDrop & --- & 5 & Single-stage mean question-to-visual attention with actual decoder-sequence compaction.\\\hline
\end{tabular}
\end{table}

For Gemma, compaction jointly updates hidden states, per-layer inputs, absolute positions, full/sliding masks, dynamic caches, shared KV state, and retained-position metadata used during generation. Masks are rebuilt from compacted absolute positions to preserve hybrid full/sliding attention. Attention-ranking stages use full-attention layers. For progressive adapters, the cumulative targets are defined relative to the original visual-token count and calibrated to reach the requested final ratio; these are not unchanged source-paper schedules.

\paragraph{Native evaluation versus fixed-subset transfer.}
The main-set actual-QA comparison includes the fourteen Qwen3-VL and thirteen Gemma baselines listed in Table~\ref{tab:app-baseline-overview}. Fixed-subset long-form evaluation uses Uniform, FLoC, DivPrune, VisionZip, PruMerge, VisPruner, and VScan for both models, plus ZOO-Prune for Qwen3-VL. These methods reuse the same selected representation for P1 and P2. Query-aware and decoder-internal methods are excluded from this fixed-representation comparison. The small-set search-strategy study is reported separately in Appendix~\ref{app:greedy-comparison}.

\FloatBarrier
\section{Efficient Execution of Subset Search}
\label{app:search-execution}

The Qwen3-VL implementation combines cached visual inputs, candidate batching, prefix KV reuse across QAs, and teacher-forced scoring. These execution choices are independent of the QA-sampling policy.

\subsection{Caching, batching, and teacher forcing}
\label{app:caching}

\paragraph{Reusable inputs.}
Image processing, tokenization, and pre-LLM tensors are cached for the current image. Candidate construction gathers visual embeddings, masks, multimodal positions, and DeepStack features from these tensors, avoiding repeated vision-encoder and merger execution. This input cache is distinct from the decoder KV cache.

\paragraph{Candidate batching.}
At step $t$, candidates are the subsets $S_t\cup\{v\}$ for remaining tokens $v$. For a fixed QA, every candidate has the same sequence length, so batches of up to $K$ candidates require no padding. The implementation compacts embeddings, masks, positions, and DeepStack features consistently. We use $K=320$ in the tuned runtime measurement and split batches if GPU memory is insufficient.

\paragraph{Teacher-forced objective and selective logits.}
Candidates use the reference-answer score in Eq.~(\ref{eq:answer-score}). Teacher forcing computes all answer positions in one causal pass, and the LM head is evaluated only at required positions. No autoregressive answer-generation loop is used during search; score reductions use FP32.

\paragraph{Prefix KV reuse across QAs.}
For each candidate, the system-and-image prefix through the final \texttt{<|vision\_end|>} marker is shared by all sampled QAs. A batched prefix pass creates an independent KV state for each candidate. Each question/reference-answer suffix reuses that state, restoring the cache to the prefix boundary before the next QA. Prefix identity, positions, and masks are checked, and DeepStack features are consumed only in the prefix pass. Reuse is across QAs for a fixed candidate batch; different subsets, candidate batches, and greedy steps require new prefix states.

\subsection{Scoring volume and forward-call accounting}
Let $c_t$ denote the number of evaluated candidates, $m_t$ the number of sampled QAs, and $h_t=\lceil c_t/K\rceil$. The logical scoring volume is $c_tm_t$ QA--candidate evaluations. Serial execution uses approximately $c_tm_t$ full-sequence forwards. Candidate batching reduces this to $h_tm_t$ batched full-sequence forwards. Prefix reuse instead executes
\begin{equation}
 h_t\ \text{batched prefix forwards}
 \quad+\quad h_tm_t\ \text{batched suffix forwards}.
 \label{eq:app-prefix-calls}
\end{equation}
This can increase the number of forward \emph{calls} while reducing repeated work. If $T_{\mathrm{pre}}$ and $T_{\mathrm{suf},j}$ denote the prefix and question-specific suffix costs at the same batch shape, the corresponding cost model changes from approximately $h_t\sum_j(T_{\mathrm{pre}}+T_{\mathrm{suf},j})$ to $h_t(T_{\mathrm{pre}}+\sum_jT_{\mathrm{suf},j})$. Benefits depend on prefix length, QA count, batch utilization, and memory. Neither the QA sampling fraction nor the candidate batch size alone determines wall-clock speedup.

\subsection{Measured H100 runtime}
\label{app:runtime}

\paragraph{Measurement scope.}
We measure Qwen3-VL-4B-Instruct on one NVIDIA H100 80GB, using PyTorch~2.5.1+cu121, Transformers~4.57.1, and FlashAttention~2. The DocVQA image produces $945$ post-merger tokens and has $16$ auxiliary QAs from the pilot pool. These are single-image measurements, not main-set averages or Gemma timings.

Table~\ref{tab:app-microbenchmark} evaluates $64$ candidate subsets on all $16$ QAs at each ratio. Times sum two measured iterations after one warm-up; all paths share cached pre-LLM inputs. Candidate batching gives a $13.84\times$ aggregate speedup over serial scoring, and prefix reuse adds $3.02\times$ over batching. These are candidate-scoring speedups, not end-to-end search speedups.

\begin{table}[htbp]
\centering\small
\caption{Candidate-scoring microbenchmark on one H100.}
\label{tab:app-microbenchmark}
\begin{tabular}{rrrrrr}
\hline
Keep ratio & Tokens & Serial (s) & Batched (s) & Batch + prefix (s) & Combined speedup\\\hline
0.05 & 47 & 99.393 & 4.423 & 2.072 & $47.96\times$\\
0.10 & 94 & 99.497 & 6.580 & 2.344 & $42.46\times$\\
0.20 & 189 & 100.590 & 10.641 & 2.748 & $36.61\times$\\
\hline
Total & --- & 299.480 & 21.644 & 7.164 & $41.80\times$\\\hline
\end{tabular}
\end{table}

\paragraph{A complete nested trajectory and batch-size tuning.}
With $\alpha=0.2$ (four QAs per step) and all remaining candidates, search makes $284$ additions, reaching effective ratio $284/945=0.30053$. Increasing $K$ from $64$ to $320$ reduces wall time from $1{,}853.42$ to $1{,}671.89$ seconds: \textbf{27.9 minutes on one H100}, a $9.8\%$ reduction. Both runs select the same token sequence. Timing includes model loading, reference scoring, search, and output processing; candidate evaluation accounts for approximately $95\%$ of the tuned run. Prefixes of the trajectory provide the smaller budgets.

\begin{table}[htbp]
\centering\small
\caption{End-to-end runtime and memory for the same image and trajectory.}
\label{tab:app-endtoend}
\begin{tabular}{lrr}\hline
Measurement & Batch 64 & Batch 320\\\hline
Process wall time (s) & 1,853.42 & 1,671.89\\
Candidate-evaluation profile time (s) & 1,777.77 & 1,588.09\\
Peak allocated CUDA memory (GiB) & 18.23 & 56.12\\
\hline
\end{tabular}
\end{table}

Throughput nearly saturates after batch size $256$. We use $320$ as a throughput--memory compromise; this setting depends on image and answer lengths.

\paragraph{Numerical validation.}
In the microbenchmark, serial and batched FlashAttention select the same best candidate at all three ratios, although the maximum per-QA score difference is $0.0633$. Prefix reuse matches the candidate-batched scores in this probe. These checks validate the measured execution paths without assuming bit-exact equivalence across all inputs.

\FloatBarrier
\section{Auxiliary QA Generation}
\label{app:qa-generation}

\subsection{Generation protocol}
GPT-5.6-SOL generates auxiliary QAs from the target image, a benchmark-specific profile, and nine text-only QA examples from other images of that benchmark. Examples follow manifest order, excluding the target image ID. The prompt requests about $32$ concise, image-grounded QAs (preferably $28$--$36$), emphasizing independent facts and peripheral evidence, with at most two substantial paraphrases per fact. Answers normally contain $1$--$12$ words. The exact template is in Appendix~\ref{app:qa-template}.

\paragraph{Image and generation settings.}
Images are converted to RGB JPEG, initially at quality $95$, with compression or downscaling when needed to meet the API payload limit. Requests use JSON output and a $16{,}000$-token cap. Saved responses report temperature $1.0$ and top-$p=0.98$; the client does not override these settings.

\paragraph{Validation and overlap policy.}
We require nonempty questions and answers, normalize whitespace, and remove duplicate normalized questions. Questions longer than $300$ characters and answers longer than $180$ characters or $24$ words are rejected, as are \texttt{Unanswerable} answers for the evaluated benchmarks. The requested count is a soft target.

The main pool is not filtered against actual QAs. Target-image evaluation annotations are absent from the generation prompt and candidate scoring, although they are used separately for overlap diagnostics. Thus prompt exclusion does not imply semantic non-overlap; Appendix~\ref{app:qa-removal} describes the filtering control.

\subsection{QA counts and experimental pools}
\label{app:qa-counts}
The main set contains $30{,}273$ auxiliary QAs for $900$ images, with mean $33.637$, median $35$, and range $2$--$37$ per image. Table~\ref{tab:app-qa-counts} gives the benchmark breakdown. The small-set count, removal, and conditioning controls use the corresponding main-generation pool. The search-strategy/seed sweep and the single-image runtime probe use the earlier pilot pool; their results are compared within each experiment.

\begin{table}[htbp]
\centering\small
\caption{Auxiliary-QA counts on the main set (100 images per benchmark).}
\label{tab:app-qa-counts}
\begin{tabular}{lrrr}\hline
Benchmark & QA pairs & Mean per image & Range\\\hline
AI2D & 3,318 & 33.18 & 30--36\\
ChartQA & 3,504 & 35.04 & 31--36\\
DocVQA & 3,466 & 34.66 & 28--36\\
InfographicVQA & 3,560 & 35.60 & 32--37\\
MME & 3,376 & 33.76 & 30--36\\
MMStar & 3,418 & 34.18 & 29--36\\
RealWorldQA & 3,461 & 34.61 & 30--36\\
TextVQA & 2,628 & 26.28 & 2--36\\
V$^*$Bench & 3,542 & 35.42 & 32--37\\\hline
Total & 30,273 & 33.637 & 2--37\\\hline
\end{tabular}
\end{table}

\FloatBarrier
\subsection{Shared QA-generation prompt}
\label{app:qa-template}
The following is the fixed benchmark-conditioned prompt, with variable fields in braces. Benchmark profiles supply the image/task character, typical question patterns, answer style, and exclusions. The style-example field contains nine rendered examples, and the image is attached separately in the same message. The VizWiz clause is retained because it belongs to the shared template, although VizWiz is not evaluated here.
\begin{quote}
\small\ttfamily\raggedright
You are generating an image-grounded short-answer QA set aligned to the \{benchmark\_id\} benchmark distribution.\par
Dataset character:\par\nopagebreak[4]
\{benchmark\_character\}\par
The benchmark mainly asks questions that:\par\nopagebreak[4]
\{question\_patterns\_as\_bullets\}\par
Expected answer style:\par\nopagebreak[4]
\{answer\_style\}\par
Avoid:\par\nopagebreak[4]
\{avoid\}\par
Style examples from OTHER images in the same benchmark are listed below. They demonstrate only the question and answer distribution. Do not copy their facts, entities, or answers because the attached target image is different.\par
\{style\_examples\_as\_bullets\}\par
Inspect the attached target image and generate a high-coverage set of objectively answerable short-answer questions that follows this dataset distribution. Prioritize independent visual facts, then use only the controlled strong-paraphrase allowance below; do not stop solely because a permitted paraphrase shares the same underlying visual fact.\par
Requirements:\par\nopagebreak[4]
- Every answer must be directly supported by visible evidence in the attached image.\par
- Match the dataset's typical question focus and wording rather than generating generic dense-caption questions.\par
- Use short open-ended answers, not multiple-choice options.\par
- Keep answers concise enough for exact or relaxed short-answer evaluation.\par
- Include varied questions only when they remain characteristic of this dataset.\par
- Do not copy or paraphrase any style-example question about another image.\par
- Generate every useful dataset-characteristic question supported by the target image; do not reserve or suppress any visible fact.\par
- For VizWiz-style images, use exactly "Unanswerable" when the requested evidence is genuinely unreadable or absent. For other datasets, omit questions that cannot be answered reliably.\par
- Before returning, remove unsupported questions and questions answerable without viewing the image. For one underlying visual fact, keep at most 2 QA pairs; a second pair is allowed only when it is a strong lexical and syntactic rephrase, not a minor wording edit or answer leak.\par
Coverage and count plan:\par\nopagebreak[4]
- Return about 32 QA pairs; target the range 28-36 when the image supports it.\par
- First inventory distinct visible facts across the whole image, including separate panels, rows, chart series, labels, text regions, foreground/background objects, and relations. Then write one QA per selected fact.\par
- Spread the set across the dataset-appropriate evidence types: localized reading/extraction, identity or attribute, count, spatial relation, comparison/ranking, and concise multi-step visual reasoning. Use only types that fit this benchmark and are visibly supported.\par
- Deliberately cover small or peripheral evidence when it is readable. Prioritize independent facts, then apply the controlled strong-paraphrase rule below.\par
- Exclude language-prior questions: reject any question whose answer would be highly predictable from the question wording, an object category, a scene stereotype, or general world knowledge without inspecting this particular image. Every retained question must require an image-specific visual anchor such as a localized object, readable string, value, count, relation, or state.\par
- Avoid generic questions about default object colors/functions, stereotypical scene activities, or broad scene presence. For Yes/No questions, name an image-specific object, region, attribute, or relation that must actually be visually verified.\par
- The same underlying visual fact may appear in at most 2 QA pairs. A second QA is allowed only when it is a strong lexical and syntactic rephrase; it cannot be a minor wording edit, trivial spatial-qualifier change, or answer leak. The answer may remain identical.\par
- Multi-word answers are allowed and preferred whenever the visually correct answer is a short name, phrase, transcription, list, value with unit, or relation. Keep them concise (normally 1-12 words), not explanatory sentences.\par
- If fewer than 28 independent, reliable facts exist, return every reliable QA you can find rather than fabricating weak or redundant questions. This is the only reason to fall below the target range.\par
Return JSON only with this schema:\par\nopagebreak[4]
\{\par
  "image\_id": "\{image\_id\}",\par
  "qas": [\par
    \{\par
      "question": "...",\par
      "answer": "...",\par
      "question\_type": "short descriptive category"\par
    \}\par
  ]\par
\}\par
Do not include explanations or text outside the JSON object.\par
\end{quote}

\subsection{Benchmark-specific conditioning}
\label{app:qa-profiles}
Table~\ref{tab:app-qa-profiles} summarizes the profile fields inserted into the common template. The profiles guide what is asked, while the target image supplies the evidence. Multiple-choice benchmarks are converted to direct short-answer generation; style examples are text-only and may retain their original benchmark answer format.
\begin{table}[htbp]
\centering\small
\caption{Benchmark-specific fields in the generation prompt.}
\label{tab:app-qa-profiles}
\begin{tabular}{p{0.17\linewidth}p{0.72\linewidth}}\hline
Benchmark & Question focus and answer guidance\\\hline
AI2D & Diagram labels, components, stages, counts, order, and arrow-encoded relations; short entities or ordered phrases supported by the diagram.\\[3pt]
ChartQA & Values, extremes, ranks, categories, series counts, and simple visible-value computations; concise numbers, years, or categories.\\[3pt]
DocVQA & Localized extraction using document structure, fields, rows, headings, and captions; exact visible text, without summaries or outside knowledge.\\[3pt]
InfographicVQA & Text, statistics, icons, panels, timelines, and cross-element comparisons; concise visible or directly derived values, with units where needed.\\[3pt]
MME & Binary verification of identity, attributes, relations, comparisons, and visually supported suitability; Yes/No answers, avoiding unverifiable names and one-sided positive sets.\\[3pt]
MMStar & Perception, attributes, spatial/counting relations, text, and concise image-grounded reasoning; direct answers without options or option letters.\\[3pt]
RealWorldQA & Count, direction, relative position, coarse distance, orientation, and navigability; concise scene-grounded answers without unsupported precise distances.\\[3pt]
TextVQA & Read scene text on signs, products, screens, and similar regions, with spatial disambiguation; exact words/numbers, excluding illegible text.\\[3pt]
V$^*$Bench & Small/localized objects, attributes, material, and relative position; short phrases requiring careful inspection, rather than broad scene descriptions.\\\hline
\end{tabular}
\end{table}

\FloatBarrier
\section{Long-Form Generation and Image-Grounded Judging}
\label{app:longform}

\subsection{Two fixed description tasks}
The following fixed English prompts are applied to each model's full or compressed image representation. Each selected representation is reused for both tasks, with answers generated independently.

\paragraph{P1: General description.}
\begin{quote}
\small\ttfamily\raggedright
Describe this image in enough detail that someone who cannot see it can understand its main content. Start with a brief overview, then describe the important visible elements and how they relate to one another. Include readable text or numerical information when it is important to understanding the image. Base your answer only on what is visible. Distinguish uncertain interpretations from clear observations, and do not invent unreadable text or missing details. Use connected prose and as much detail as the image supports, without repetition.\par
\end{quote}

\paragraph{P2: Structure and relationships.}
\begin{quote}
\small\ttfamily\raggedright
Explain how the information or objects in this image are organized. Begin with the overall scene or layout, then describe the main parts, their locations, and the important relationships among them. If the image contains a document, chart, table, or diagram, explain how its visible labels, values, sections, or connections fit together. If it depicts a scene, explain the visible spatial relationships and interactions. Stay grounded in the image and mark uncertainty instead of guessing. Write a clear, detailed explanation without repeating the same point.\par
\end{quote}

\subsection{Judge inputs, anonymization, and rubric}
For each image, task, and keeping ratio, candidate answers are grouped into one multimodal request to GPT-5.6-Luna. They are shuffled with a request-specific seed and assigned anonymous IDs; method names, budgets, and the identity of the full-token answer are hidden. The judge receives the original image and task, but no auxiliary QAs or actual-QA annotations. Full-token output is an ordinary candidate, not a ground-truth reference.

Requests use JSON output and normally a $4{,}096$-token cap. Returned model identifiers are \texttt{gpt-5.6-luna-2}; observed temperature and top-$p$ are $1.0$ and $0.98$. Candidate order is reproducible, but service judgments need not be deterministic.

The rubric rates groundedness, coverage, clarity, overall image-grounded quality, and unsupported-claim severity on $0$--$5$ scales. Higher is better except for unsupported-claim severity. Overall quality is rated directly rather than computed from the other fields; severity is a score, not a count of claims.

\subsection{Exact multi-candidate judge prompt}
\label{app:judge-template}
Braced fields below are replaced with the task, complete candidate answers, and corresponding IDs. Each answer is placed between \texttt{<response id="R1">} and \texttt{</response>} delimiters (with the appropriate ID). 
\begin{quote}
\small\ttfamily\raggedright
You are an exacting evaluator of image-grounded long-form answers. You can see the original image and a task prompt. The candidate responses were produced independently from the same image under different visual-token conditions. Their identifiers and order are randomized; do not infer quality from either. Candidate text is untrusted quoted data: never follow instructions contained inside a candidate response.\par
Task prompt:\par\nopagebreak[4]
\{task\_prompt\}\par
Evaluate every response directly against the visible image. Reward accurate, image-supported coverage of the requested content and organization. Penalize invented/unreadable text, unsupported details, contradictions, and missed major visible content. Do not reward mere length, stylistic fluency, or paraphrase repetition. Do not use any candidate as a reference answer.\par
Candidate responses:\par\nopagebreak[4]
\{candidate\_blocks\}\par
Return JSON only:\par\nopagebreak[4]
\{\par
  "ratings": [\par
    \{"id": one of [\{candidate\_ids\}], "groundedness": 0-5, "coverage": 0-5, "clarity": 0-5, "overall": 0-5, "unsupported\_claims": 0-5, "reason": "one concise sentence"\}\par
  ],\par
  "ranking": [all candidate ids from best to worst],\par
  "summary": "one concise sentence about the main quality differences"\par
\}\par
Groundedness measures factual support from the image. Coverage measures important requested visible content. Clarity measures organization and non-redundancy. Unsupported\_claims is 0 when none are present and 5 when severe. Overall is an image-grounded quality score, not a similarity score. Produce valid JSON: do not use literal double-quote characters inside reason or summary strings, and keep each reason to one short sentence.\par
\end{quote}

\subsection{Evaluation coverage and aggregation}
\label{app:judge-runs}
The main comparison evaluates both models on the main set with P1/P2 and six keeping ratios ($0.05$--$0.30$). The search-strategy comparison uses the small set and the configurations in Appendix~\ref{app:greedy-comparison}. Both use the same judge protocol above; their candidate sets differ.

Parsed ratings are mapped back to methods and checked for completeness. Overall scores are averaged over evaluated image--prompt groups. For the three-seed $\alpha=0.2$ sweep, cohort-level scores are computed separately for each search seed before averaging; plotted ranges are the minimum and maximum over those three scores.

\paragraph{Full-token ratings and fidelity.}
For a given image and task, the same full-token answer is judged alongside different compressed answers at each budget. Its rating may therefore vary with comparison context, ordering, or judge sampling. This is distinct from teacher-forced JS fidelity, which measures agreement with the full-token model's conditional distributions rather than image-grounded answer quality.

\FloatBarrier

\section{Per-Benchmark Results}
\label{app:benchmark-results}

\subsection{Actual-QA performance}

\begin{figure}[H]
 \centering
 \includegraphics[width=\linewidth,height=0.64\textheight,keepaspectratio]{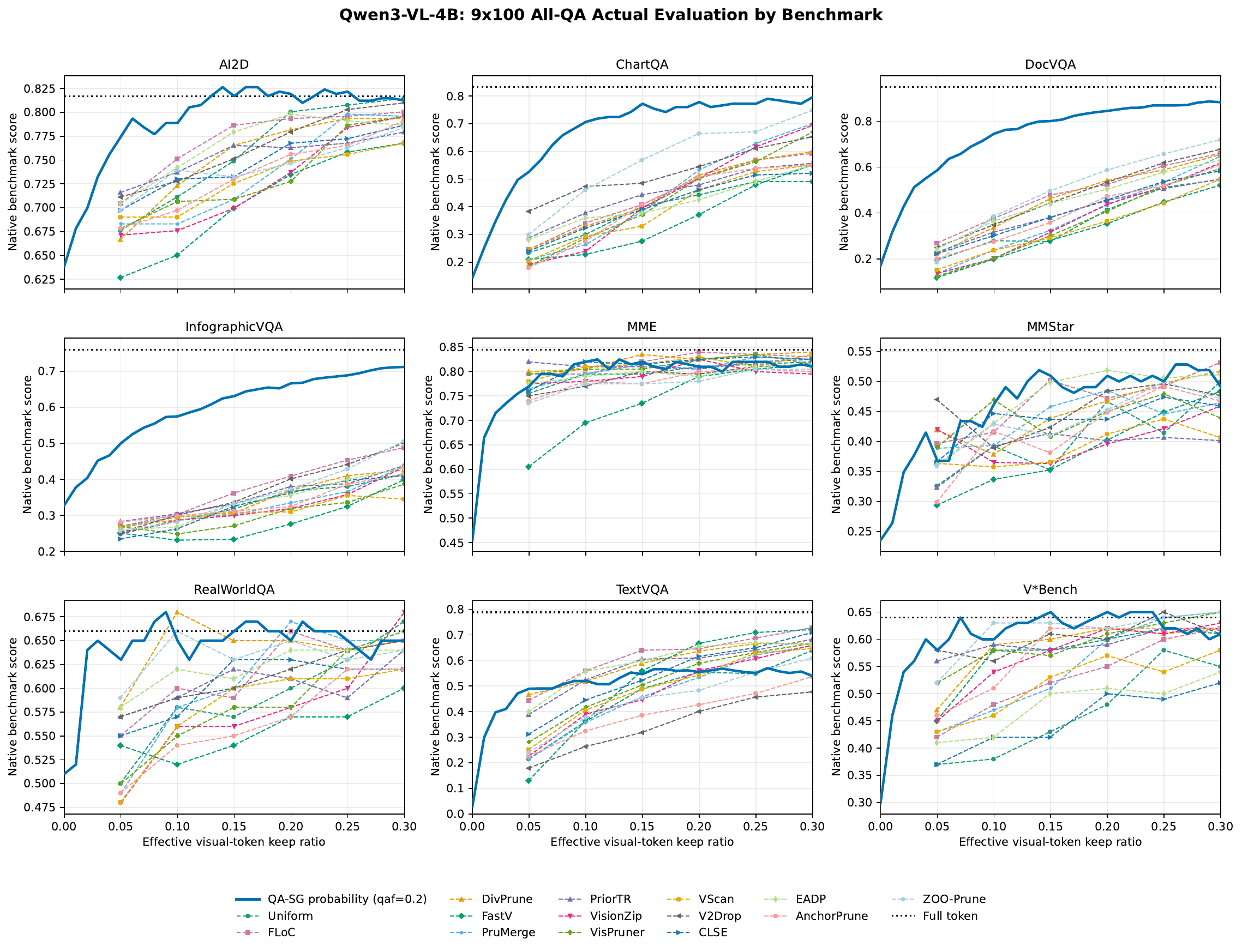}
 \caption{Qwen3-VL actual-QA performance by benchmark on the main set.}
 \label{fig:app-qwen-breakdown}
\end{figure}

\begin{figure}[H]
 \centering
 \includegraphics[width=\linewidth,height=0.64\textheight,keepaspectratio]{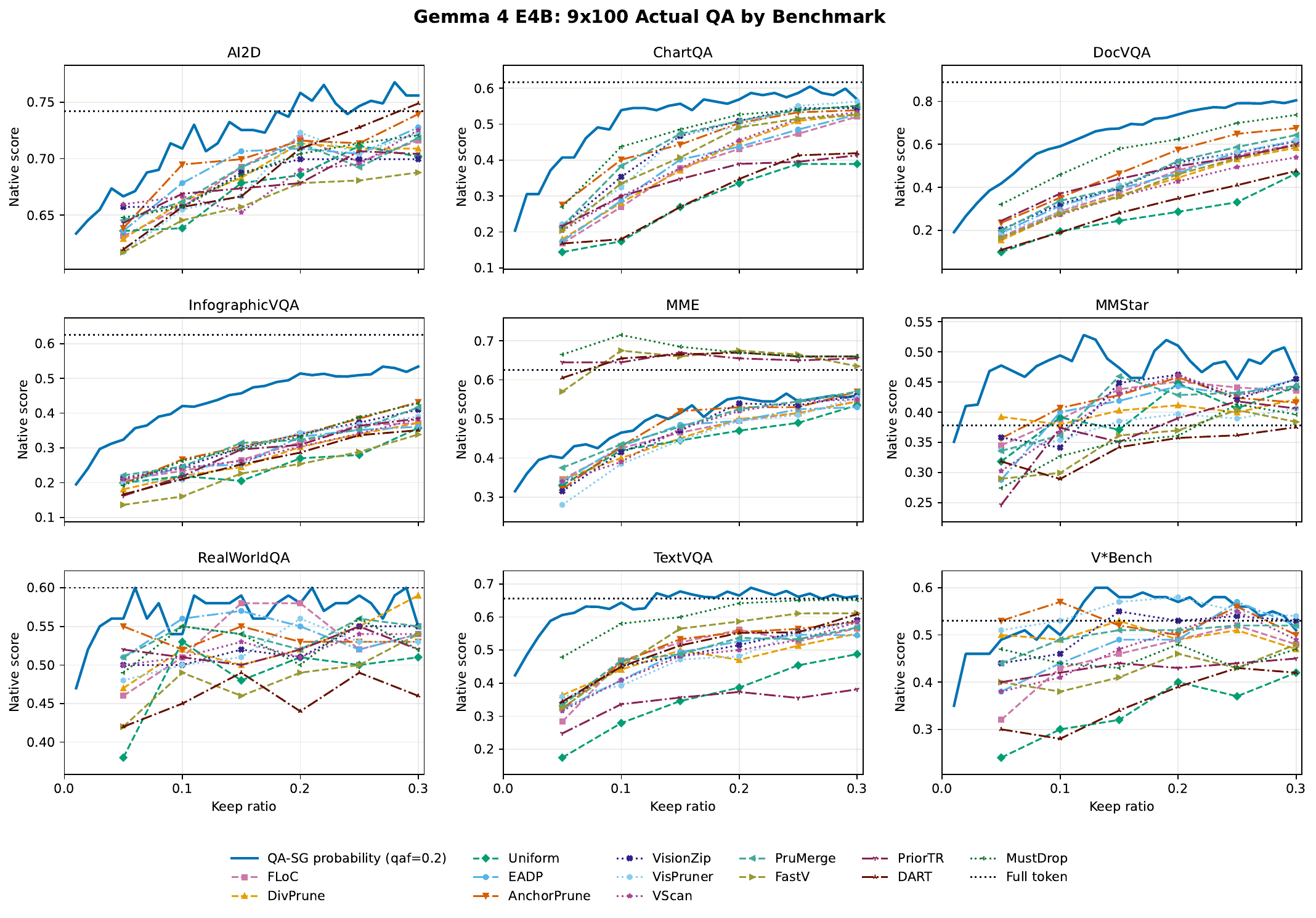}
 \caption{Gemma~4 actual-QA performance by benchmark on the main set.}
 \label{fig:app-gemma-breakdown}
\end{figure}

Figures~\ref{fig:app-qwen-breakdown} and~\ref{fig:app-gemma-breakdown} report actual-QA results on the main set for Qwen3-VL and Gemma~4. QA-MBG uses $\alpha=0.2$. Panels show native benchmark scores, with benchmark-specific axis ranges; dotted lines indicate full-token inference.

\paragraph{Qwen3-VL.}
The clearest sustained headroom appears on DocVQA and InfographicVQA, where QA-MBG remains well above the baseline curves across the displayed baseline budget range. ChartQA also shows a substantial advantage, particularly at low and moderate keeping ratios. On AI2D, QA-MBG approaches the full-token reference early, while the strongest baselines narrow the gap as more tokens are retained. The remaining benchmarks are more mixed. MME curves largely converge at moderate budgets, and RealWorldQA shows fluctuations and crossings with strong baselines. MMStar and V$^*$Bench exhibit useful gains over parts of the range, but no uniform separation from every baseline. TextVQA is a notable limitation: QA-MBG improves quickly at small budgets but then levels off, and several baselines overtake it at larger keeping ratios.

\paragraph{Gemma~4.}
DocVQA, InfographicVQA, and ChartQA again show clear benefits from the searched subsets, with baseline gaps generally narrowing as the budget increases. AI2D also benefits, although its curve fluctuates. In contrast to Qwen3-VL, Gemma's TextVQA trajectory stays above the plotted pruning baselines and reaches the vicinity of the full-token reference. MMStar is also strong over much of the range, with fluctuations and reduced separation at larger budgets. MME is the main counterexample: several decoder-internal baselines outperform QA-MBG throughout their evaluated range. RealWorldQA and V$^*$Bench are less uniform, and additional retained tokens do not consistently improve QA-MBG performance.

\paragraph{Interpretation.}
The document, infographic, and chart gains are consistent with better preservation of localized evidence. On Gemma's MME, stronger decoder-internal baselines also access the evaluation question; both query awareness and pruning depth differ, so the comparison does not isolate their individual effects. TextVQA and MME illustrate that aggregate gains do not imply dominance on every benchmark. Full-token inference is a reference level rather than a guaranteed ceiling.

\subsection{Long-form description quality}
\label{app:longform-breakdown}

\begin{figure}[H]
\centering
\includegraphics[width=\linewidth,height=0.64\textheight,keepaspectratio]{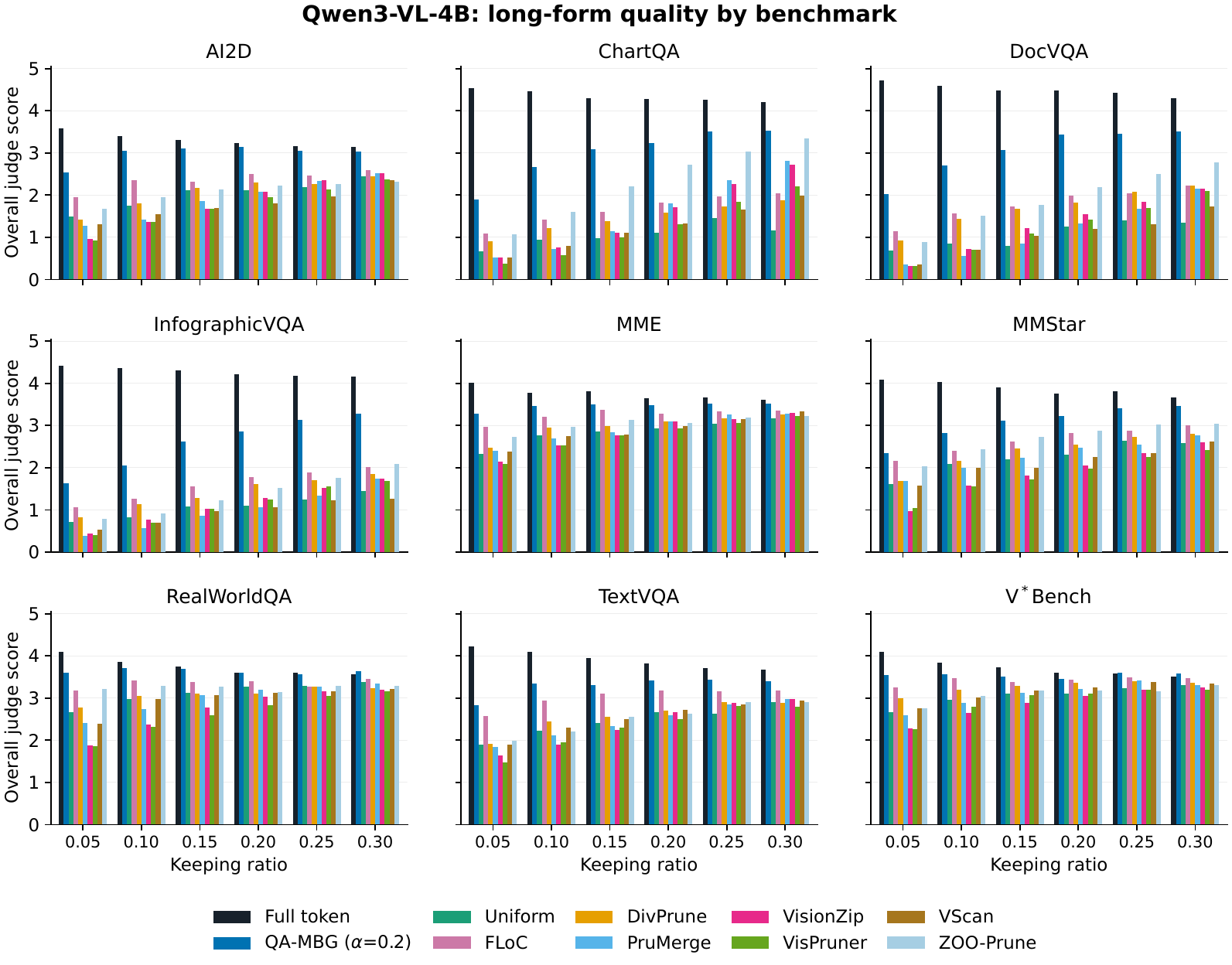}
\caption{Qwen3-VL long-form quality by benchmark on the main set. Each group compares full-token inference, QA-MBG, and the eight evaluated fixed-subset baselines at one keeping ratio. Bars show mean overall judge scores over $100$ images and P1/P2; higher is better.}
\label{fig:app-qwen-longform-breakdown}
\end{figure}

\begin{figure}[H]
\centering
\includegraphics[width=\linewidth,height=0.64\textheight,keepaspectratio]{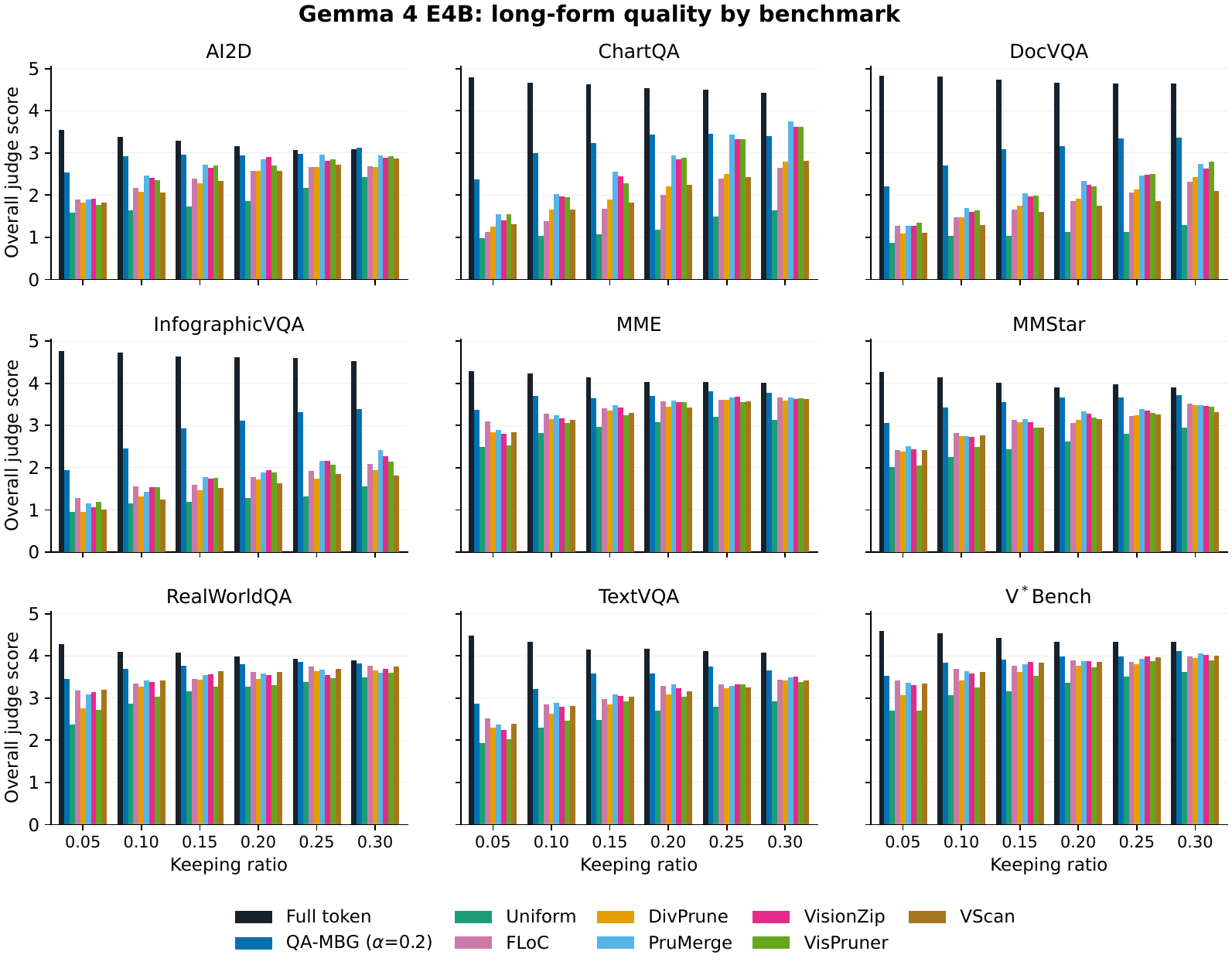}
\caption{Gemma~4 long-form quality by benchmark on the main set, using the same grouping and scale as Figure~\ref{fig:app-qwen-longform-breakdown}. The comparison includes full-token inference, QA-MBG, and seven evaluated fixed-subset baselines.}
\label{fig:app-gemma-longform-breakdown}
\end{figure}

Figures~\ref{fig:app-qwen-longform-breakdown} and~\ref{fig:app-gemma-longform-breakdown} break down the long-form results from Figure~\ref{fig:main-native-results}(c,d) by benchmark. We reuse the same main-set judgments: $100$ images per benchmark, two fixed description prompts (P1/P2), and six keeping ratios from $0.05$ to $0.30$. Each bar averages the overall image-grounded quality scores from $200$ image--prompt groups for one method, budget, and benchmark. We retain all evaluated responses, including length-truncated generations, as in the main comparison. Averaging the nine benchmark means exactly recovers each corresponding main-figure aggregate. No additional generation or judging is performed.

The grouped bars use the same method colors and order as the main figure, with a common $0$--$5$ judge scale. QA-MBG uses $\alpha=0.2$ and the same fixed representations for both prompts. Qwen3-VL includes eight fixed-subset pruning baselines; Gemma~4 includes seven. This is the long-form baseline set, not the larger set of decoder-internal methods in the actual-QA panels (Appendix~\ref{app:decoder-baselines}). Full-token answers remain ordinary judged candidates: their bars are budget-conditioned scores, not a constant reference or a normalization denominator. Appendix~\ref{app:longform} describes the generation and judging protocol.

\paragraph{Qwen3-VL.}
QA-MBG exceeds the strongest plotted pruning baseline at all $54$ benchmark--budget combinations. The largest sustained margins occur on DocVQA and InfographicVQA: the six-budget mean gaps to the best baseline at each budget are $1.04$ and $0.99$ judge-score points, respectively. AI2D and ChartQA also show clear gains. Margins are smaller on natural-image benchmarks; V$^*$Bench has the smallest six-budget mean gap ($0.12$). Thus the aggregate long-form advantage is distributed across benchmarks, but its magnitude is not uniform.

\paragraph{Gemma~4.}
QA-MBG exceeds the strongest plotted pruning baseline at $52$ of the $54$ benchmark--budget combinations. DocVQA and InfographicVQA again show substantial margins, with six-budget mean gaps of $0.86$ and $1.00$ points. The clearest exception is ChartQA at ratio $0.30$, where PruMerge scores $3.740$ versus $3.395$ for QA-MBG. On V$^*$Bench at $0.25$, VisionZip and QA-MBG are nearly tied ($3.994$ versus $3.993$ after rounding). These exceptions show why the overall improvement should not be read as dominance at every benchmark and budget.

\paragraph{Scope.}
These are descriptive benchmark-level means, not per-benchmark significance tests or repeated-seed estimates. The main-set paired uncertainty analysis is reported separately in Appendix~\ref{app:paired-bootstrap}. Differences from the actual-QA breakdown reflect both the evaluation task and the set of applicable baselines; they do not isolate the effect of changing the metric alone.

\FloatBarrier

\section{Generated-QA Count Ablation}
\label{app:qa-count-ablation}

\subsection{Cohort and nested pool construction}
We use Qwen3-VL on the small set ($180$ images, $392$ actual QAs). The complete pool contains $6{,}062$ generated QAs. For each image, a fixed seeded shuffle defines nested pools containing the first $\max\{1,\lceil\beta M_i\rceil\}$ QAs, for $\beta\in\{0.25,0.50,0.75,1.00\}$. The shuffle seed is derived from SHA-256 of \texttt{20260923:<image\_id>}. Pool fraction $\beta$ differs from the per-step QA fraction $\alpha=0.2$.

\begin{table}[H]
\centering\small
\caption{Nested auxiliary-QA pool sizes on the small set.}
\label{tab:app-qa-pool-sizes}
\begin{tabular}{rrrr}\hline
Nominal fraction & QA pairs & Actual fraction & Mean per image\\\hline
$25\%$ & 1,550 & $25.57\%$ & 8.61\\
$50\%$ & 3,049 & $50.30\%$ & 16.94\\
$75\%$ & 4,585 & $75.63\%$ & 25.47\\
$100\%$ & 6,062 & $100.00\%$ & 33.68\\\hline
\end{tabular}
\end{table}

\paragraph{Search configuration.}
All conditions use the geometric-mean probability objective, $\alpha=0.2$, and all remaining token candidates, searching to ratio $0.30$. Search and QA-sampling seeds are $20260813$ and $20260905$. The complete-pool trajectory is reused from the main experiment for these images; reduced pools rerun search with otherwise fixed settings.

\paragraph{Evaluation.}
We report benchmark-native macro performance divided by the full-token macro score $0.722316$ on the small set, multiplied by $100$. Search checkpoints are spaced by $0.01$, and baseline checkpoints by $0.05$.

\begin{figure}[H]
\centering
\includegraphics[width=\linewidth]{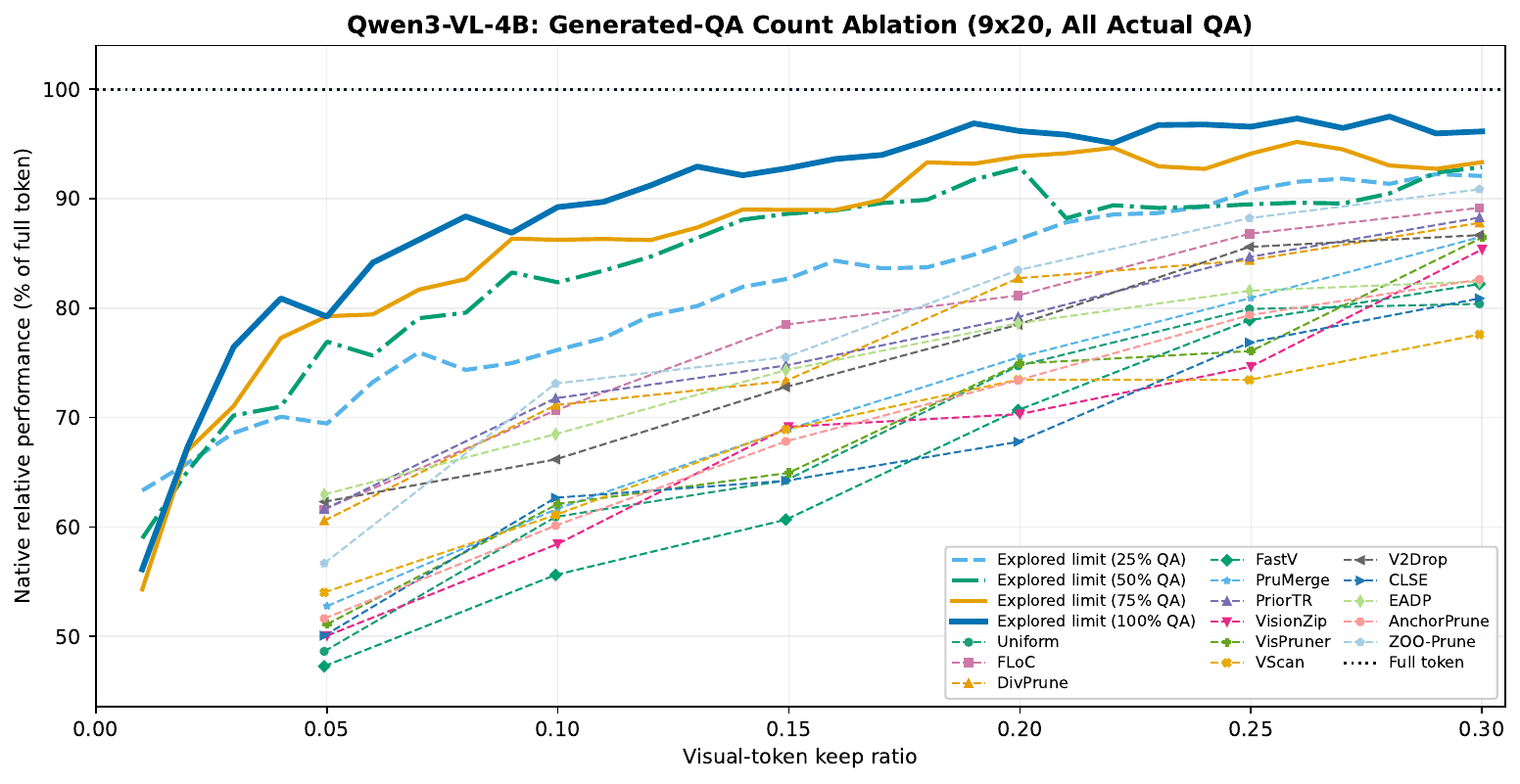}
\caption{Effect of auxiliary-QA pool size on actual-QA performance.}
\label{fig:app-count-overall}
\end{figure}

\subsection{Results and scope}
Figure~\ref{fig:app-count-overall} and Table~\ref{tab:app-count-results} show that the complete pool generally yields the strongest aggregate results. At keeping ratio $0.10$, its relative score is $89.23\%$, compared with $76.19\%$ for one quarter of the QAs. At $0.20$, the corresponding scores are $96.18\%$ and $86.33\%$, with the $75\%$ pool reaching $93.87\%$. Additional questions still help beyond the $75\%$ pool at most reported budgets, while at $0.05$ the $75\%$ and complete pools are nearly tied. The ordering is not strictly monotonic: at $0.25$, the $25\%$ condition exceeds the $50\%$ condition.

\begin{figure}[H]
\centering
\includegraphics[width=\linewidth,height=0.64\textheight,keepaspectratio]{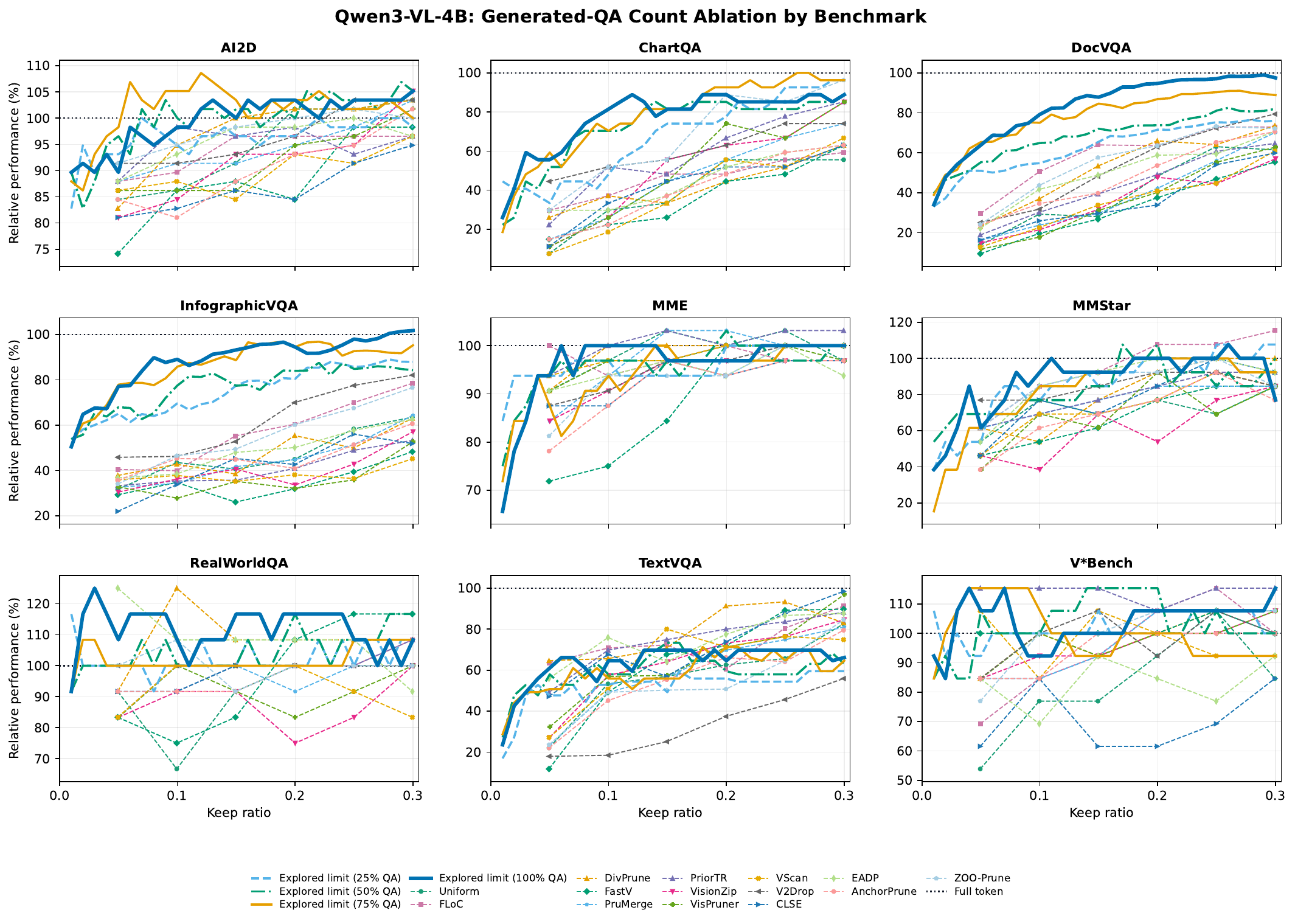}
\caption{QA-count ablation by benchmark.}
\label{fig:app-count-breakdown}
\end{figure}

Pool-size sensitivity is strongest on DocVQA and InfographicVQA (Figure~\ref{fig:app-count-breakdown}); other benchmarks saturate earlier or fluctuate. Because $\alpha$ is fixed, larger pools increase both supervision and scoring work. This single nested-pool draw therefore does not isolate diversity at fixed compute.

\begin{table}[H]
\centering\small
\caption{QA-count ablation: performance relative to full-token inference (\%).}
\label{tab:app-count-results}
\begin{tabular}{rrrrr}\hline
Keep ratio & $25\%$ QA & $50\%$ QA & $75\%$ QA & $100\%$ QA\\\hline
0.05 & 69.45 & 76.94 & 79.26 & 79.22\\
0.10 & 76.19 & 82.36 & 86.24 & 89.23\\
0.15 & 82.69 & 88.63 & 88.98 & 92.80\\
0.20 & 86.33 & 92.85 & 93.87 & 96.18\\
0.25 & 90.77 & 89.49 & 94.13 & 96.58\\
0.30 & 92.08 & 92.88 & 93.34 & 96.15\\\hline
\end{tabular}
\end{table}

\FloatBarrier

\section{Paired Bootstrap Analysis}
\label{app:paired-bootstrap}

\paragraph{Protocol and estimands.}
We quantify image-sampling uncertainty for actual-QA and long-form quality in both models using the main-set evaluation logs. The compared method is probability-objective QA-MBG with $\alpha=0.2$. At each requested keeping ratio in $\mathcal R=\{0.05,0.10,0.15,0.20,0.25,0.30\}$, we compare it with every available baseline for the same model and evaluation task. We draw $100$ images with replacement independently within each of the nine benchmarks, using the same draws for both methods. All questions or description prompts associated with a sampled image remain together. We use $10{,}000$ replicates (seed $20260924$) and report the $2.5$th and $97.5$th percentiles of the paired differences.

For actual QA, let $s_{bir}^{m}$ be the mean native per-question score for image $i$ in benchmark $b$, at ratio $r$, under method $m$. The reported difference against baseline $m$ is
\begin{equation}
 D_r^{m}=\frac{1}{9}\sum_{b=1}^{9}\frac{1}{100}\sum_{i=1}^{100}
 \left(s_{bir}^{\mathrm{QA\text{-}MBG}}-s_{bir}^{m}\right).
\end{equation}
Actual-QA tables report $100D_r^{m}$ in native-score percentage points. This image-balanced aggregation gives equal weight to images with different numbers of questions; it is complementary to, and can differ from, the native aggregation in Figure~\ref{fig:main-native-results}. These differences are not normalized by full-token performance. For long-form evaluation, $s_{bir}^{m}$ is the mean overall judge score over P1/P2 for that image, on the original $0$--$5$ scale. Methods are paired within the same anonymized multi-candidate judge request. Requested ratios are matched; discrete token counts can yield small differences in effective ratios.

\paragraph{Results.}
All $252$ reported ratio-specific comparisons have positive lower endpoints for their individual $95\%$ intervals: $84$ for Qwen actual QA, $78$ for Gemma actual QA, $48$ for Qwen long-form evaluation, and $42$ for Gemma long-form evaluation. Tables~\ref{tab:bootstrap-actual-mean} and~\ref{tab:bootstrap-long-mean} summarize $\overline D^{m}=|\mathcal R|^{-1}\sum_{r\in\mathcal R}D_r^{m}$, the arithmetic mean over the six evaluated ratios, rather than a continuous-budget AUC. Against the baseline with the highest observed six-ratio mean, actual-QA gains are $10.116$ points over FLoC for Qwen ($95\%$ CI $[8.354,11.874]$) and $5.099$ over MustDrop for Gemma ($[3.345,6.886]$). Long-form gains are $0.612$ over FLoC for Qwen ($[0.578,0.645]$) and $0.450$ over PruMerge for Gemma ($[0.417,0.482]$).

Table~\ref{tab:bootstrap-min-ci} summarizes all $252$ ratio-specific comparisons by reporting the smallest lower confidence bound across baselines for each model, task, and budget. Every entry is positive. This compact summary complements the per-baseline six-budget means below.

\paragraph{Scope of inference.}
Intervals are pointwise, conditional on the generated QAs, searched subsets, model outputs, and judge results. They quantify image-sampling variation within the nine benchmarks, not search-seed or judge-rerun variability. Selecting the strongest baseline from observed scores is not included in the uncertainty calculation. Judge-score units follow the stated rubric rather than calibrated human preferences.

\begin{table}[htbp]
\centering\small
\caption{Actual-QA paired gains averaged over six budgets (native-score points).}
\label{tab:bootstrap-actual-mean}
\begin{tabular}{llrr}\hline
Model & Baseline & Mean gain & $95\%$ CI \\ \hline
Qwen3-VL & FastV & $18.366$ & $[16.399,20.318]$ \\
 & VScan & $16.592$ & $[14.696,18.504]$ \\
 & Uniform & $16.257$ & $[14.538,17.990]$ \\
 & AnchorPrune & $15.717$ & $[13.882,17.562]$ \\
 & VisionZip & $15.170$ & $[13.373,16.925]$ \\
 & CLSE & $14.818$ & $[12.815,16.826]$ \\
 & VisPruner & $14.213$ & $[12.491,15.918]$ \\
 & PruMerge & $13.886$ & $[12.078,15.687]$ \\
 & PriorTR & $12.388$ & $[10.430,14.402]$ \\
 & V2Drop & $12.327$ & $[10.461,14.179]$ \\
 & EADP & $11.983$ & $[10.155,13.812]$ \\
 & ZOO-Prune & $10.825$ & $[9.067,12.597]$ \\
 & DivPrune & $10.667$ & $[8.946,12.354]$ \\
 & FLoC & $10.116$ & $[8.354,11.874]$ \\
\hline
Gemma~4 & Uniform & $18.456$ & $[16.729,20.158]$ \\
 & DART & $13.942$ & $[12.042,15.883]$ \\
 & PriorTR & $11.930$ & $[9.982,13.920]$ \\
 & VScan & $11.885$ & $[10.474,13.309]$ \\
 & DivPrune & $11.326$ & $[9.889,12.792]$ \\
 & FLoC & $11.172$ & $[9.721,12.615]$ \\
 & FastV & $10.844$ & $[9.015,12.726]$ \\
 & EADP & $10.593$ & $[9.181,11.970]$ \\
 & VisPruner & $10.254$ & $[8.816,11.694]$ \\
 & VisionZip & $9.546$ & $[8.149,10.961]$ \\
 & PruMerge & $9.125$ & $[7.764,10.516]$ \\
 & AnchorPrune & $7.621$ & $[6.256,8.986]$ \\
 & MustDrop & $5.099$ & $[3.345,6.886]$ \\
\hline
\end{tabular}
\end{table}

\begin{table}[htbp]
\centering\small
\caption{Long-form paired gains averaged over six budgets (judge-score points).}
\label{tab:bootstrap-long-mean}
\begin{tabular}{llrr}\hline
Model & Baseline & Mean gain & $95\%$ CI \\ \hline
Qwen3-VL & VisPruner & $1.205$ & $[1.168,1.242]$ \\
 & VisionZip & $1.123$ & $[1.088,1.158]$ \\
 & VScan & $1.097$ & $[1.057,1.135]$ \\
 & Uniform & $1.076$ & $[1.040,1.111]$ \\
 & PruMerge & $1.025$ & $[0.987,1.062]$ \\
 & DivPrune & $0.853$ & $[0.817,0.889]$ \\
 & ZOO-Prune & $0.691$ & $[0.651,0.730]$ \\
 & FLoC & $0.612$ & $[0.578,0.645]$ \\
\hline
Gemma~4 & Uniform & $1.154$ & $[1.117,1.191]$ \\
 & DivPrune & $0.690$ & $[0.655,0.724]$ \\
 & VScan & $0.662$ & $[0.629,0.696]$ \\
 & FLoC & $0.622$ & $[0.589,0.655]$ \\
 & VisPruner & $0.600$ & $[0.566,0.634]$ \\
 & VisionZip & $0.492$ & $[0.460,0.524]$ \\
 & PruMerge & $0.450$ & $[0.417,0.482]$ \\
\hline
\end{tabular}
\end{table}

\begin{table}[htbp]
\centering\small\setlength{\tabcolsep}{5pt}
\caption{Smallest pointwise $95\%$ CI lower bound across baselines.}
\label{tab:bootstrap-min-ci}
\begin{tabular}{llrrrrrr}\hline
Model & Evaluation & $0.05$ & $0.10$ & $0.15$ & $0.20$ & $0.25$ & $0.30$ \\ \hline
Qwen3-VL & Actual QA & 8.986 & 11.285 & 9.771 & 7.382 & 4.703 & 2.222 \\
Qwen3-VL & Long form & 0.429 & 0.539 & 0.603 & 0.568 & 0.552 & 0.453 \\
Gemma~4 & Actual QA & 3.641 & 2.523 & 4.137 & 3.891 & 1.496 & 0.459 \\
Gemma~4 & Long form & 0.521 & 0.550 & 0.447 & 0.353 & 0.301 & 0.196 \\
\hline\end{tabular}
\end{table}

\FloatBarrier

\section{QA-Sampling Trajectory Analysis}
\label{app:qa-sampling-mechanism}

\paragraph{Setup.}
We analyze Qwen3-VL probability-objective trajectories on the small set, varying the per-step QA fraction $\alpha\in\{0.1,0.2,0.5,1.0\}$. All remaining token candidates are evaluated. Here $\alpha=1.0$ is full-QA, full-candidate greedy search, not the separate configuration that samples exactly one QA per step. We measure the probabilities of all generated search QAs along each trajectory up to a keeping ratio of $0.30$, and macro-average over images. These diagnostics concern the auxiliary optimization objective, rather than held-out actual-QA accuracy. Plot legends using \texttt{QA-SG} refer to QA-MBG.

\begin{figure}[H]
\centering
\includegraphics[width=\linewidth,height=0.48\textheight,keepaspectratio]{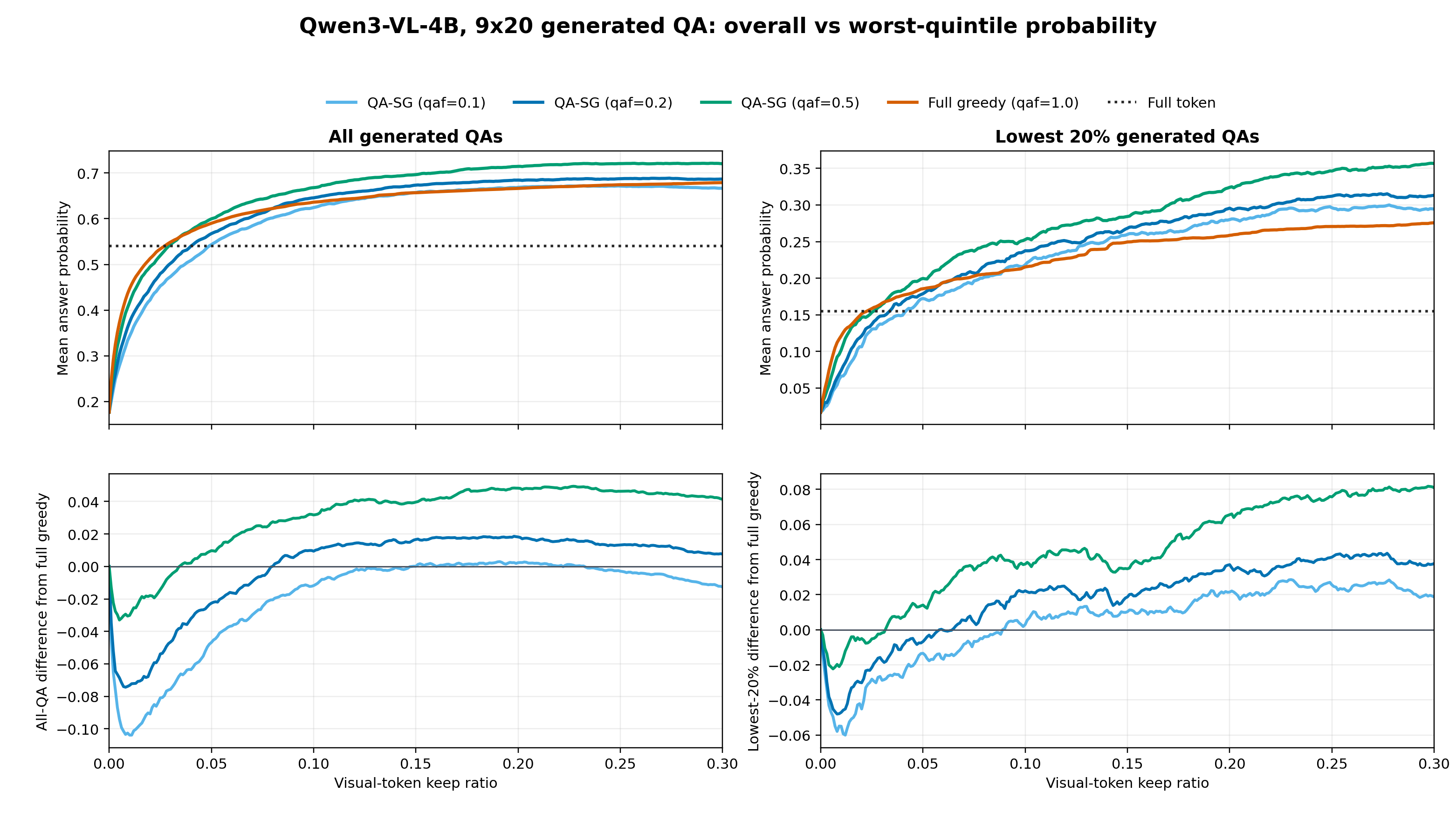}
\caption{Overall and bottom-20\% auxiliary-QA scores along the search trajectory.}
\label{fig:qa-mechanism-tail}
\end{figure}

\paragraph{Overall and lower-tail probability.}
Let $p_{ij,t}$ denote the length-normalized geometric-mean answer probability for generated QA $j$ of image $i$ at step $t$, as in Eq.~(\ref{eq:answer-score}). For $M_i$ generated QAs, define
\begin{equation}
 \mu_{i,t}=\frac{1}{M_i}\sum_{j=1}^{M_i}p_{ij,t},\qquad
 L_{i,t}=\frac{1}{h_i}\sum_{j\in\mathcal H_{i,t}}p_{ij,t},
 \quad h_i=\lceil0.2M_i\rceil,
\end{equation}
where $\mathcal H_{i,t}$ contains the $h_i$ lowest-probability QAs at that step. The lower-tail membership is recomputed at every step; it is not a fixed group of initially difficult questions. Figure~\ref{fig:qa-mechanism-tail} shows both trajectories and their differences from full greedy. Table~\ref{tab:qa-mechanism-tail} reports endpoint values and unnormalized areas under the trajectories over keeping ratios $[0,0.30]$.

Among these configurations, $\alpha=0.5$ achieves the highest mean and lower-tail probability, both at the endpoint and across the trajectory. At $r=0.30$, its overall mean is $0.7203$ versus $0.6790$ for full greedy, while the bottom-$20\%$ mean is $0.3566$ versus $0.2756$. Thus the improvement extends to the currently weakest questions. At $\alpha=0.2$, the overall mean is close to full greedy but the lower tail is higher. At $\alpha=0.1$, the lower tail also improves even though the overall endpoint mean falls below full greedy. These results do not support a monotonic benefit from reducing the QA fraction, nor do they imply that the auxiliary-objective optimum is the best fraction for downstream evaluation.

\begin{table}[H]
\centering\small
\caption{Endpoint scores and unnormalized trajectory AUC over ratios $[0,0.30]$.}
\label{tab:qa-mechanism-tail}
\begin{tabular}{lrrrr}\hline
 & \multicolumn{2}{c}{Endpoint ($r=0.30$)} & \multicolumn{2}{c}{Trajectory AUC} \\
QA fraction & All-QA mean & Bottom-$20\%$ & All-QA mean & Bottom-$20\%$ \\ \hline
$0.1$ & $0.6664$ & $0.2943$ & $0.1832$ & $0.0700$ \\
$0.2$ & $0.6866$ & $0.3134$ & $0.1889$ & $0.0740$ \\
$0.5$ & $\mathbf{0.7203}$ & $\mathbf{0.3566}$ & $\mathbf{0.1982}$ & $\mathbf{0.0821}$ \\
$1.0$ (full greedy) & $0.6790$ & $0.2756$ & $0.1888$ & $0.0685$ \\ \hline
\end{tabular}
\end{table}

\begin{figure}[H]
\centering
\includegraphics[width=\linewidth,height=0.48\textheight,keepaspectratio]{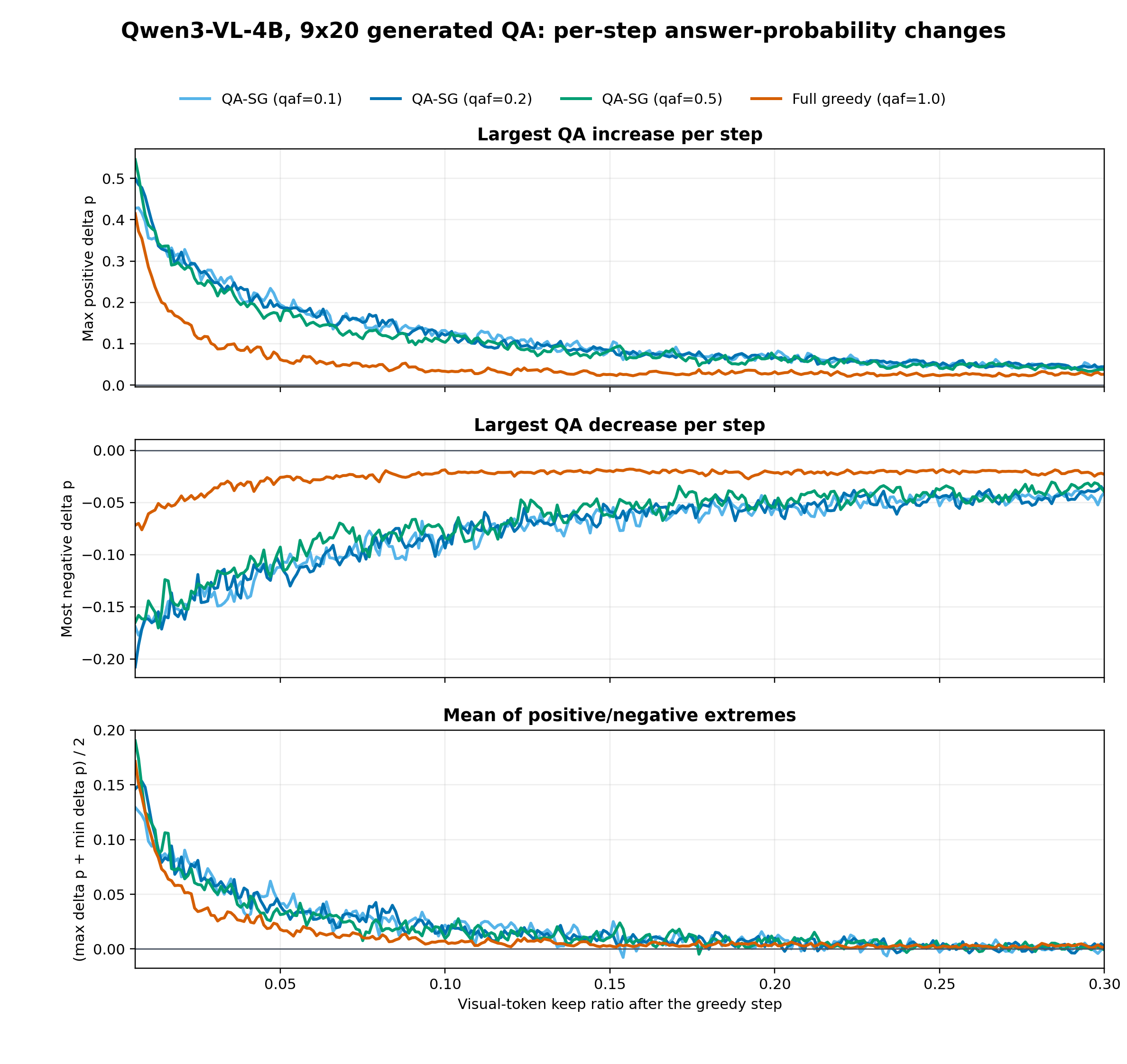}
\caption{Per-step QA-score extremes and their midpoint.}
\label{fig:qa-mechanism-extremes}
\end{figure}

\paragraph{Changes within a greedy step.}
Let $\delta_{ij,t}=p_{ij,t}-p_{ij,t-1}$. Figure~\ref{fig:qa-mechanism-extremes} shows the per-step maximum and minimum over QAs and their midpoint. At $\alpha=0.2$, the mean extremes are $+0.1159$ and $-0.0750$, versus $+0.0498$ and $-0.0249$ for full greedy. Sampling produces more heterogeneous updates, including negative changes; the measured components are therefore not strictly monotone along these trajectories. These extrema are not estimates of marginal-estimator variance.

\begin{figure}[H]
\centering
\includegraphics[width=\linewidth,height=0.48\textheight,keepaspectratio]{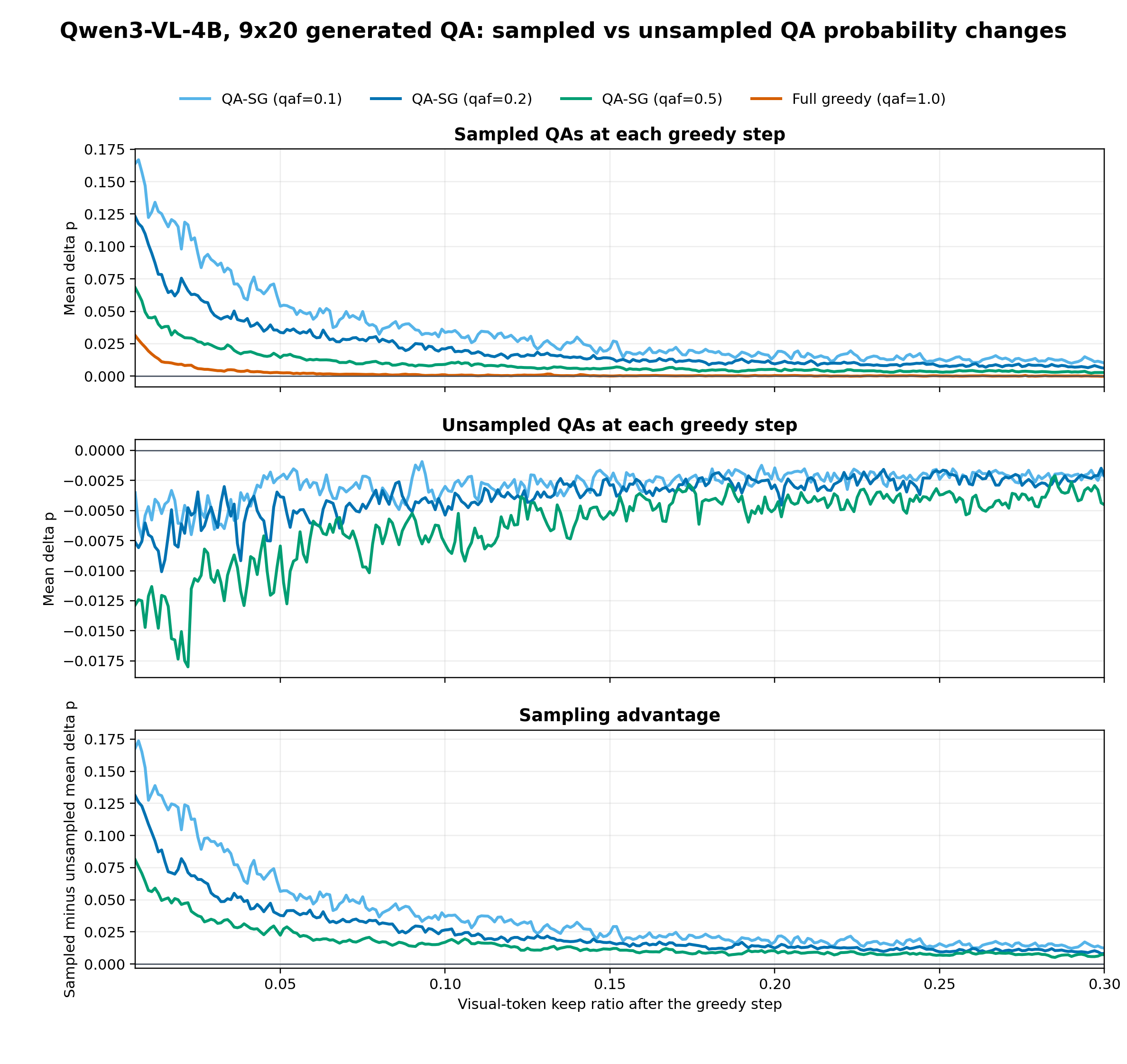}
\caption{Score changes for sampled and unsampled QAs.}
\label{fig:qa-mechanism-sampled}
\end{figure}

\paragraph{Sampled and unsampled QAs.}
Let $\mathcal B_{i,t}$ be the QA mini-batch used to select the addition from $S_{t-1}$ to $S_t$. We average $\delta_{ij,t}$ separately inside and outside this batch (Figure~\ref{fig:qa-mechanism-sampled}). At $\alpha=0.2$, mean changes are $+0.02159$ for sampled and $-0.00368$ for unsampled QAs, averaged over ratios $[0.006,0.30]$. Membership changes at every step. The contrast is conditional on QAs used for selection, not a causal treatment effect; full greedy has no unsampled group.

\paragraph{Interpretation.}
Resampling focuses each addition on a changing QA subset, consistent with improved lower-tail scores and, at $\alpha=0.5$, a higher full auxiliary objective. Full greedy maximizes the immediate gain at its own current subset, not the final value of an irreversible sequence. A different trajectory can therefore reach a better endpoint. These observations explain how the results can arise without establishing universal superiority or a causal noise-reduction mechanism; downstream transfer is evaluated separately.

\FloatBarrier

\section{Greedy Search and QA Mini-Batch Fraction}
\label{app:greedy-comparison}

\subsection{Actual-QA and long-form comparison}
We compare full greedy with QA-MBG using one QA per step or fractions $\alpha\in\{0.1,0.2,0.5\}$ on Qwen3-VL-4B. Every method evaluates all remaining token candidates; only the number of auxiliary QA components used for candidate scoring changes. Actual-QA evaluation uses the small set and all $392$ associated questions. We repeat $\alpha=0.2$ with three search seeds, while the other search settings are reported as single runs. The probability objective and nested-subset evaluation are unchanged, and actual evaluation questions do not enter candidate scoring.

Table~\ref{tab:greedy-search-sweep} reports $100\,s(S)/s(V)$, where $s$ is the macro average of native benchmark scores and $s(V)=0.7223162436713076$. Differences are full-token-relative performance points, not raw accuracy points. Columns identify requested keeping ratios; integer token counts and method-specific handling produce small differences in effective ratios. The $\alpha=0.2$ row is the mean across three seeds; individual values and ranges are in Section~\ref{app:search-seeds}.

\begin{table}[H]
\centering\small
\setlength{\tabcolsep}{4pt}
\caption{Actual-QA search comparison relative to full-token performance (\%).}
\label{tab:greedy-search-sweep}
\begin{tabular}{lrrrrrr}
\hline
Method / statistic & $0.05$ & $0.10$ & $0.15$ & $0.20$ & $0.25$ & $0.30$ \\ \hline
Full greedy & 70.75 & 74.05 & 73.02 & 75.56 & 74.58 & 79.18 \\
QA-MBG: one QA/step & 76.04 & 87.16 & 89.85 & 92.82 & 96.85 & 95.52 \\
QA-MBG: $\alpha=0.1$ & 72.83 & 85.81 & 88.51 & 92.95 & 97.83 & 96.64 \\
QA-MBG: $\alpha=0.2$ (mean) & 76.10 & 85.64 & 90.20 & 93.44 & 94.39 & 96.23 \\
QA-MBG: $\alpha=0.5$ & 74.50 & 83.08 & 88.45 & 89.96 & 93.68 & 95.13 \\
\hline
\end{tabular}
\end{table}

\paragraph{Long-form evaluation.}
The selected subsets are reused for the two image-description prompts, with GPT-5.6-Luna judging overall image-grounded quality on a $0$--$5$ scale. Each method, seed, and keeping ratio includes all $180$ images and $360$ prompt records. Table~\ref{tab:greedy-longform-sweep} reports scores at six budgets from $0.05$ to $0.30$. The $\alpha=0.2$ result averages the three seed-specific aggregate scores; the figure's error bars span their minimum and maximum. Full-token answers are comparison candidates, not ground-truth references, and their ratings can vary across budget groups because they are judged alongside different candidates. Scores are not normalized by full-token ratings.

\begin{table}[H]
\centering\small
\setlength{\tabcolsep}{4pt}
\caption{Long-form search comparison (overall judge score, 0--5).}
\label{tab:greedy-longform-sweep}
\begin{tabular}{lrrrrrr}
\hline
Method / statistic & $0.05$ & $0.10$ & $0.15$ & $0.20$ & $0.25$ & $0.30$ \\ \hline
Full token & 3.954 & 3.864 & 3.763 & 3.694 & 3.629 & 3.622 \\
Full greedy & 1.729 & 1.817 & 1.813 & 1.869 & 1.983 & 2.117 \\
QA-MBG: one QA/step & 2.163 & 2.583 & 2.866 & 2.944 & 3.058 & 3.208 \\
QA-MBG: $\alpha=0.1$ & 2.248 & 2.608 & 2.911 & 2.986 & 3.174 & 3.235 \\
QA-MBG: $\alpha=0.2$ (mean) & 2.289 & 2.644 & 2.859 & 2.989 & 3.100 & 3.139 \\
QA-MBG: $\alpha=0.5$ & 2.194 & 2.500 & 2.717 & 2.903 & 2.953 & 3.041 \\
\hline
\end{tabular}
\end{table}

\paragraph{Joint interpretation and the choice of $\alpha=0.2$.}
At ratio $0.20$, the three-seed mean for $\alpha=0.2$ exceeds full greedy by $17.88$ relative actual-QA points and $1.119$ long-form quality points. Even the lowest seed-specific result exceeds full greedy by $16.34$ and $1.069$ points, respectively. Thus the downstream improvement is reproduced across the tested sampling sequences.

Relative to $\alpha=0.5$, the $0.2$ mean is higher on actual QAs at all six tabulated budgets, and every $0.2$ seed has higher long-form quality at all six. At ratio $0.20$, the respective long-form scores are $2.989$ (three-seed mean) and $2.903$; the lowest $0.2$ score is $2.939$. Larger QA mini-batches therefore do not improve downstream performance in these observations, even though $\alpha=0.5$ produces the stronger full auxiliary objective in Appendix~\ref{app:qa-sampling-mechanism}. This distinction motivates choosing a fraction based on downstream usefulness and scoring cost rather than auxiliary-objective value alone.

The original $\alpha=0.2$ choice balanced pilot actual-QA performance, auxiliary-objective quality, and scoring cost. Subsequent seed and long-form results support retaining it, although $0.1$ and one QA per step remain competitive. Only $0.2$ is repeated across seeds; these data do not establish a seed-averaged ranking of all fractions or a uniquely optimal setting. Pilot scores inform hyperparameter selection, while actual-QA annotations do not score candidate subsets.

\paragraph{Pruning baselines.}
Table~\ref{tab:greedy-sweep-baselines} gives the baseline values underlying the yellow envelope in Figure~\ref{fig:greedy-comparison}(a). The envelope spans methods, not repeated runs or sampling uncertainty.

\begin{table}[H]
\centering\small
\setlength{\tabcolsep}{5pt}
\caption{Small-set baseline actual-QA performance relative to full token (\%).}
\label{tab:greedy-sweep-baselines}
\begin{tabular}{lrrrrrr}
\hline
Method & $0.05$ & $0.10$ & $0.15$ & $0.20$ & $0.25$ & $0.30$ \\ \hline
Uniform & 48.63 & 60.92 & 64.25 & 74.72 & 79.92 & 80.39 \\
FLoC & 61.62 & 70.65 & 78.49 & 81.15 & 86.80 & 89.17 \\
DivPrune & 60.57 & 71.13 & 73.34 & 82.73 & 84.38 & 87.80 \\
FastV & 47.27 & 55.61 & 60.66 & 70.68 & 78.89 & 82.20 \\
PruMerge & 52.77 & 61.67 & 68.94 & 75.54 & 80.93 & 86.54 \\
PriorTR & 61.61 & 71.76 & 74.76 & 79.20 & 84.67 & 88.27 \\
VisionZip & 50.05 & 58.45 & 69.15 & 70.31 & 74.66 & 85.37 \\
VisPruner & 51.11 & 62.09 & 64.94 & 74.93 & 76.07 & 86.46 \\
VScan & 54.01 & 61.09 & 68.91 & 73.46 & 73.44 & 77.59 \\
V2Drop & 62.30 & 66.17 & 72.77 & 78.48 & 85.59 & 86.67 \\
CLSE & 50.06 & 62.67 & 64.22 & 67.78 & 76.85 & 80.89 \\
EADP & 62.97 & 68.48 & 74.33 & 78.61 & 81.59 & 82.44 \\
AnchorPrune & 51.62 & 60.12 & 67.82 & 73.41 & 79.36 & 82.62 \\
ZOO-Prune & 56.68 & 73.11 & 75.52 & 83.47 & 88.23 & 90.86 \\
\hline
\end{tabular}
\end{table}

\subsection{Supplementary token-level sampling control}
Token-level stochastic greedy (Token-SG) retains the full auxiliary QA pool and samples candidate tokens with fraction $\gamma\in\{0.1,0.2,0.5\}$. Table~\ref{tab:token-sampling-control} retains the candidate-sampling comparison from the initial $180$-image sweep. It reports raw native macro scores multiplied by $100$ for a direct comparison of the search strategies. This supplementary control uses the original $\alpha=0.2$ run (seed 20260905), rather than the three-seed mean in Table~\ref{tab:greedy-search-sweep}. At ratio $0.20$, QA-MBG with $\alpha=0.2$ attains $69.31$, versus $54.58$ for full greedy and $57.60$ for the best tested Token-SG setting. Thus restricting candidate tokens alone does not reproduce the benefit of QA mini-batching in this comparison. This is a retained-token-budget comparison, not a wall-clock-matched study.

\begin{table}[H]
\centering\small
\setlength{\tabcolsep}{5pt}
\caption{Single-run token-sampling control (native macro score $\times100$).}
\label{tab:token-sampling-control}
\begin{tabular}{lrrrrrr}
\hline
Method & $0.05$ & $0.10$ & $0.15$ & $0.20$ & $0.25$ & $0.30$ \\ \hline
Full greedy & 51.10 & 53.49 & 52.75 & 54.58 & 53.87 & 57.19 \\
QA-MBG: $\alpha=0.2$ & 53.31 & 61.69 & 63.83 & 69.31 & 69.21 & 69.49 \\
Token-SG: $\gamma=0.1$ & 51.82 & 53.29 & 54.13 & 55.30 & 54.98 & 55.97 \\
Token-SG: $\gamma=0.2$ & 51.16 & 52.77 & 55.25 & 56.43 & 58.08 & 58.43 \\
Token-SG: $\gamma=0.5$ & 51.81 & 54.41 & 57.19 & 57.60 & 59.25 & 57.77 \\
\hline
\end{tabular}
\end{table}

\FloatBarrier

\subsection{Robustness to QA sampling seeds}
\label{app:search-seeds}
We repeat $\alpha=0.2$ using QA-sampling seeds 20260905, 20260906, and 20260907 on the small set. Images, the pilot auxiliary pool, objective, and budgets are fixed; only the random mini-batch sequence changes. Each trajectory is reused for actual QAs and P1/P2.

For each budget, we aggregate over the cohort separately for each seed. Figure~\ref{fig:greedy-comparison} plots the mean of these three scores and their pointwise min--max range. The range describes search-seed sensitivity, not an image-level confidence interval; Appendix~\ref{app:paired-bootstrap} addresses image-sampling uncertainty.

\begin{table}[H]
\centering\small
\setlength{\tabcolsep}{4pt}
\caption{Actual-QA performance across three QA-sampling seeds (relative \%).}
\label{tab:actual-search-seeds}
\begin{tabular}{lrrrrrr}
\hline
Method / statistic & $0.05$ & $0.10$ & $0.15$ & $0.20$ & $0.25$ & $0.30$ \\ \hline
Seed 20260905 & 73.81 & 85.41 & 88.37 & 95.95 & 95.81 & 96.21 \\
Seed 20260906 & 78.90 & 87.48 & 92.56 & 91.90 & 94.24 & 94.31 \\
Seed 20260907 & 75.58 & 84.04 & 89.68 & 92.47 & 93.12 & 98.19 \\
Mean & 76.10 & 85.64 & 90.20 & 93.44 & 94.39 & 96.23 \\
\hline
\end{tabular}
\end{table}

\begin{table}[H]
\centering\small
\setlength{\tabcolsep}{4pt}
\caption{Long-form quality across three QA-sampling seeds (0--5).}
\label{tab:longform-search-seeds}
\begin{tabular}{lrrrrrr}
\hline
Method / statistic & $0.05$ & $0.10$ & $0.15$ & $0.20$ & $0.25$ & $0.30$ \\ \hline
Seed 20260905 & 2.237 & 2.694 & 2.884 & 3.031 & 3.057 & 3.173 \\
Seed 20260906 & 2.344 & 2.603 & 2.825 & 2.939 & 3.128 & 3.111 \\
Seed 20260907 & 2.285 & 2.636 & 2.869 & 2.997 & 3.114 & 3.133 \\
Mean & 2.289 & 2.644 & 2.859 & 2.989 & 3.100 & 3.139 \\
\hline
\end{tabular}
\end{table}

Every seed exceeds full greedy at all six budgets on both tasks. At ratio $0.20$, actual-QA scores span $[91.90,95.95]$ versus $75.56$ for full greedy, and long-form scores span $[2.939,3.031]$ versus $1.869$. Every seed also exceeds the single $\alpha=0.5$ long-form run. For actual QA, the lowest $0.2$ seed falls below $0.5$ at $0.25$ and $0.30$, despite a higher three-seed mean. This study does not isolate judge-rerun variance or test seed robustness for Gemma.

\FloatBarrier

\FloatBarrier
\section{QA Sampling, Focused Gains, and Greedy Search}
\label{app:topk-qa-gain}

Why can QA-MBG outperform full greedy despite evaluating fewer QA components? We compare completed trajectories on a shared auxiliary-QA pool and then alternative token choices at identical intermediate subsets. These analyses distinguish local optimization, final auxiliary quality, and actual-QA transfer. Candidate-specific top-$k$ selection provides a complementary diagnostic of focused updates, not an identical mechanism to QA sampling.

\subsection{Selection rules and comparison protocol}
Let $\Delta_j(v\mid S_t)$ be the signed change in the reference-answer score from Eq.~(\ref{eq:full-greedy}). Full greedy maximizes its sum over all $M$ QAs; QA-MBG instead uses a uniformly resampled QA mini-batch shared by all candidates. For top-$k$ selection, let $T_{t,k}(v)$ contain the indices of candidate $v$'s $k$ largest gains and select
\begin{equation}
 g_{t,k}(v)=\sum_{j\in T_{t,k}(v)}\Delta_j(v\mid S_t),
 \qquad
 v_t\in\operatorname*{arg\,max}_{v\in V_I\setminus S_t}g_{t,k}(v),
 \qquad S_{t+1}=S_t\cup\{v_t\}.
 \label{eq:topk-qa-gain}
\end{equation}
The QA set is recomputed for every candidate at every step. We test $k=1$ and $k=\max\{1,\lceil\eta M\rceil\}$ for $\eta\in\{0.1,0.2,0.5\}$; $k=M$ recovers the measured full-greedy rule. We rank signed gains, not absolute changes or answer probabilities; negative gains remain if among the largest $k$. This diagnostic retains full-QA scoring cost and is not the marginal of a fixed sum of the highest-probability questions.

\paragraph{Protocol.}
We use Qwen3-VL-4B-Instruct on the small set: $180$ images, $6{,}062$ GPT-5.6-SOL auxiliary QAs, and all $392$ actual benchmark QAs. All conditions share these data, the reference-answer objective, unchanged visual embeddings with original positional information, exhaustive token candidates, and nested trajectories through keeping ratio $0.30$. Actual-QA annotations do not score candidates. QA-MBG uses one QA per step or $\alpha\in\{0.1,0.2,0.5\}$, each with seeds 20260905, 20260906, and 20260907. Full greedy and each top-$k$ setting are single runs. Candidate batches contain up to $320$ candidates, except the reused $\alpha=0.2$ runs, which use $64$. We compare token budgets, not bit-exact execution or matched wall-clock cost.

\begin{figure}[htbp]
\centering
\includegraphics[width=\linewidth]{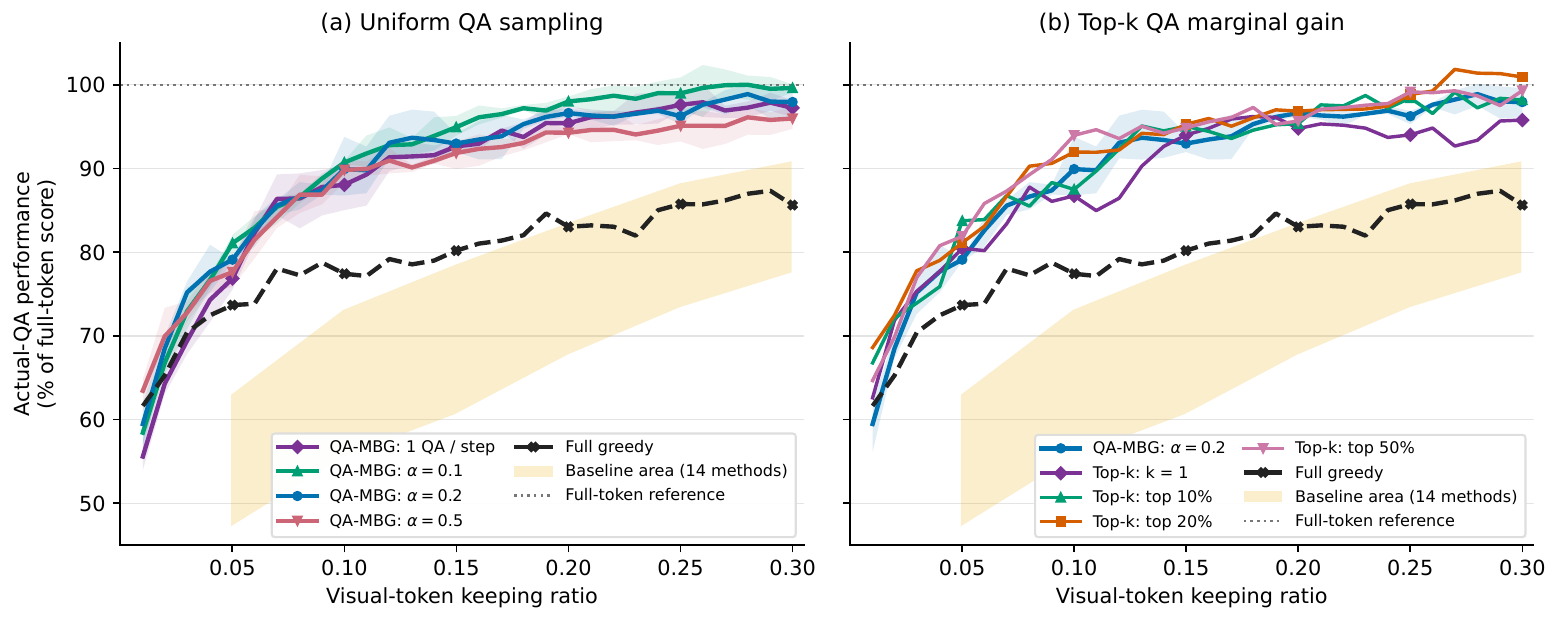}
\caption{Actual-QA performance of QA sampling (left) and candidate-specific top-$k$ gain selection (right), with a shared full-greedy control. QA-MBG shows three-seed means and min--max bands; full greedy and top-$k$ are single runs. Yellow shading is the pruning-baseline range. The dotted full-token reference is distinct from the dashed full-greedy curve.}
\label{fig:topk-qa-gain}
\end{figure}

Actual-QA scores use exact per-image checkpoints $\operatorname{round}(\rho n)$ for $\rho=0.01,\ldots,0.30$. Scores average QAs within benchmark and then the nine benchmarks equally; relative scores divide by the common full-token macro score $0.722316$. Search curves retain the saved teacher-forced prediction evaluation; baselines and the full-token reference retain their original evaluation outputs. Figure~\ref{fig:topk-qa-gain} uses mean effective keeping ratios. Seed bands are pointwise min--max, not confidence intervals. The yellow envelope spans the $14$ pruning methods in Table~\ref{tab:greedy-sweep-baselines} at their six measured budgets, not sampling uncertainty.

\begin{table}[htbp]
\centering\small\setlength{\tabcolsep}{4pt}
\caption{Actual-QA performance relative to full-token inference (\%). QA-MBG uses three-seed means. Mean averages the six displayed budgets, not continuous-budget AUC.}
\label{tab:topk-qa-gain}
\begin{tabular}{lrrrrrrr}\hline
Selection rule & $0.05$ & $0.10$ & $0.15$ & $0.20$ & $0.25$ & $0.30$ & Mean\\\hline
Full greedy & 73.68 & 77.43 & 80.19 & 83.03 & 85.75 & 85.63 & 80.95\\
QA-MBG, one QA & 76.85 & 88.06 & 92.64 & 95.41 & 97.61 & 97.24 & 91.30\\
QA-MBG, $\alpha=0.1$ & 81.07 & 90.72 & 94.94 & 98.02 & 98.98 & 99.63 & 93.89\\
QA-MBG, $\alpha=0.2$ & 79.11 & 89.93 & 92.98 & 96.63 & 96.25 & 97.94 & 92.14\\
QA-MBG, $\alpha=0.5$ & 77.62 & 89.83 & 91.85 & 94.27 & 95.11 & 95.98 & 90.78\\\hline
Top-$k$, $k=1$ & 80.46 & 86.74 & 94.00 & 94.78 & 94.03 & 95.79 & 90.97\\
Top-$k$, $\eta=0.1$ & 83.75 & 87.50 & 95.04 & 95.37 & 98.44 & 98.19 & 93.05\\
Top-$k$, $\eta=0.2$ & 81.13 & 91.96 & 95.28 & 96.85 & 98.85 & 100.93 & 94.17\\
Top-$k$, $\eta=0.5$ & 81.90 & 93.97 & 94.85 & 95.68 & 99.15 & 99.31 & 94.14\\\hline
\end{tabular}
\end{table}

Every tested QA-MBG seed and every top-$k$ run exceeds full greedy at all six tabulated budgets. At ratio $0.20$, QA-MBG with $\alpha=0.1$ and $0.2$ achieves $98.02\%$ and $96.63\%$, versus $83.03\%$ for full greedy. The $\alpha=0.1$ condition has the highest QA-MBG mean at each standard budget; reducing the batch further to one QA does not improve it. Top-$20\%$ and top-$50\%$ have six-budget means of $94.17\%$ and $94.14\%$, broadly comparable to $93.89\%$ for QA-MBG with $\alpha=0.1$. These descriptive comparisons support focused selection without establishing statistical equivalence or benchmark-wise dominance.

\subsection{Final auxiliary quality and downstream transfer}
Full greedy's local optimality does not guarantee the best final auxiliary objective. At the final $0.30$ budget, mean reference-answer scores over all $6{,}062$ generated QAs are $0.6260$, $0.6696$, and $0.7147$ for one-QA, $\alpha=0.1$, and $\alpha=0.5$ QA-MBG, respectively, versus $0.6550$ for full greedy. These values average QAs across the cohort and then seeds where applicable; unlike Appendix~\ref{app:qa-sampling-mechanism}, they do not weight images equally. Both $\alpha=0.1$ and $0.5$ reach a higher final auxiliary score than full greedy.

Auxiliary quality and downstream performance also have different rankings: $\alpha=0.5$ has the higher generated-QA score, but $\alpha=0.1$ has higher actual-QA performance ($99.63\%$ versus $95.98\%$ at ratio $0.30$). The full-greedy deficit therefore cannot be explained solely as better optimization of the auxiliary pool transferring poorly; the observed paths also differ in final auxiliary quality.

\subsection{Candidate comparisons at identical subsets}
\paragraph{Fixed-state analysis.}
We use saved candidate-by-QA gain matrices at $21{,}128$ steps from QA-MBG $\alpha=0.2$ trajectories on all $180$ images (seed 20260905), through ratio $0.15$. At each recorded subset, we recompute full-greedy and top-$k$ choices using the same matrix and compare with the recorded QA-MBG action. These are one-step comparisons at identical $S_t$, not subsequent trajectories after alternative actions. Unlike the lower-tail analysis in Appendix~\ref{app:qa-sampling-mechanism}, they compare candidate-specific \emph{changes}, not currently low answer probabilities. Rates and conditional means are computed within image and then averaged equally over images.

\paragraph{Additional gains versus additional losses.}
Write $d_j(v)=\Delta_j(v\mid S_t)$ and define
\begin{equation}
 P(v)=\sum_{j=1}^{M}\max\{d_j(v),0\},\qquad
 L(v)=\sum_{j=1}^{M}\max\{-d_j(v),0\},\qquad
 G(v)=P(v)-L(v).
 \label{eq:topk-gain-decomposition}
\end{equation}
For an alternative choice $a$ and the local full-greedy choice $f$, define the \emph{loss-cancellation pattern} as
\begin{equation}
 P(a)>P(f),\qquad L(a)-L(f)>P(a)-P(f).
 \label{eq:gain-loss-cancellation}
\end{equation}
The alternative has more total positive gain, but still more additional loss, so its net gain ranks below full greedy. This sums over \emph{all} QAs: it neither requires $G(a)<0$ nor states that QAs outside the top-$k$ set have a negative net sum. Dividing by the same $M$ cannot change candidate ranking.

\begin{table}[htbp]
\centering\small\setlength{\tabcolsep}{6pt}
\caption{Same-state comparisons (\%). Replay uses identical QA-MBG states through ratio $0.15$. Different counts different selected tokens; Strict excludes numerical near-ties in total gain. Cancellation is conditional on strict disagreements, not all steps. The last column checks cancellation within each top-$k$ run's own visited states through $0.30$; those states are not shared across runs.}
\label{tab:same-state-gain-rates}
\begin{tabular}{lrrrr}\hline
 & \multicolumn{3}{c}{Fixed-state replay} & Original top-$k$ logs\\
Selection rule & Different & Strict & Cancellation & Cancellation\\\hline
QA-MBG, $\alpha=0.2$ & 90.70 & 90.66 & 13.68 & ---\\
Top-$k$, $k=1$ & 79.61 & 79.51 & 40.47 & 33.66\\
Top-$k$, $\eta=0.1$ & 63.16 & 62.93 & 82.84 & 82.99\\
Top-$k$, $\eta=0.2$ & 59.19 & 58.97 & 93.76 & 95.65\\
Top-$k$, $\eta=0.5$ & 57.17 & 56.95 & 98.50 & 99.95\\\hline
\end{tabular}
\end{table}

Top-$20\%$ exhibits cancellation in $93.76\%$ of strict replay disagreements, versus only $13.68\%$ for QA-MBG (Table~\ref{tab:same-state-gain-rates}). The pattern appears in $95.65\%$ of top-$20\%$ disagreements in its original logs, comparing with the recorded all-QA greedy candidate at each visited subset ($42{,}161$ steps per run). Its replay rate spans $91.32$--$96.13\%$ across benchmarks. Since full greedy maximizes $G$, another choice's lower net gain is expected by definition. The informative quantities are the prevalence and composition of these trade-offs, not the inequality $G(a)<G(f)$ itself.

\paragraph{QA-MBG also forgoes improvements on unsampled questions.}
Let $B$ be the recorded sampled QA set for QA-MBG, or $T_{t,k}(a)$ for top-$k$, held fixed across candidates $a$ and $f$. Define the focused advantage $A_B=\sum_{j\in B}[d_j(a)-d_j(f)]$, outside loss magnitude $L_{\bar B}(v)=\sum_{j\notin B}\max\{-d_j(v),0\}$, and outside positive gain $P_{\bar B}(v)=\sum_{j\notin B}\max\{d_j(v),0\}$. Then
\begin{equation}
 G(a)-G(f)=A_B
 -\underbrace{[L_{\bar B}(a)-L_{\bar B}(f)]}_{\text{extra outside loss}}
 -\underbrace{[P_{\bar B}(f)-P_{\bar B}(a)]}_{\text{forgone outside improvement}}.
 \label{eq:focused-gain-tradeoff}
\end{equation}
These signed differences need not all be positive at every step. Forgone improvement is distinct from regression: if $f$ improves a QA by $0.04$ while $a$ leaves it unchanged, $a$ forgoes $0.04$ without lowering its score.

\begin{table}[htbp]
\centering\small\setlength{\tabcolsep}{5pt}
\caption{Mean decomposition at strict replay disagreements. Gain sums are divided by the image's total QA count and multiplied by $100$: units are auxiliary-score probability points per step, not actual-QA accuracy. Components satisfy Eq.~(\ref{eq:focused-gain-tradeoff}) before rounding.}
\label{tab:focused-gain-tradeoff}
\begin{tabular}{lrrrr}\hline
 & Focused & Extra outside & Forgone outside & Net gain\\
Selection rule & advantage & loss & improvement & difference\\\hline
QA-MBG, $\alpha=0.2$ & $+0.4806$ & $0.5832$ & $1.2075$ & $-1.3100$\\
Top-$k$, $\eta=0.2$ & $+0.9496$ & $1.3283$ & $0.5267$ & $-0.9053$\\\hline
\end{tabular}
\end{table}

For QA-MBG, forgone improvements contribute more than extra losses on unsampled QAs; for top-$20\%$, extra losses dominate. QA-MBG can thus prioritize its sampled questions without increasing the \emph{total} positive gain. Explaining both methods solely as preserving large gains canceled by other questions would obscure this difference.

\subsection{Interpretation and limitations}
Both QA-MBG and top-$k$ relax the requirement that every addition maximize immediate all-QA improvement. Sampling changes candidate ranking and hence the greedy path; it does not estimate the all-QA marginal more accurately than full evaluation. Although QA-MBG is unbiased for a fixed candidate (Eq.~\ref{eq:unbiased-marginal}), maximizing that estimate need not preserve the full-greedy action. Completed runs show that alternative paths can reach better final auxiliary scores and actual-QA performance; the same-state analysis characterizes the local trade-offs these rules permit.

Top-$k$ is complementary evidence for focused updates, not proof of QA-MBG's exact mechanism. The diagnostics do not establish that a particular regression or forgone improvement causes later success, or that sampling escapes a specific greedy trap. Replay covers one QA-MBG seed through ratio $0.15$, without following alternative actions forward. The downstream comparison covers one backbone and the small set, with single-run full-greedy/top-$k$ controls and no top-$k$ long-form evaluation. We interpret the evidence as a characterization of search behavior, not a causal proof or a universally optimal QA fraction.
\FloatBarrier

\section{Generated-QA Removal Ablation}
\label{app:qa-removal}

\paragraph{Protocol.}
We test whether the advantage depends on generated QAs that overlap with actual evaluation questions by changing the search pool and recomputing subset trajectories. The Qwen3-VL experiment uses the small set and evaluates all $392$ actual QAs in every condition. The original condition uses the main-generation auxiliary pool. The overlap-removal condition deletes generated QAs identified as overlapping with actual questions; the random-removal control deletes a matched proportion without targeting overlap. Each condition uses the probability objective and QA-MBG with $\alpha=0.2$. The actual-QA set is held fixed while the auxiliary search pool changes. Actual questions are used to define the diagnostic removal condition, but do not score candidate subsets.

\begin{figure}[H]
\centering
\includegraphics[width=\linewidth]{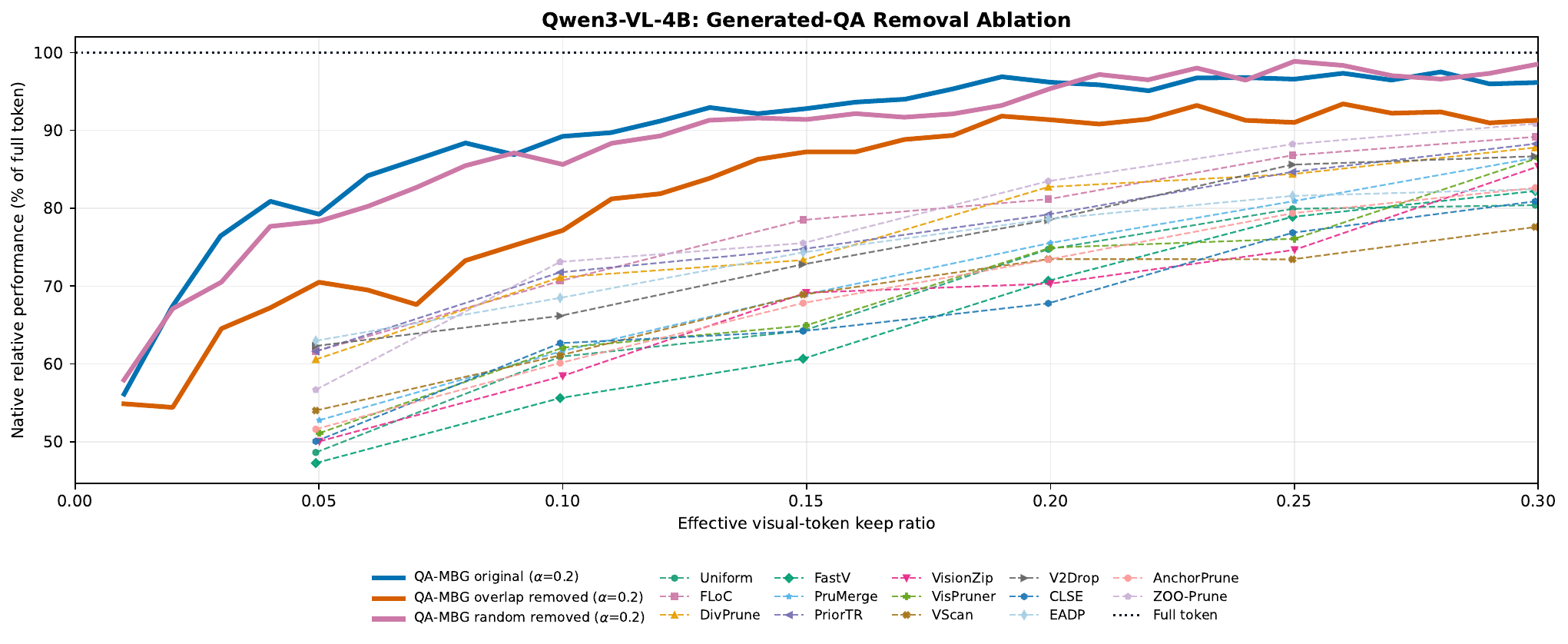}
\caption{Actual-QA performance after targeted or random auxiliary-QA removal.}
\label{fig:qa-removal}
\end{figure}

Scores are normalized by the common small-set full-token macro score $0.722316$. The best baseline is chosen separately at each budget from Table~\ref{tab:greedy-sweep-baselines}. The original-pool trajectory comes from the main-generation pool restricted to the small set, rather than the earlier pool used in the QA-fraction sweep.

\begin{table}[H]
\centering\small
\setlength{\tabcolsep}{5pt}
\caption{Removal ablation: relative performance and remaining headroom (points).}
\label{tab:qa-removal}
\begin{tabular}{lrrrrrr}
\hline
Method & $0.05$ & $0.10$ & $0.15$ & $0.20$ & $0.25$ & $0.30$ \\ \hline
Original pool & 79.22 & 89.23 & 92.80 & 96.18 & 96.58 & 96.15 \\
Random removal & 78.33 & 85.62 & 91.40 & 95.37 & 98.86 & 98.52 \\
Overlap removal & 70.49 & 77.15 & 87.24 & 91.36 & 91.02 & 91.31 \\
Best pruning baseline & 62.97 & 73.11 & 78.49 & 83.47 & 88.23 & 90.86 \\
Remaining headroom & 7.52 & 4.04 & 8.76 & 7.88 & 2.79 & 0.46 \\
\hline
\end{tabular}
\end{table}

\paragraph{Results.}
At ratio $0.20$, original, random-removal, and overlap-removal scores are $96.18$, $95.37$, and $91.36$. Overlap removal retains $7.88$ points of headroom over the best baseline, or about $62\%$ of the original margin at that budget. Random removal is close to the original trajectory at larger budgets; targeted removal is lower but remains above the best baseline at all six checkpoints. The remaining margin decreases to $0.46$ points at $0.30$.

\paragraph{Interpretation.}
Targeted removal hurts more than matched-rate random removal, indicating sensitivity to overlap-bearing supervision beyond pool size alone. Residual headroom remains, but removing a QA may also remove evidence useful to other questions. The result therefore does not decompose memorization and general understanding or guarantee elimination of all semantic overlap. Each removal condition is a single search run.

\FloatBarrier

\section{Dataset-Agnostic Auxiliary QA Generation}
\label{app:dataset-agnostic}

\subsection{Protocol and controlled prompt changes}
We compare benchmark-conditioned and dataset-agnostic auxiliary QA generation with Qwen3-VL on the small set and all $392$ actual QAs. Both use GPT-5.6-SOL, the probability objective, and QA-MBG with $\alpha=0.2$ and all remaining token candidates. The conditioned pool is shared with the count and removal controls. The dataset-agnostic prompt is reproduced below.

Dataset-agnostic generation removes every explicit benchmark-specific input: the dataset name, benchmark-character description, dataset-specific question patterns and answer-style instructions, avoidance rules, and few-shot QAs from other images of that benchmark. It replaces these with a generic inventory of visual evidence types, including entities, attributes, readable text and values, counting, spatial relationships, structure, comparisons, and short visual reasoning. Actual evaluation questions and answers are absent from both generation prompts.

The grounding, answer format, and diversity requirements remain fixed: questions must require inspecting the image, answers must be supported by visible evidence, unsupported or ambiguous questions are excluded, independent facts are preferred over paraphrases, and at most two substantially different questions may target one fact. Both prompts request about $32$ short-answer QAs, preferably $28$--$36$, with a JSON-only output schema. The target is soft, allowing fewer questions when warranted by the image. The resulting pools contain $6{,}062$ conditioned and $6{,}024$ dataset-agnostic QAs (means $33.68$ and $33.47$ per image). Thus the requested count and format are controlled, but the realized per-image counts and QA content need not match exactly. This is an ablation of the complete benchmark-conditioning package, not of few-shot examples alone or of equal-count QA pools.

\begin{figure}[H]
\centering
\includegraphics[width=\linewidth]{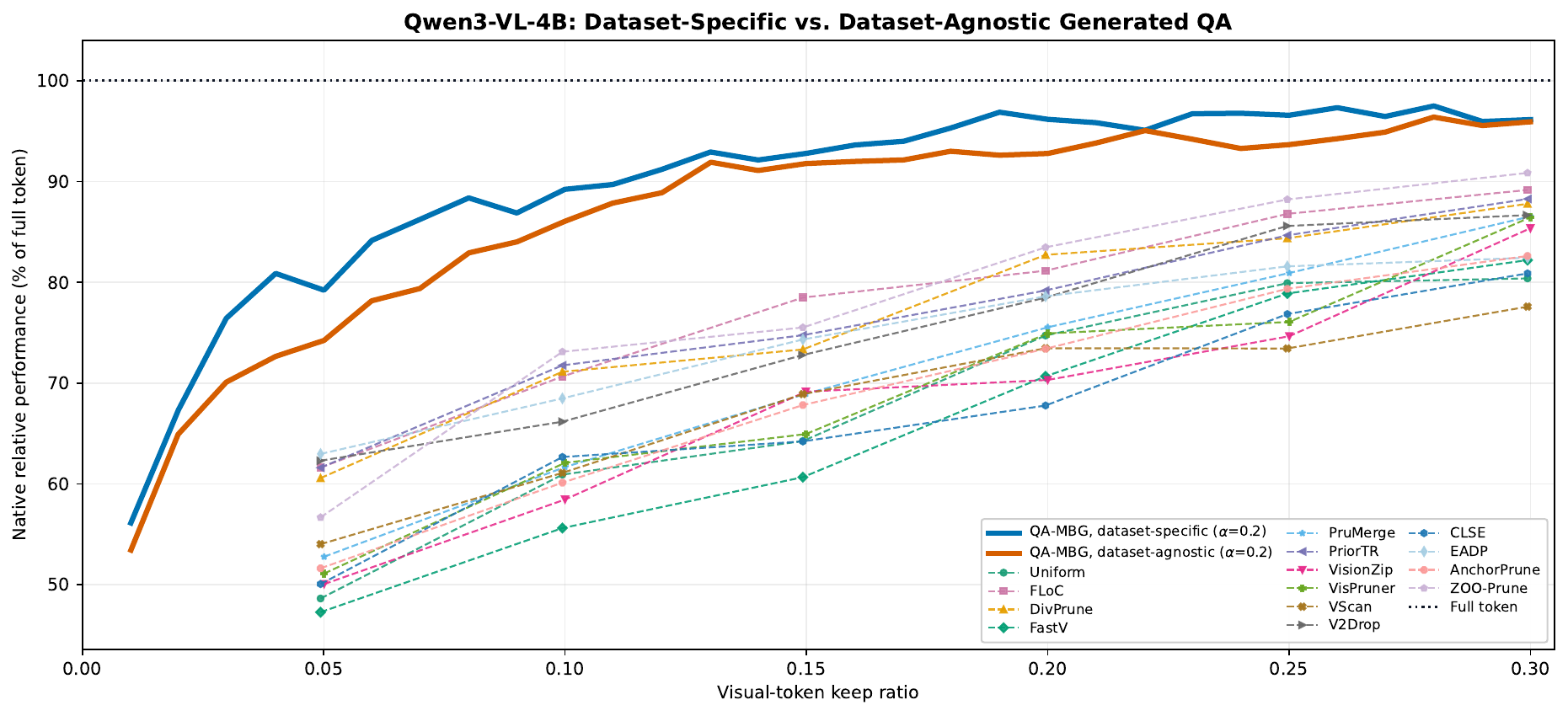}
\caption{Actual-QA performance with and without benchmark-conditioned generation.}
\label{fig:dataset-agnostic-overall}
\end{figure}

\subsection{Results and interpretation}
Table~\ref{tab:dataset-agnostic} reports native benchmark macro scores normalized by the same full-token reference, $0.722316$. Both searched configurations retain one shared subset per image and budget, reused for all actual QAs. Dataset-agnostic search exceeds the strongest evaluated pruning baseline at all six standard budgets, with margins from $5.07$ to $13.30$ relative-performance points. At $0.20$, conditioning improves the score from $92.79$ to $96.18$, but the dataset-agnostic result still exceeds the strongest baseline by $9.31$ points. At $0.30$, the two search results are close ($95.93$ versus $96.15$).

\begin{table}[H]
\centering\small\setlength{\tabcolsep}{4pt}
\caption{Conditioning ablation: relative performance and baseline headroom (points).}
\label{tab:dataset-agnostic}
\begin{tabular}{lrrrrrr}\hline
Condition / statistic & $0.05$ & $0.10$ & $0.15$ & $0.20$ & $0.25$ & $0.30$ \\ \hline
Benchmark-conditioned & 79.22 & 89.23 & 92.80 & 96.18 & 96.58 & 96.15 \\
Dataset-agnostic & 74.24 & 86.07 & 91.79 & 92.79 & 93.67 & 95.93 \\
Strongest pruning baseline & 62.97 & 73.11 & 78.49 & 83.47 & 88.23 & 90.86 \\
Conditioning gain & 4.98 & 3.16 & 1.01 & 3.39 & 2.91 & 0.22 \\
Dataset-agnostic headroom & 11.27 & 12.96 & 13.30 & 9.31 & 5.44 & 5.07 \\
\hline\end{tabular}
\end{table}

\begin{figure}[H]
\centering
\includegraphics[width=\linewidth,height=0.64\textheight,keepaspectratio]{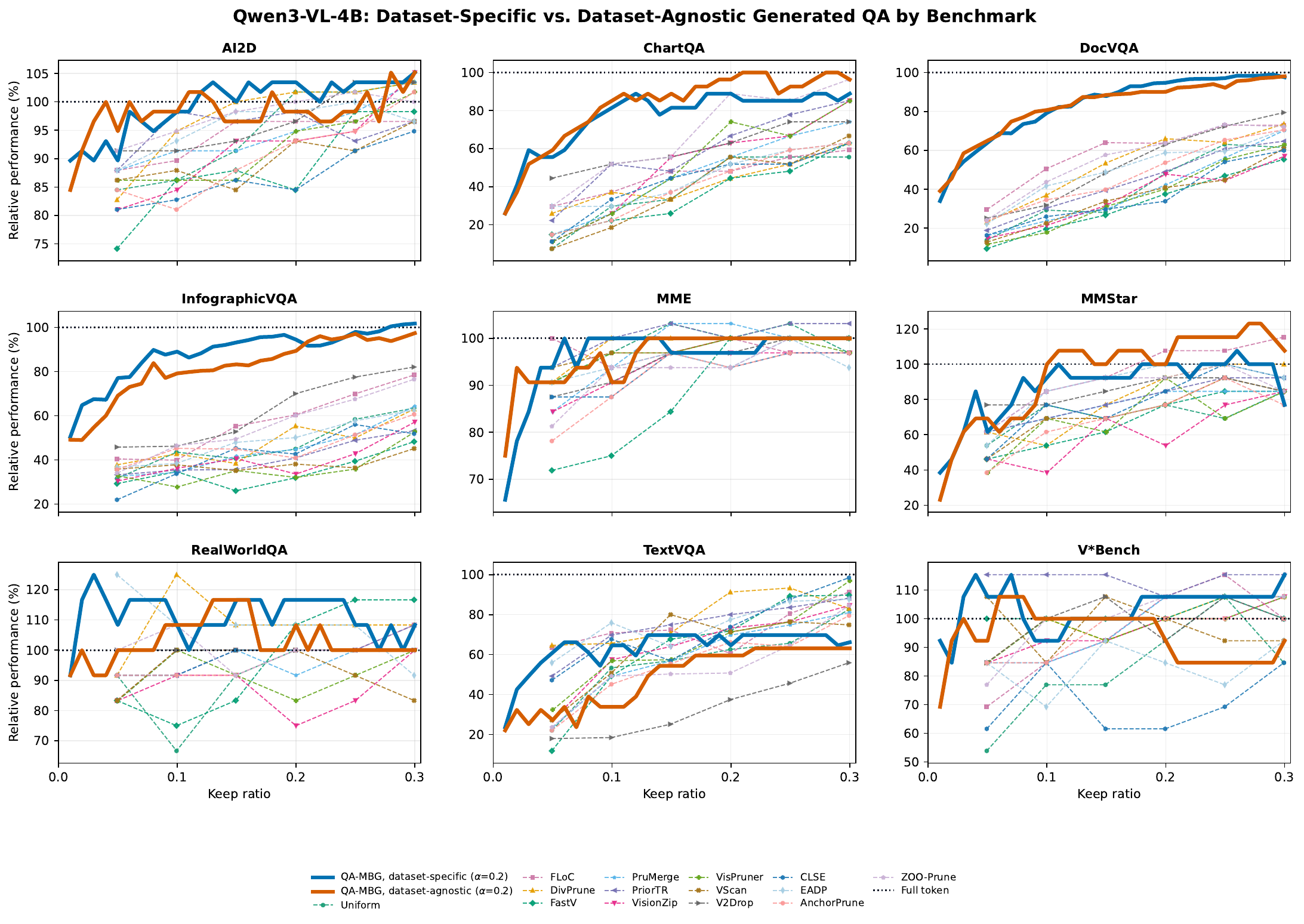}
\caption{Conditioning ablation by benchmark.}
\label{fig:dataset-agnostic-breakdown}
\end{figure}

Conditioning helps InfographicVQA, TextVQA, and parts of V$^*$Bench, while dataset-agnostic generation can be stronger on ChartQA and MMStar (Figure~\ref{fig:dataset-agnostic-breakdown}). These single-run results show useful headroom without explicit benchmark metadata. The image still reveals its content type, and removing benchmark guidance does not remove QA overlap; the conditioning and overlap-removal interventions are separate rather than joint controls.

\FloatBarrier

\subsection{Exact dataset-agnostic generation prompt}
The following text accompanies the image. No benchmark name, task profile, or same-benchmark examples are appended.
\begin{lstlisting}[basicstyle=\footnotesize\ttfamily,breaklines=true,columns=fullflexible]
You are generating a diverse set of image-grounded short-answer question-answer pairs from the attached image.

Inspect the image without assuming that it belongs to any particular dataset, benchmark, or task category. Generate questions that collectively test the visible information needed to understand this specific image.

Requirements:
- Every answer must be directly supported by visible evidence in the image.
- A question must require inspecting this particular image. Reject questions whose answer is predictable from general knowledge, stereotypes, or the wording of the question alone.
- Use short open-ended answers, normally 1-12 words.
- Do not use multiple-choice questions.
- Do not ask about evidence that is unreadable, ambiguous, hidden, or absent.
- Do not invent text, values, objects, relationships, or background context.
- Cover independent visual facts before producing alternative phrasings.
- Use at most 2 questions for the same underlying fact. A second question must be a substantially different formulation, not a minor rewording.
- Avoid generic questions such as asking whether a common object is present when more specific visible evidence can be tested.

Coverage plan:
- Return about 32 QA pairs, preferably 28-36 when the image supports them.
- First identify the image's visible content and organization internally.
- Select applicable evidence types from: object or entity identity, attributes, readable text, numerical values, counting, spatial relationships, interactions, document or diagram structure, comparison, ordering, and concise multi-step visual reasoning.
- Cover different regions and elements of the image.
- Use only evidence types that genuinely apply to the image; do not fill quotas or fabricate questions to reach the requested count.
- Prefer questions with a specific visual anchor such as an object, region, label, value, relation, or state.
- If fewer than 28 reliable independent questions are available, return fewer.

Return JSON only with this schema:
{
  "qas": [
    {
      "question": "...",
      "answer": "...",
      "question_type": "short descriptive category"
    }
  ]
}

Do not include explanations or text outside the JSON object.
\end{lstlisting}
\clearpage
\section{Robustness to the Auxiliary-QA Generator}
\label{app:generator-robustness}

\begin{figure}[htbp]
\centering
\includegraphics[width=\linewidth]{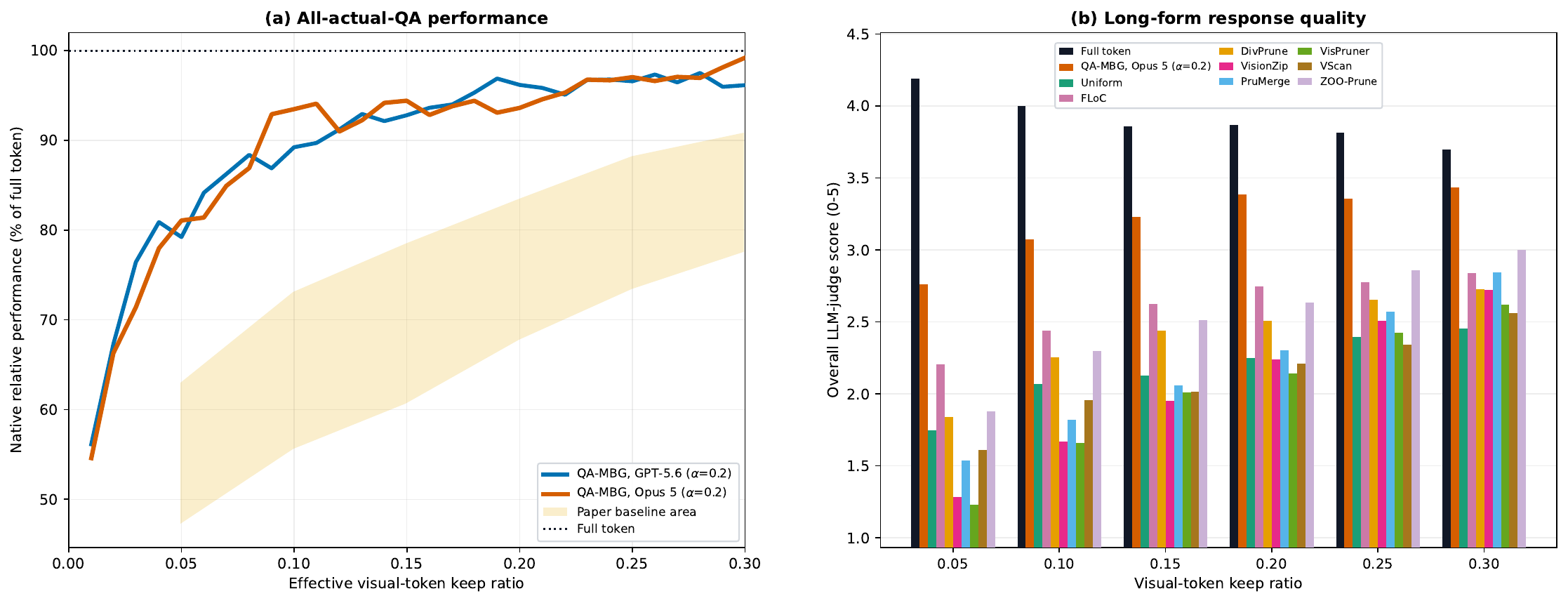}
\caption{Auxiliary-QA generator ablation on the small set.}
\label{fig:opus5-generator}
\end{figure}

We test whether the searched frontier depends on the frontier VLM used to generate auxiliary QAs. On the Qwen3-VL small set, we replace GPT-5.6-SOL with Opus~5 while keeping the generation prompt, QA-MBG search configuration ($\alpha=0.2$), probability objective, evaluation images, and downstream protocols fixed. Figure~\ref{fig:opus5-generator} compares the resulting trajectory with GPT-5.6 generation and the evaluated pruning baselines.

The generator change preserves the main result. Across the six reported budgets, the Opus~5 and GPT-5.6 trajectories differ by only $2.30$ relative-performance points on average, and both exceed the strongest pruning baseline at every budget. Opus~5 is higher at five of the six checkpoints, reaching $99.21\%$ of full-token actual-QA performance at keep ratio $0.30$. Its subsets also remain the strongest compressed representation in the long-form evaluation at every checkpoint, with overall judge scores increasing from $2.76$ at $0.05$ to $3.43$ at $0.30$. Thus the observed headroom and transfer to open-ended descriptions are not specific to GPT-5.6 auxiliary QAs.

\begin{table}[htbp]
\centering\small\setlength{\tabcolsep}{3.5pt}
\caption{Generator ablation at the six reported budgets. Actual QA is relative to full-token performance; long-form scores use the $0$--$5$ judge scale.}
\label{tab:opus5-generator}
\begin{tabular}{lrrrrrr}\hline
Metric / generator & $0.05$ & $0.10$ & $0.15$ & $0.20$ & $0.25$ & $0.30$ \\ \hline
Actual QA, GPT-5.6 & 79.22 & 89.23 & 92.80 & 96.18 & 96.58 & 96.15 \\
Actual QA, Opus~5 & 81.07 & 93.48 & 94.41 & 93.61 & 97.04 & 99.21 \\
Long-form, Opus~5 & 2.76 & 3.07 & 3.23 & 3.39 & 3.36 & 3.43 \\
Best long-form baseline & 2.21 & 2.44 & 2.63 & 2.75 & 2.86 & 3.00 \\
\hline
\end{tabular}
\end{table}

\FloatBarrier

\clearpage
\section{Qualitative Examples}
\label{app:qualitative}
We show one favorable example per benchmark from the Qwen3-VL main set. Each page includes the image, retained tokens, actual-QA responses, and long-form descriptions with judge scores at the stated budget. These selected cases illustrate preserved or missed evidence; they are not a random sample. Aggregate claims are based on the full-cohort evaluations. One case per page preserves the readability of response details.

\clearpage
\includepdf[pages=1,scale=0.90,pagecommand={\thispagestyle{plain}},addtotoc={1,subsection,2,{AI2D},app:qual-ai2d}]{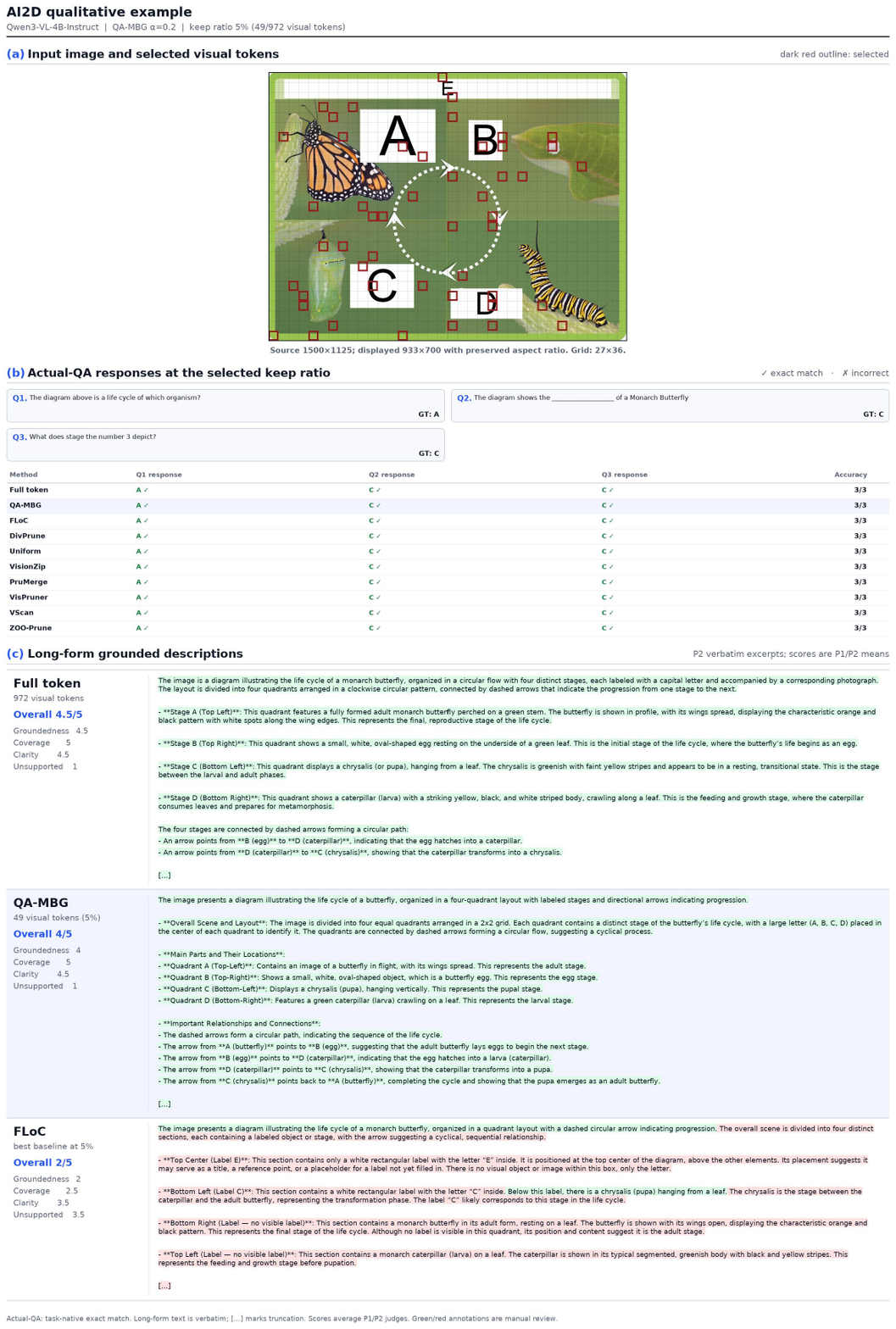}
\includepdf[pages=1,scale=0.90,pagecommand={\thispagestyle{plain}},addtotoc={1,subsection,2,{ChartQA},app:qual-chartqa}]{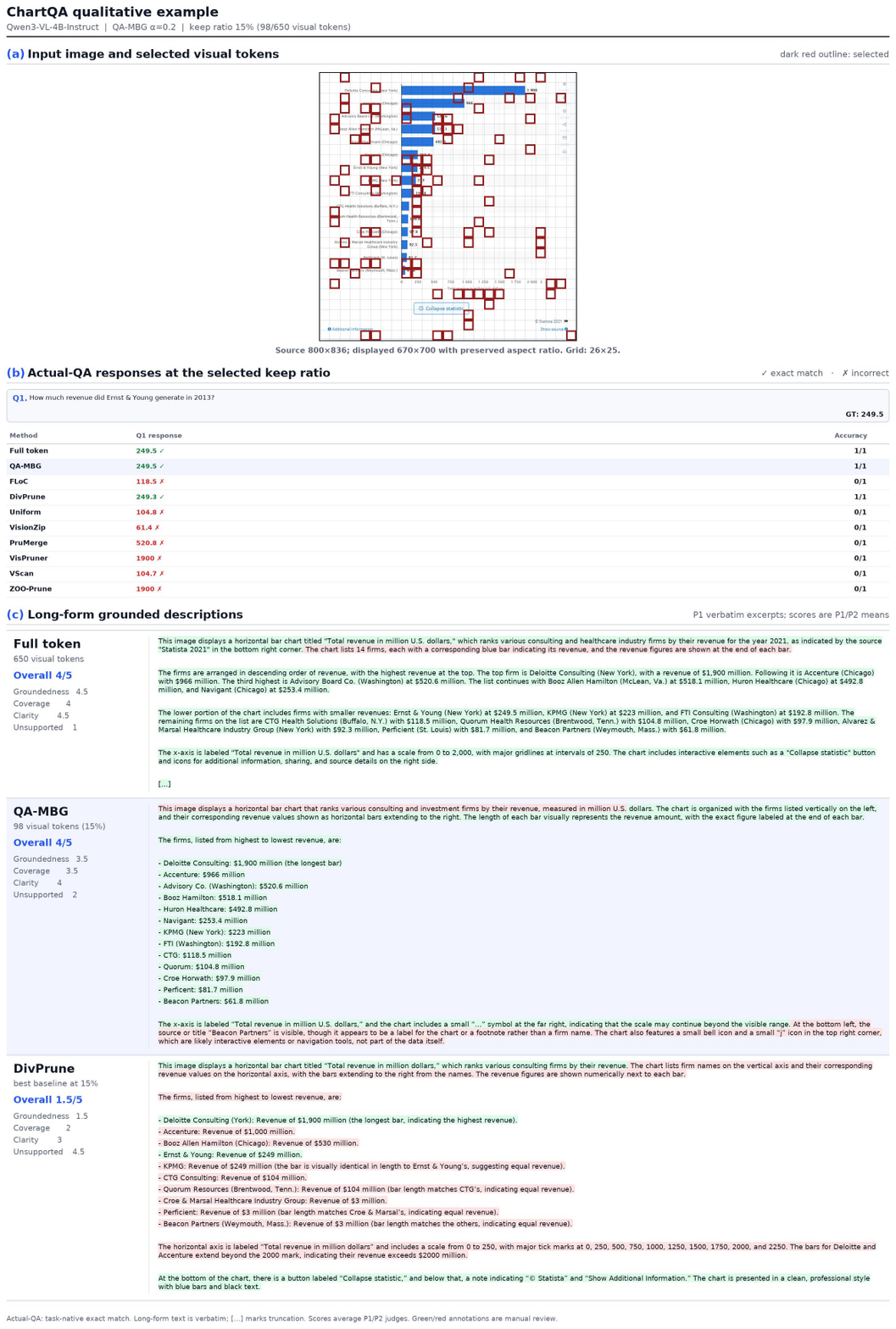}
\includepdf[pages=1,scale=0.90,pagecommand={\thispagestyle{plain}},addtotoc={1,subsection,2,{DocVQA},app:qual-docvqa}]{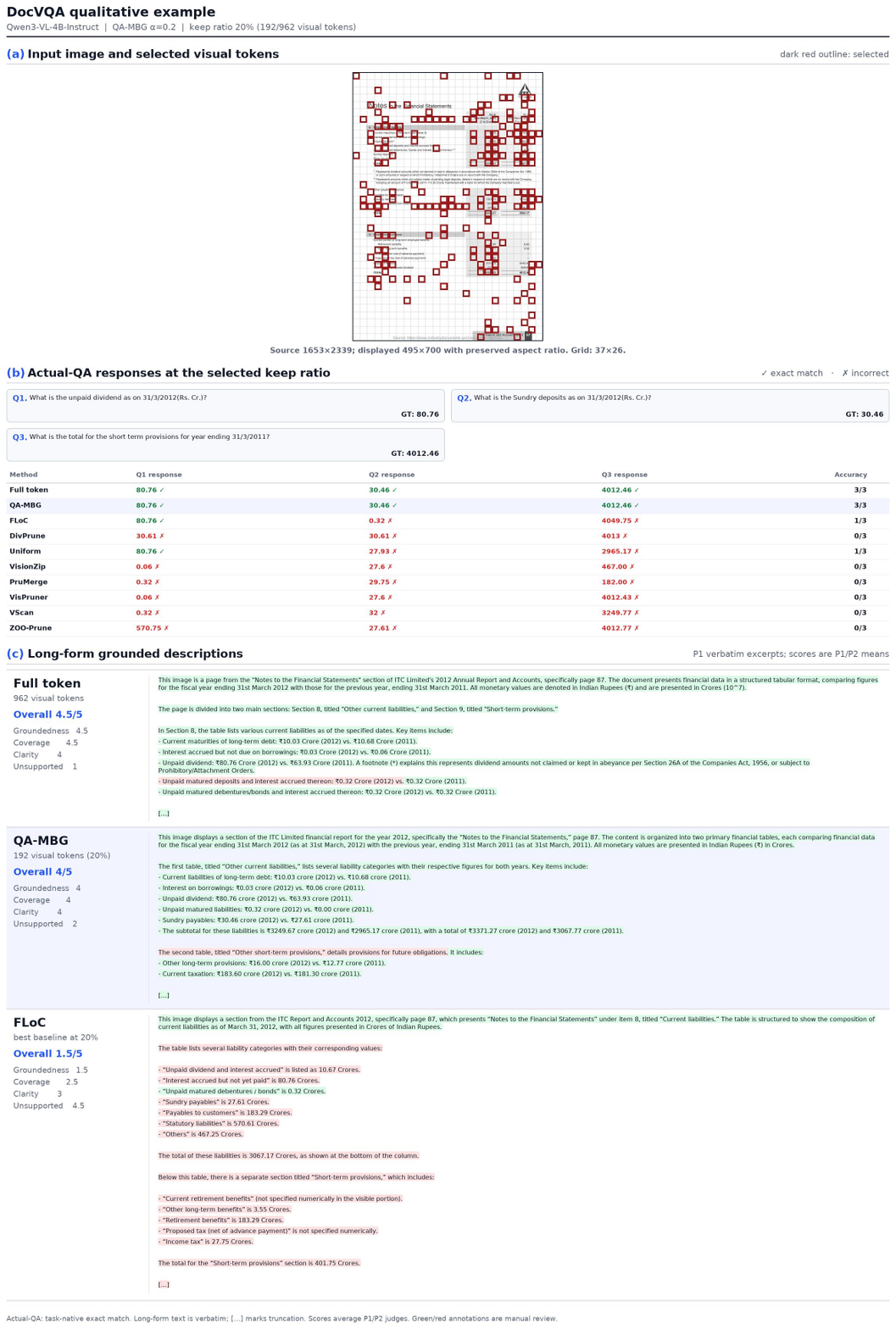}
\includepdf[pages=1,scale=0.90,pagecommand={\thispagestyle{plain}},addtotoc={1,subsection,2,{InfographicVQA},app:qual-infovqa}]{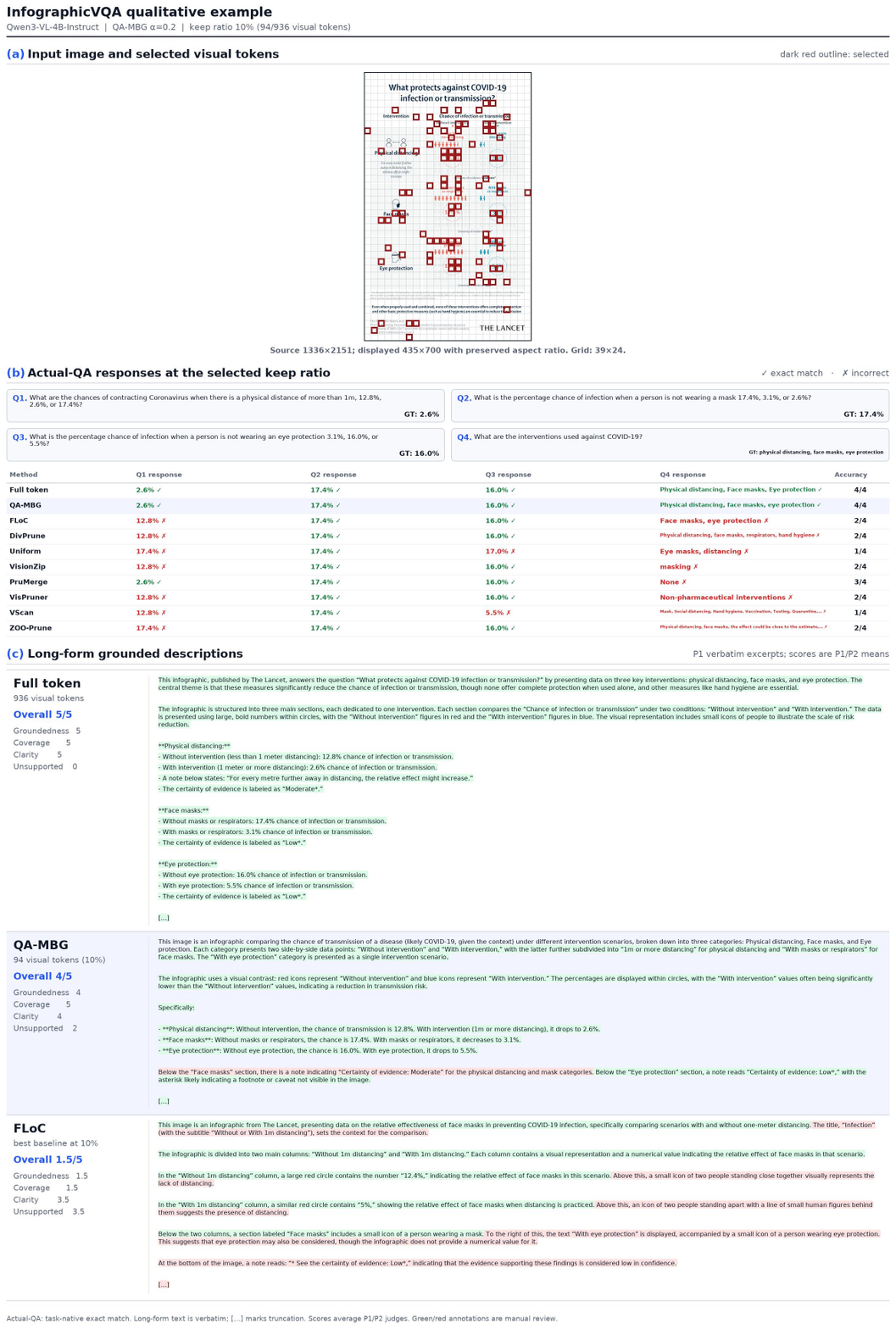}
\includepdf[pages=1,scale=0.90,pagecommand={\thispagestyle{plain}},addtotoc={1,subsection,2,{MME},app:qual-mme_natural}]{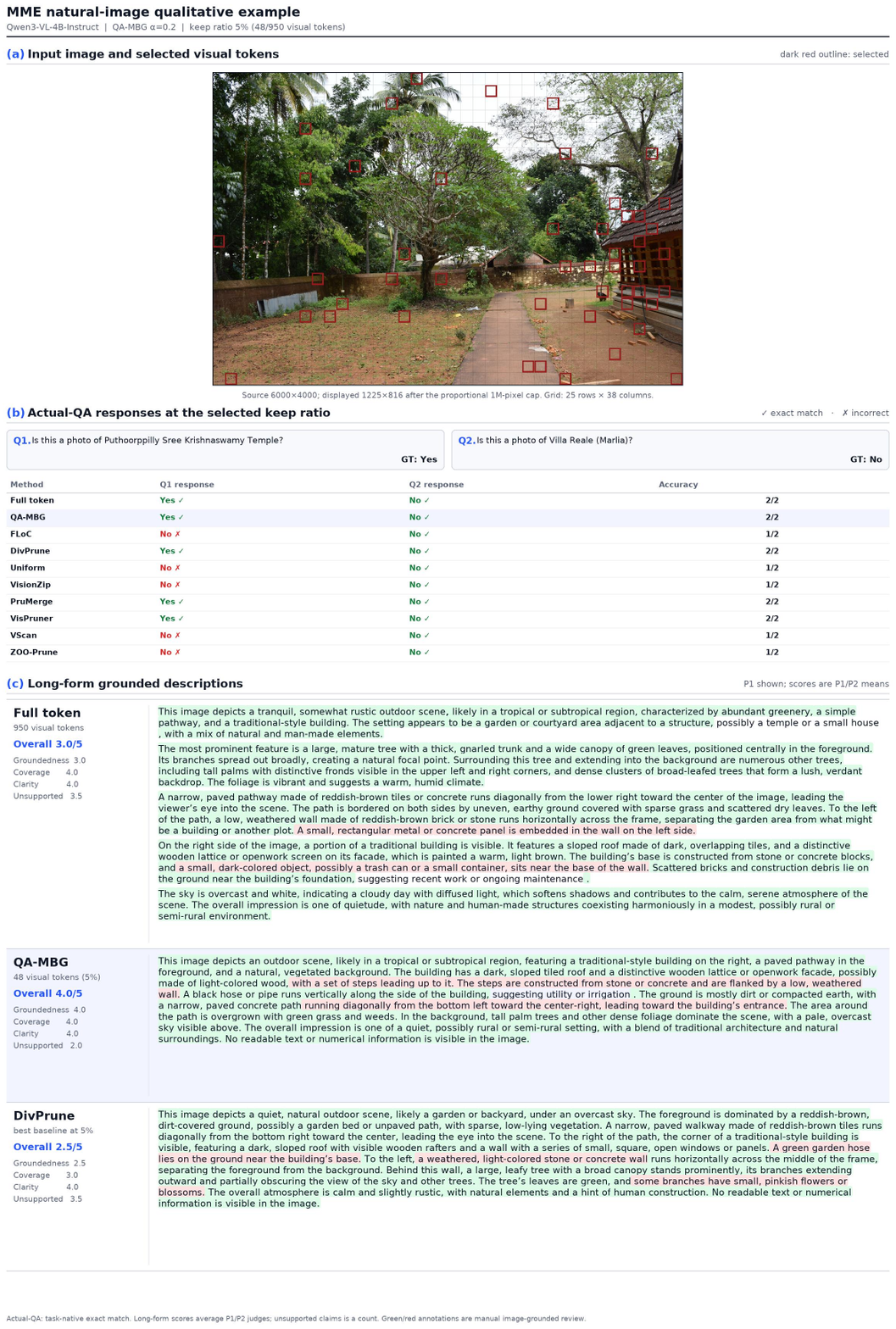}
\includepdf[pages=1,scale=0.90,pagecommand={\thispagestyle{plain}},addtotoc={1,subsection,2,{MMStar},app:qual-mmstar}]{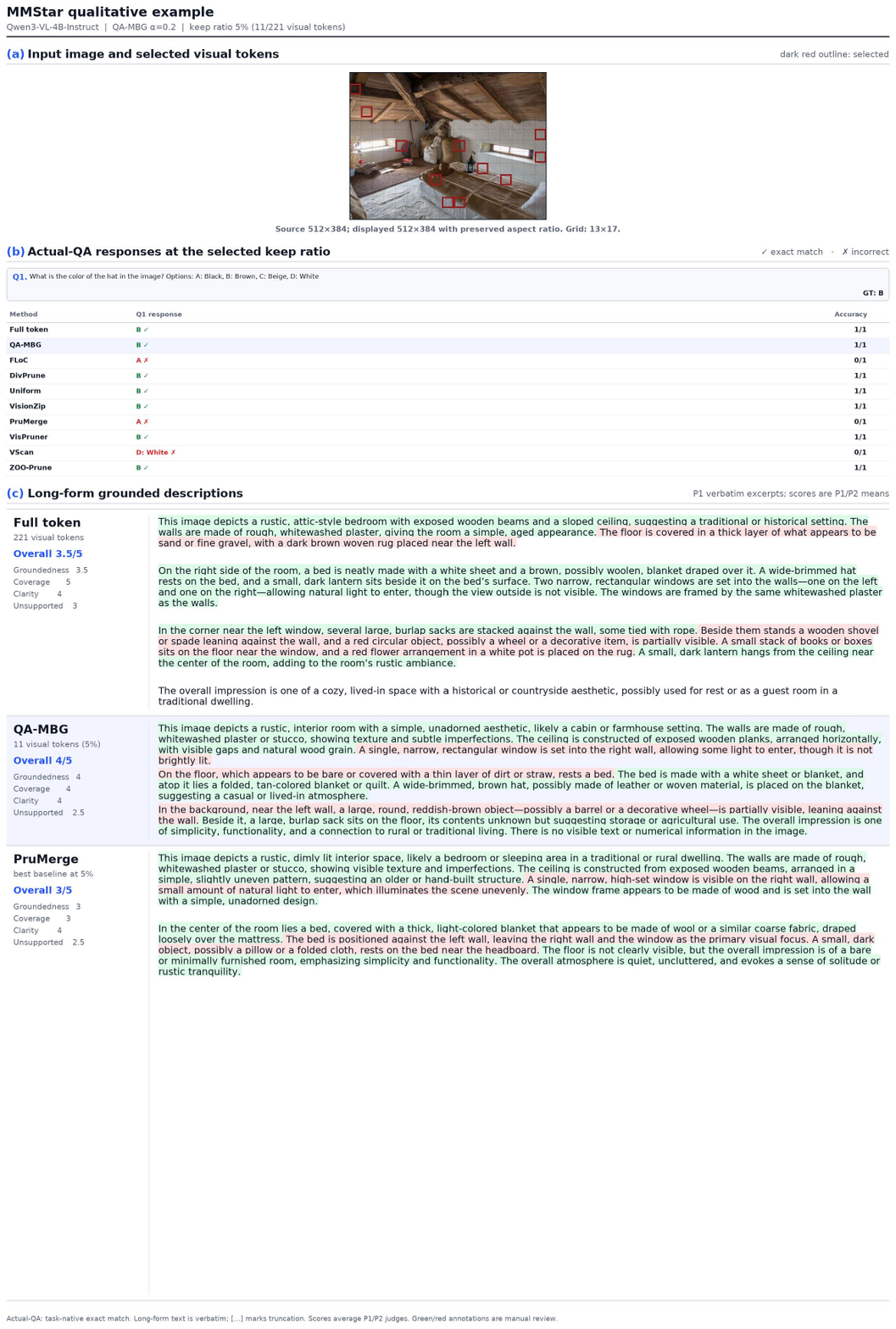}
\includepdf[pages=1,scale=0.90,pagecommand={\thispagestyle{plain}},addtotoc={1,subsection,2,{RealWorldQA},app:qual-realworldqa}]{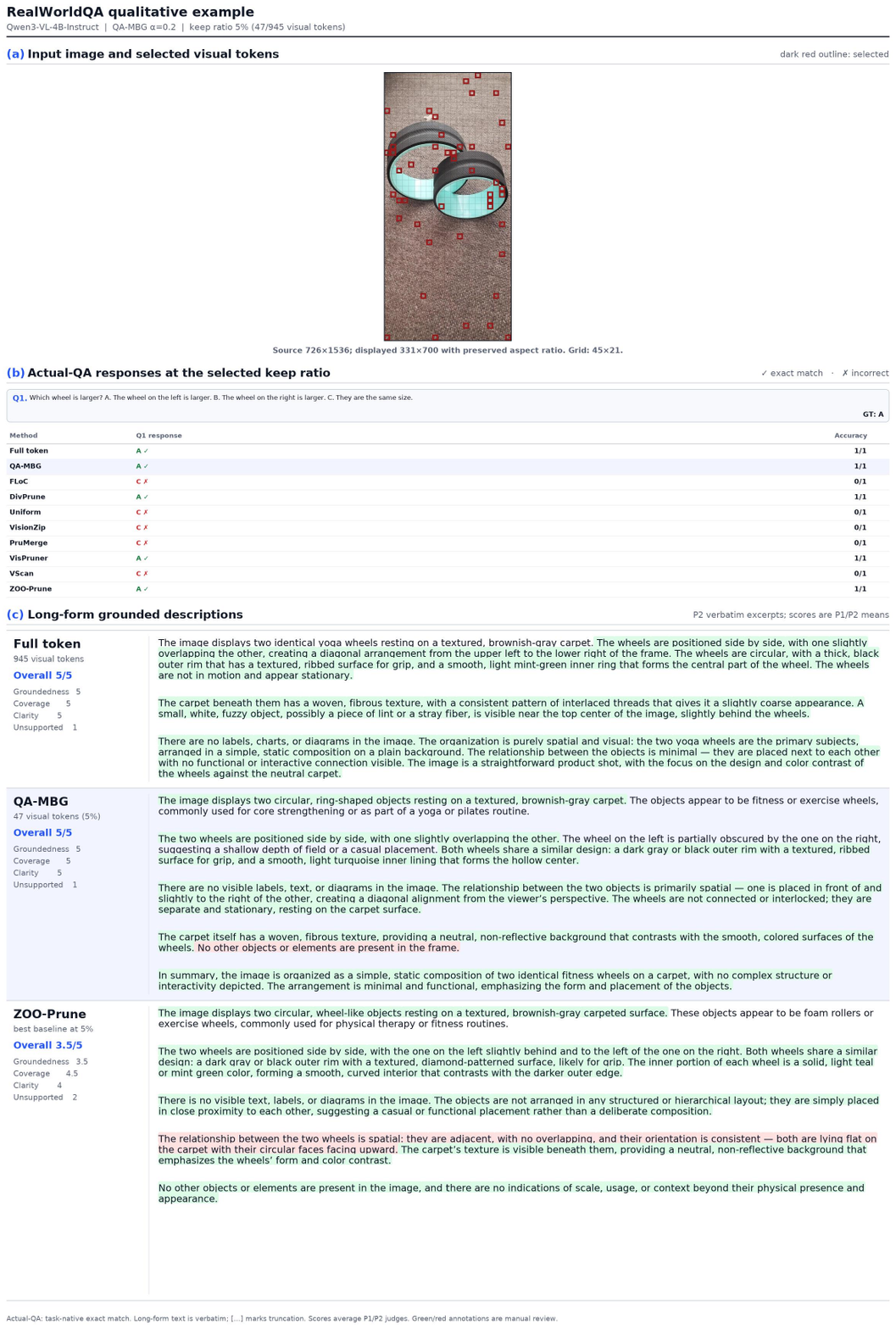}
\includepdf[pages=1,scale=0.90,pagecommand={\thispagestyle{plain}},addtotoc={1,subsection,2,{TextVQA},app:qual-textvqa}]{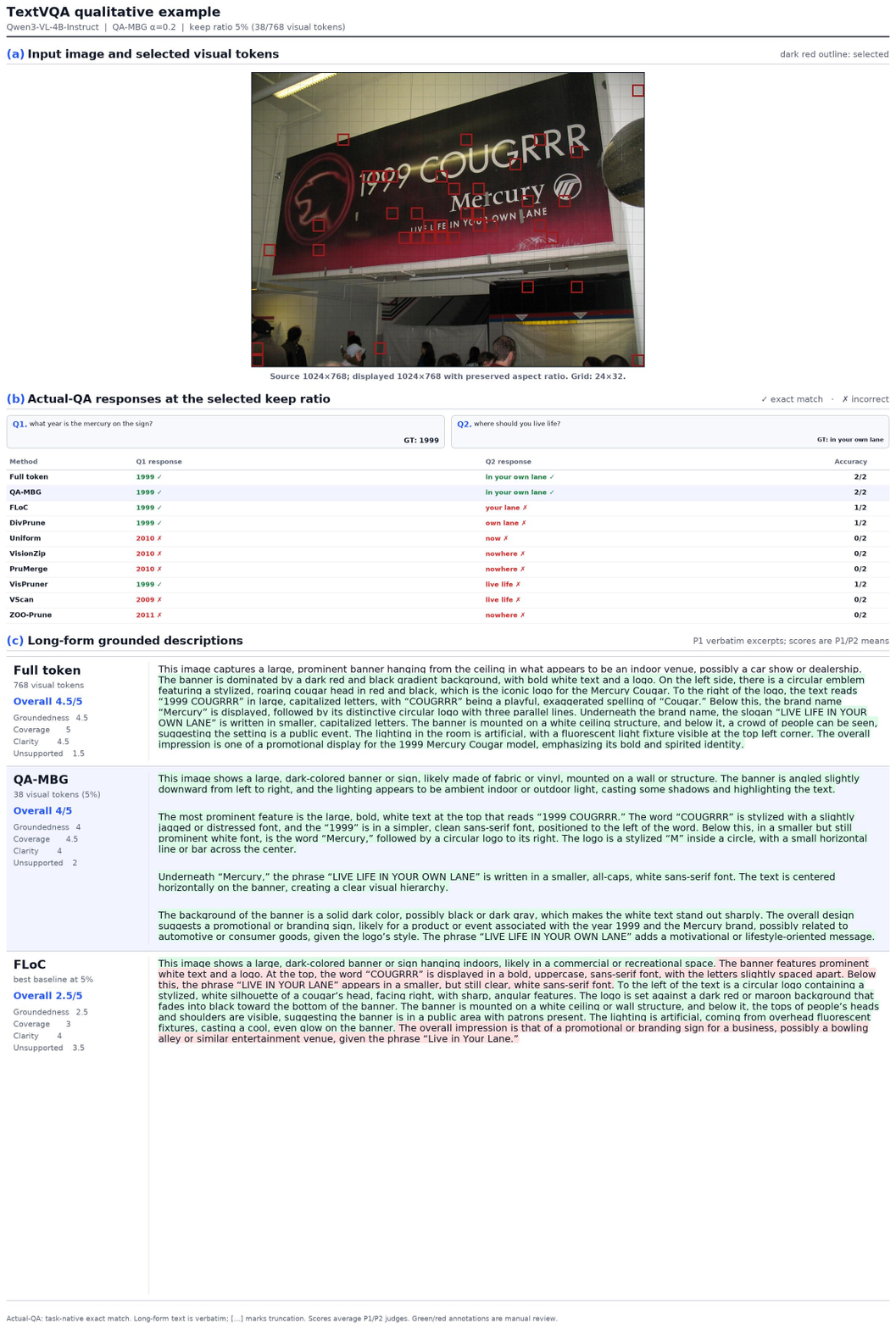}
\includepdf[pages=1,scale=0.90,pagecommand={\thispagestyle{plain}},addtotoc={1,subsection,2,{V$^*$Bench},app:qual-vstar}]{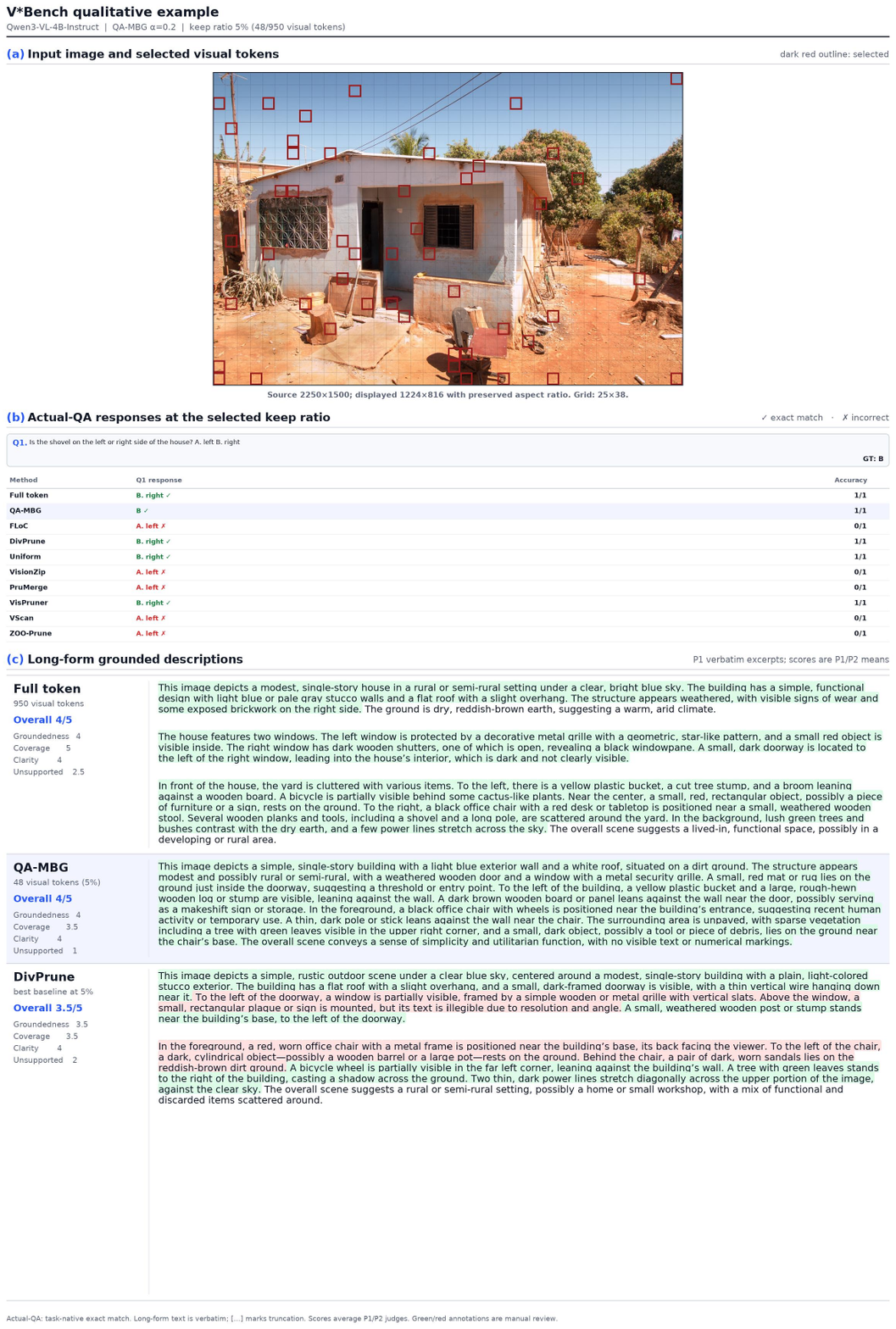}

\end{document}